\pdfoutput=1
\pdfminorversion=7
\documentclass[10pt,a4paper,onecolumn]{rho-class/rho}
\pdfmapfile{=syncfonts.map}
\usepackage[english]{babel}
\renewcommand{\titlefont}{\bfseries\color{rhocolor}\fontsize{17}{21}\sffamily\selectfont}
\geometry{left=2.4cm,right=2.4cm,top=2.2cm,bottom=2.4cm,headheight=18.7pt}
\setlength{\headwidth}{\textwidth}
\fancyhf{}\fancyhead[R]{\footerfont SYNC: Share the Symbol, Not the Vector}\fancyfoot[C]{\footerfont\thepage}
\fancypagestyle{firststyle}{\fancyhf{}\fancyfoot[C]{\footerfont\thepage}}
\setlength{\emergencystretch}{3em}
\definecolor{rhocolor}{HTML}{1B1A17}
\definecolor{rhoaccent}{HTML}{4A6378}
\definecolor{rhoabsback}{HTML}{EFEEEA}
\renewcommand{\absheadfont}{\color{rhoaccent}\normalfont\sffamily\fontsize{9}{11}\selectfont}
\AtBeginDocument{\hypersetup{allcolors=rhoaccent}}
\ifdefined\pdfsuppressptexinfo\pdfsuppressptexinfo=-1\fi
\usepackage{booktabs,longtable,array}
\usepackage{float}
\makeatletter
\let\jc@defaultaddtocurcol\@addtocurcol
\makeatother
\usepackage{flafter} %
\usepackage{placeins}
\usepackage{amsmath}\makeatletter\@ifpackageloaded{stix2}{}{\usepackage{amssymb}}\makeatother
\usepackage{graphicx}
\usepackage{xcolor}
\usepackage{colortbl}
\definecolor{parity}{RGB}{222,235,247}
\usepackage{tcolorbox}
\usepackage{accsupp}
\definecolor{findingblue}{HTML}{4A6378}
\newcounter{paperfinding}
\newcommand{\findingbadge}{\tcbox[on line,arc=2pt,boxrule=0pt,
  colback=findingblue,coltext=white,boxsep=0pt,
  left=4pt,right=4pt,top=1.5pt,bottom=1.5pt]{\small\bfseries Finding~\thepaperfinding}}
\newtcolorbox{findingbox}[1]{
  enhanced,arc=3pt,
  colback=findingblue!2!white,colframe=findingblue!65!white,boxrule=0.5pt,
  boxsep=0pt,left=8pt,right=8pt,top=4pt,bottom=4pt,
  before skip=4pt,after skip=4pt,
  before upper={\refstepcounter{paperfinding}\raggedright
    \findingbadge\hspace{5pt}\textbf{#1}\par\vspace{2pt}}
}
\usepackage{hyperref}
\usepackage{url}

\usepackage[utf8]{inputenc}
\DeclareUnicodeCharacter{03B3}{\ensuremath{\gamma}}
\DeclareUnicodeCharacter{03B1}{\ensuremath{\alpha}}
\DeclareUnicodeCharacter{03BC}{\ensuremath{\mu}}
\DeclareUnicodeCharacter{0394}{\ensuremath{\Delta}}
\DeclareUnicodeCharacter{00D7}{\ensuremath{\times}}
\DeclareUnicodeCharacter{2264}{\ensuremath{\leq}}
\DeclareUnicodeCharacter{2265}{\ensuremath{\geq}}
\DeclareUnicodeCharacter{2192}{\ensuremath{\rightarrow}}
\DeclareUnicodeCharacter{2212}{\ensuremath{-}}
\DeclareUnicodeCharacter{2011}{-}
\DeclareUnicodeCharacter{2013}{--}
\DeclareUnicodeCharacter{2014}{---}
\DeclareUnicodeCharacter{2018}{`}
\DeclareUnicodeCharacter{2019}{'}
\DeclareUnicodeCharacter{201C}{``}
\DeclareUnicodeCharacter{201D}{''}

\newcommand{\RECScenes}{572}
\newcommand{\RECDevScenes}{92}

\newcommand{\RECBaseRate}{97.7}

\newcommand{\RECGLMRate}{74.3}

\newcommand{\WIConjBase}{95.7}
\newcommand{\WIConjGLM}{41.8}
\newcommand{\WIConjGemma}{94.4}
\newcommand{\WIConjRecall}{99.9}

\usepackage{verbatim}
\makeatletter
\newcommand{\jcinputmacros}[1]{%
  \let\jc@savedtable\table \let\jc@savedendtable\endtable
  \let\table\comment \let\endtable\endcomment
  \input{#1}%
  \let\table\jc@savedtable \let\endtable\jc@savedendtable}
\newcommand{\jcinputtables}[1]{%
  \begingroup\let\newcommand\providecommand\input{#1}\endgroup}
\makeatother

\begin{table}[t]
\centering
\small
\resizebox{\textwidth}{!}{%
\begin{tabular}{llllll}
\toprule
dictionary (log key) & kind & seconds & hours & log & how read \\
\midrule
glm-4-9b & J-lens C=221 & 2560 & 0.71 & fit\_glm-4-9b\_inv2.log & integrated from per-batch s/batch \\
gpt-oss-20b & J-lens C=221 & 2304 & 0.64 & fit\_gpt-oss-20b\_inv2.log & integrated from per-batch s/batch \\
llama & J-lens C=221 & 1416 & 0.39 & fit\_llama\_inv2.log & integrated from per-batch s/batch \\
mistral-nemo & J-lens C=221 & 1800 & 0.50 & fit\_mistral-nemo\_inv2.log & integrated from per-batch s/batch \\
phi-4 & J-lens C=221 & 2824 & 0.78 & fit\_phi-4\_inv2.log & integrated from per-batch s/batch \\
qwen & J-lens C=221 & 1800 & 0.50 & fit\_qwen\_inv2.log & integrated from per-batch s/batch \\
qwen3.5-9b-base & J-lens C=221 & 20736 & 5.76 & fit\_qwen3.5-9b-base\_inv2.log & integrated from per-batch s/batch \\
qwen3.6-27b & J-lens C=221 & 44304 & 12.31 & fit\_qwen3.6-27b\_inv2.log & integrated from per-batch s/batch \\
l & J-lens ladder C=20,000 & 7866 & 2.18 & logd\_fit\_l\_c20k.log & log JSON tail \\
q & J-lens ladder C=20,000 & 11205 & 3.11 & logd\_fit\_q\_c20k.log & log JSON tail \\
14b & probe221 & 187 & 0.05 & logp\_fit\_14b.log & log JSON tail \\
32b & probe221 & 311 & 0.09 & logp\_fit\_32b.log & log JSON tail \\
glm & probe221 & 210 & 0.06 & logp\_fit\_glm.log & log JSON tail \\
llama & probe221 & 110 & 0.03 & logp\_fit\_llama.log & log JSON tail \\
llama-3.1-8b-instruct & probe221 & 88 & 0.02 & fit\_probe\_llama-3.1-8b-instruct.log & log JSON tail \\
phi & probe221 & 156 & 0.04 & logp\_fit\_phi.log & log JSON tail \\
qwen & probe221 & 149 & 0.04 & logp\_fit\_qwen.log & log JSON tail \\
qwen2.5-14b-instruct & probe221 & 132 & 0.04 & fit\_probe\_qwen2.5-14b-instruct.log & log JSON tail \\
qwen3-0.6b & probe221 & 118 & 0.03 & logp\_fit\_qwen3-0.6b.log & log JSON tail \\
qwen3-1.7b & probe221 & 122 & 0.03 & logp\_fit\_qwen3-1.7b.log & log JSON tail \\
qwen3-30b-a3b & probe221 & 154 & 0.04 & fit\_probe\_qwen3-30b-a3b.log & log JSON tail \\
qwen3-4b & probe221 & 153 & 0.04 & logp\_fit\_qwen3-4b.log & log JSON tail \\
qwen3.5-27b & probe221 & 831 & 0.23 & fit\_probe\_qwen3.5-27b.log & log JSON tail \\
qwen3.5-2b & probe221 & 270 & 0.08 & fit\_probe\_qwen3.5-2b.log & log JSON tail \\
qwen3.5-4b & probe221 & 360 & 0.10 & fit\_probe\_qwen3.5-4b.log & log JSON tail \\
qwen3.5-9b & probe221 & 366 & 0.10 & fit\_probe\_qwen3.5-9b.log & log JSON tail \\
r1-llama-8b & probe221 & 83 & 0.02 & fit\_probe\_r1-llama-8b.log & log JSON tail \\
r1-qwen-14b & probe221 & 123 & 0.03 & fit\_probe\_r1-qwen-14b.log & log JSON tail \\
\bottomrule
\end{tabular}}
\caption{Dictionary fit wall-clock per model, read from each fit's own log (the probe and 20k-row fits print their elapsed seconds; the J-lens logs are integrated from their per-batch wall-clock, approximate to one logging window). One GPU except where the log says otherwise.}
\label{tab:wire_setup}
\end{table}

\begin{table}[t]
\centering
\small
\resizebox{\textwidth}{!}{%
\begin{tabular}{llllllllllllll}
\toprule
sender & bank & k & dict & min\_cand & source\_dir & n & n\_matched & fallback\_rate & gis@4 grounded & gis@4 open & ev\_cov grounded & disc.-token recall grounded & disc.-token recall open \\
\midrule
llama-3.1-8b & ac-countries & 4 & acaug & 2 & m1\_dev\_k4post & 200 & 4.71 & 0.000 & 0.000 & 0.000 & 1.000 & 0.560 & 0.220 \\
llama-3.1-8b & ac-countries & 4 & acaug & 2 & m1\_dev\_mc2 & 200 & 4.71 & 0.000 & 0.000 & 0.000 & 1.000 & 0.350 & 0.045 \\
llama-3.1-8b & ac-countries & 4 & acaug & 8 & m1\_dev & 200 & 4.71 & 0.960 & 0.000 & 0.000 & 0.251 & 0.205 & 0.045 \\
llama-3.1-8b & ac-countries & 4 & ladder20000 & 8 & m1\_dev & 200 & 1.68 & 1.000 & 0.000 & 0.000 & 0.179 & 0.130 & 0.020 \\
llama-3.1-8b & ac-countries & 6 & acaug & 2 & m1\_dev\_k6 & 200 & 4.71 & 0.000 & -- & -- & 1.000 & 0.560 & 0.255 \\
llama-3.1-8b & ac-countries & 8 & acaug & 2 & m1\_dev\_k8 & 200 & 4.71 & 0.000 & -- & -- & 1.000 & 0.560 & 0.280 \\
llama-3.1-8b & ac-tipsheets & 4 & acaug & 2 & m1\_dev\_k4post & 200 & 51.98 & 0.000 & 0.000 & 0.000 & 1.000 & 0.000 & 0.000 \\
llama-3.1-8b & ac-tipsheets & 4 & acaug & 2 & m1\_dev\_mc2 & 200 & 51.98 & 0.000 & 0.000 & 0.000 & 1.000 & 0.000 & 0.000 \\
llama-3.1-8b & ac-tipsheets & 4 & acaug & 8 & m1\_dev & 200 & 51.98 & 0.000 & 0.000 & 0.000 & 1.000 & 0.000 & 0.000 \\
llama-3.1-8b & ac-tipsheets & 4 & ladder20000 & 8 & m1\_dev & 200 & 44.63 & 0.000 & 0.000 & 0.000 & 1.000 & 0.000 & 0.000 \\
llama-3.1-8b & ac-tipsheets & 6 & acaug & 2 & m1\_dev\_k6 & 200 & 51.98 & 0.000 & -- & -- & 1.000 & 0.000 & 0.000 \\
llama-3.1-8b & ac-tipsheets & 8 & acaug & 2 & m1\_dev\_k8 & 200 & 51.98 & 0.000 & -- & -- & 1.000 & 0.000 & 0.000 \\
llama-3.1-8b & arc-c & 4 & acaug & 2 & m1\_dev\_mc2 & 200 & 21.48 & 0.000 & 0.390 & 0.159 & 1.000 & 0.800 & 0.385 \\
llama-3.1-8b & arc-c & 4 & ladder20000 & 8 & m1\_dev & 200 & 21.25 & 0.030 & 0.394 & 0.167 & 0.989 & 0.795 & 0.385 \\
llama-3.1-8b & obqa & 4 & acaug & 2 & m1\_dev\_mc2 & 200 & 6.89 & 0.020 & 0.181 & 0.169 & 0.991 & 0.625 & 0.340 \\
llama-3.1-8b & obqa & 4 & ladder20000 & 8 & m1\_dev & 200 & 6.82 & 0.620 & 0.181 & 0.163 & 0.890 & 0.620 & 0.335 \\
qwen3-8b-base & ac-countries & 4 & acaug & 2 & m1\_dev\_k4post & 200 & 4.71 & 0.000 & 0.000 & 0.000 & 1.000 & 0.615 & 0.070 \\
qwen3-8b-base & ac-countries & 4 & acaug & 2 & m1\_dev\_mc2 & 200 & 4.71 & 0.000 & 0.000 & 0.000 & 1.000 & 0.285 & 0.010 \\
qwen3-8b-base & ac-countries & 4 & acaug & 8 & m1\_dev & 200 & 4.71 & 0.960 & 0.000 & 0.000 & 0.159 & 0.130 & 0.010 \\
qwen3-8b-base & ac-countries & 4 & ladder20000 & 8 & m1\_dev & 200 & 1.68 & 1.000 & 0.000 & 0.000 & 0.111 & 0.125 & 0.010 \\
qwen3-8b-base & ac-countries & 6 & acaug & 2 & m1\_dev\_k6 & 200 & 4.71 & 0.000 & -- & -- & 1.000 & 0.615 & 0.075 \\
qwen3-8b-base & ac-countries & 8 & acaug & 2 & m1\_dev\_k8 & 200 & 4.71 & 0.000 & -- & -- & 1.000 & 0.615 & 0.075 \\
qwen3-8b-base & ac-tipsheets & 4 & acaug & 2 & m1\_dev\_k4post & 200 & 51.98 & 0.000 & 0.000 & 0.000 & 1.000 & 0.000 & 0.000 \\
qwen3-8b-base & ac-tipsheets & 4 & acaug & 2 & m1\_dev\_mc2 & 200 & 51.98 & 0.000 & 0.000 & 0.000 & 1.000 & 0.000 & 0.000 \\
qwen3-8b-base & ac-tipsheets & 4 & acaug & 8 & m1\_dev & 200 & 51.98 & 0.000 & 0.000 & 0.000 & 1.000 & 0.000 & 0.000 \\
qwen3-8b-base & ac-tipsheets & 4 & ladder20000 & 8 & m1\_dev & 200 & 44.63 & 0.000 & 0.000 & 0.000 & 1.000 & 0.000 & 0.000 \\
qwen3-8b-base & ac-tipsheets & 6 & acaug & 2 & m1\_dev\_k6 & 200 & 51.98 & 0.000 & -- & -- & 1.000 & 0.000 & 0.000 \\
qwen3-8b-base & ac-tipsheets & 8 & acaug & 2 & m1\_dev\_k8 & 200 & 51.98 & 0.000 & -- & -- & 1.000 & 0.000 & 0.000 \\
qwen3-8b-base & arc-c & 4 & acaug & 2 & m1\_dev\_mc2 & 200 & 21.48 & 0.000 & 0.412 & 0.077 & 1.000 & 0.915 & 0.225 \\
qwen3-8b-base & arc-c & 4 & ladder20000 & 8 & m1\_dev & 200 & 21.25 & 0.030 & 0.428 & 0.078 & 0.986 & 0.910 & 0.225 \\
qwen3-8b-base & obqa & 4 & acaug & 2 & m1\_dev\_mc2 & 200 & 6.89 & 0.020 & 0.193 & 0.054 & 0.984 & 0.760 & 0.235 \\
qwen3-8b-base & obqa & 4 & ladder20000 & 8 & m1\_dev & 200 & 6.82 & 0.620 & 0.181 & 0.054 & 0.700 & 0.700 & 0.230 \\
\bottomrule
\end{tabular}}
\caption{M1 grounded-extraction dev calibration across $k$ and dictionary variants, plus a recomputed discriminating-token recall (support word with bank document frequency $\leq 3$ in the item's sender\_evidence text; empty support counts as 0) computed directly from the raw M1 rows and the referenced bench bank, never read from a doc. Rows are one per source summary.json (never merged), identified by source\_dir; see notes for (sender, bank, k, dict, min\_cand) keys realized by more than one directory.}
\label{tab:m1_extraction}
\end{table}

\begin{table}[t]
\centering
\small
\resizebox{\textwidth}{!}{%
\begin{tabular}{lllllll}
\toprule
model & role & eff. rank (of 221) & mean off-diag cos & p90 cos & max cos & distinct packets / MC episodes \\
\midrule
llama-3.1-8b & sender & 19.7 & 0.492 & 0.606 & 0.876 & 267/300 \\
qwen3-8b-base & sender & 14.7 & 0.532 & 0.665 & 0.883 & 237/300 \\
qwen3.5-9b & sender & 11.7 & 0.570 & 0.715 & 0.902 & 27/300 \\
gemma-3-27b & sender & 6.0 & 0.588 & 0.879 & 0.985 & 1/300 \\
qwen3.5-27b & receiver & 16.1 & 0.521 & 0.661 & 0.879 & -- \\
qwen3-32b & receiver & 24.7 & 0.468 & 0.573 & 0.852 & -- \\
\bottomrule
\end{tabular}}
\caption{Sender-side extraction health: distinct packets (distinct tuples of per-lane sorted concept-id supports, the pinned admission statistic) over the development bank against the probe221 dictionary's effective rank (participation ratio of the squared singular values of the row-normalised dictionary). Two working receivers' dictionaries are listed for scale.}
\label{tab:extraction_health}
\end{table}

\begin{table}[t]
\centering
\small
\resizebox{\textwidth}{!}{%
\begin{tabular}{llllll}
\toprule
anchor & $C$ & $E$ = replace $-$ random [95\% CI] & clusters & entities & write \\
\midrule
qwen3-8b-base & 221 & +6.156 [+5.456,+6.907] & 200 & 48 & dose 1 \\
qwen3-8b-base & 1,000 & +6.168 [+5.464,+6.924] & 200 & 48 & dose 1 \\
qwen3-8b-base & 5,000 & +6.191 [+5.492,+6.942] & 200 & 48 & dose 1 \\
qwen3-8b-base & 20,000 & +6.205 [+5.505,+6.958] & 200 & 48 & dose 1 \\
llama-3.1-8b & 221 & +1.795 [+1.538,+2.058] & 200 & 48 & dose 1 \\
llama-3.1-8b & 1,000 & +1.790 [+1.532,+2.053] & 200 & 48 & dose 1 \\
llama-3.1-8b & 5,000 & +1.801 [+1.539,+2.066] & 200 & 48 & dose 1 \\
llama-3.1-8b & 20,000 & +1.814 [+1.552,+2.083] & 200 & 48 & dose 1 \\
\bottomrule
\end{tabular}}
\caption{Alphabet ladder, writer side: matched-amplitude specificity $E$ of the probe-row dictionary at each alphabet size ($C{=}221$ is the J4 probe cell, the frozen G5 reference; larger $C$ are the ladder runs of the same construction at the same amplitude-matched write), entity-level bootstrap.}
\label{tab:alphabet_e}
\end{table}

\begin{table}[t]
\centering
\small
\resizebox{\textwidth}{!}{%
\begin{tabular}{lllll}
\toprule
anchor (sender) & extractor & $C$ & n & gold-in-support@4 \\
\midrule
llama-3.1-8b & J-lens & 221 & 150 & 0.727 \\
llama-3.1-8b & J-lens & 1,000 & 150 & 0.620 \\
llama-3.1-8b & J-lens & 5,000 & 150 & 0.393 \\
llama-3.1-8b & probe & 221 & 300 & 0.463 \\
llama-3.1-8b & probe & 5,000 & 150 & 0.127 \\
llama-3.1-8b & probe & 20,000 & 150 & 0.047 \\
qwen3-8b-base & J-lens & 221 & 150 & 0.453 \\
qwen3-8b-base & J-lens & 1,000 & 150 & 0.307 \\
qwen3-8b-base & J-lens & 5,000 & 150 & 0.200 \\
qwen3-8b-base & probe & 221 & 300 & 0.053 \\
qwen3-8b-base & probe & 5,000 & 150 & 0.000 \\
qwen3-8b-base & probe & 20,000 & 150 & 0.000 \\
\bottomrule
\end{tabular}}
\caption{Alphabet ladder, extraction side: gold-in-support at $k{=}4$ from natural text for the probe and J-lens extractors on both senders, per alphabet size (rows on disk only; absent points are not interpolated).}
\label{tab:alphabet_recall}
\end{table}

\begin{table}[t]
\centering
\small
\resizebox{\textwidth}{!}{%
\begin{tabular}{llllllllllllll}
\toprule
link & $M$ & source & none & none\_padded & $\Delta$ inject\_packet & $\Delta$ inject\_wrongset & $\Delta$ inject\_shuffled & $\Delta$ inject\_oracle & text\_packet & text\_concise & text\_matched\_text & text\_matched\_wire & text\_oracle \\
\midrule
L$\rightarrow$Q & 1 & jlens & 0.500 & 0.500 & +0.000 [+0.000,+0.000] & +0.000 [+0.000,+0.000] & +0.000 [+0.000,+0.000] & +0.000 [+0.000,+0.000] & 0.645 & 0.566 & 0.531 & 0.500 & 0.997 \\
L$\rightarrow$Q & 2 & jlens & 0.500 & 0.500 & +0.000 [+0.000,+0.000] & +0.000 [+0.000,+0.000] & +0.000 [+0.000,+0.000] & +0.000 [+0.000,+0.000] & 0.642 & 0.565 & 0.536 & 0.500 & 1.000 \\
L$\rightarrow$Q & 4 & jlens & 0.500 & 0.500 & +0.000 [+0.000,+0.000] & +0.000 [+0.000,+0.000] & +0.000 [+0.000,+0.000] & +0.000 [+0.000,+0.000] & 0.650 & 0.583 & 0.542 & 0.497 & 1.000 \\
L$\rightarrow$Q & 8 & jlens & 0.500 & 0.500 & +0.000 [+0.000,+0.000] & +0.000 [+0.000,+0.000] & +0.000 [+0.000,+0.000] & +0.000 [+0.000,+0.000] & 0.650 & 0.530 & 0.533 & 0.500 & 1.000 \\
L$\rightarrow$Q & 16 & jlens & 0.500 & 0.500 & +0.000 [+0.000,+0.000] & +0.000 [+0.000,+0.000] & +0.000 [+0.000,+0.000] & +0.000 [+0.000,+0.000] & 0.598 & 0.525 & 0.512 & 0.500 & 1.000 \\
L$\rightarrow$Q & 1 & mix pilot & 0.500 & 0.500 & +0.000 [+0.000,+0.000] & +0.000 [+0.000,+0.000] & +0.000 [+0.000,+0.000] & +0.000 [+0.000,+0.000] & 0.645 & 0.566 & 0.531 & 0.500 & 0.997 \\
L$\rightarrow$Q & 8 & mix pilot & 0.500 & 0.500 & +0.000 [+0.000,+0.000] & +0.000 [+0.000,+0.000] & +0.000 [+0.000,+0.000] & +0.000 [+0.000,+0.000] & 0.650 & 0.530 & 0.533 & 0.500 & 1.000 \\
Q$\rightarrow$L & 1 & jlens & 0.514 & 0.486 & -0.024 [-0.059,+0.010] & -0.024 [-0.059,+0.010] & -0.048 [-0.083,-0.014] & -0.034 [-0.072,+0.003] & 0.617 & 0.531 & 0.476 & 0.493 & 0.976 \\
Q$\rightarrow$L & 2 & jlens & 0.455 & 0.500 & +0.022 [-0.005,+0.047] & +0.026 [+0.000,+0.051] & +0.011 [-0.013,+0.034] & +0.018 [-0.005,+0.041] & 0.622 & 0.592 & 0.433 & 0.451 & 0.991 \\
Q$\rightarrow$L & 4 & jlens & 0.483 & 0.497 & -0.013 [-0.034,+0.009] & -0.006 [-0.028,+0.016] & -0.023 [-0.045,-0.002] & -0.017 [-0.039,+0.005] & 0.634 & 0.630 & 0.430 & 0.452 & 0.998 \\
Q$\rightarrow$L & 8 & jlens & 0.469 & 0.472 & -0.005 [-0.023,+0.014] & -0.003 [-0.023,+0.016] & -0.009 [-0.030,+0.011] & -0.005 [-0.023,+0.014] & 0.666 & 0.667 & 0.466 & 0.447 & 0.994 \\
Q$\rightarrow$L & 16 & jlens & 0.480 & 0.398 & -0.009 [-0.033,+0.013] & -0.008 [-0.032,+0.015] & -0.006 [-0.030,+0.017] & -0.014 [-0.037,+0.009] & 0.617 & 0.617 & 0.472 & 0.433 & 0.995 \\
\bottomrule
\end{tabular}}
\caption{Aggregation (H1) per $M$, both links: accuracy per arm at the calibrated dose (mult 1, qbits 0) and $\Delta$(inject $-$ none) with cluster-bootstrap 95\% CIs (nb=8000, seed 20260802); the mix-channel pilot (jlens extract + probe write) is listed as its own source and never pooled with the jlens rows. The L$\to$Q block prints $+0.000$ with a zero-width interval on every injection arm, which is this paper's own signature for an arm that never fired --- it is not that here. Those writes fire (24 writes per row, $\lVert\delta\rVert$ 16.8--134.2), and the receiver's yes/no answer is simply identical to no-message on 764 of 764 episodes under the ORACLE injection, while text moves the same rows (text oracle at the right of the table). The zero is a measured no-op of the write on this endpoint, not an unfired arm.}
\label{tab:aggregation}
\end{table}

\begin{table}[t]
\centering
\small
\resizebox{\textwidth}{!}{%
\begin{tabular}{lllllll}
\toprule
link & $M$ & source & dose $\times$ & $\Delta$ inject\_packet & $\Delta$ inject\_oracle & $\Delta$ inject\_shuffled \\
\midrule
L$\rightarrow$Q & 1 & jlens & 0.5 & +0.000 [+0.000,+0.000] & +0.000 [+0.000,+0.000] & +0.000 [+0.000,+0.000] \\
L$\rightarrow$Q & 1 & jlens & 2 & +0.000 [+0.000,+0.000] & +0.000 [+0.000,+0.000] & +0.000 [+0.000,+0.000] \\
L$\rightarrow$Q & 1 & jlens & 4 & +0.000 [+0.000,+0.000] & +0.000 [+0.000,+0.000] & +0.000 [+0.000,+0.000] \\
L$\rightarrow$Q & 8 & jlens & 0.5 & +0.000 [+0.000,+0.000] & +0.000 [+0.000,+0.000] & +0.000 [+0.000,+0.000] \\
L$\rightarrow$Q & 8 & jlens & 2 & +0.000 [+0.000,+0.000] & +0.000 [+0.000,+0.000] & +0.000 [+0.000,+0.000] \\
L$\rightarrow$Q & 8 & jlens & 4 & -0.003 [-0.008,+0.000] & -0.003 [-0.008,+0.000] & +0.000 [+0.000,+0.000] \\
L$\rightarrow$Q & 16 & jlens & 0.5 & +0.000 [+0.000,+0.000] & +0.000 [+0.000,+0.000] & +0.000 [+0.000,+0.000] \\
L$\rightarrow$Q & 16 & jlens & 2 & +0.000 [+0.000,+0.000] & +0.000 [+0.000,+0.000] & +0.000 [+0.000,+0.000] \\
L$\rightarrow$Q & 16 & jlens & 4 & +0.000 [+0.000,+0.000] & +0.000 [+0.000,+0.000] & -0.002 [-0.005,+0.000] \\
L$\rightarrow$Q & 1 & mix pilot & 0.5 & +0.000 [+0.000,+0.000] & +0.000 [+0.000,+0.000] & +0.000 [+0.000,+0.000] \\
L$\rightarrow$Q & 1 & mix pilot & 1 & +0.000 [+0.000,+0.000] & +0.000 [+0.000,+0.000] & +0.000 [+0.000,+0.000] \\
L$\rightarrow$Q & 1 & mix pilot & 2 & +0.000 [+0.000,+0.000] & +0.000 [+0.000,+0.000] & +0.000 [+0.000,+0.000] \\
L$\rightarrow$Q & 8 & mix pilot & 0.5 & +0.000 [+0.000,+0.000] & +0.000 [+0.000,+0.000] & +0.000 [+0.000,+0.000] \\
L$\rightarrow$Q & 8 & mix pilot & 1 & +0.000 [+0.000,+0.000] & +0.000 [+0.000,+0.000] & +0.000 [+0.000,+0.000] \\
L$\rightarrow$Q & 8 & mix pilot & 2 & +0.000 [+0.000,+0.000] & +0.000 [+0.000,+0.000] & +0.000 [+0.000,+0.000] \\
Q$\rightarrow$L & 1 & jlens & 0.5 & -0.059 [-0.093,-0.024] & -0.072 [-0.107,-0.041] & -0.059 [-0.090,-0.028] \\
Q$\rightarrow$L & 1 & jlens & 2 & -0.017 [-0.052,+0.017] & -0.028 [-0.062,+0.007] & -0.024 [-0.059,+0.010] \\
Q$\rightarrow$L & 1 & jlens & 4 & +0.014 [-0.021,+0.048] & -0.038 [-0.079,+0.003] & -0.024 [-0.059,+0.010] \\
Q$\rightarrow$L & 8 & jlens & 0.5 & -0.017 [-0.036,+0.003] & -0.033 [-0.052,-0.014] & -0.025 [-0.044,-0.006] \\
Q$\rightarrow$L & 8 & jlens & 2 & +0.003 [-0.014,+0.020] & -0.006 [-0.025,+0.013] & -0.005 [-0.025,+0.016] \\
Q$\rightarrow$L & 8 & jlens & 4 & +0.000 [-0.017,+0.017] & -0.013 [-0.031,+0.006] & -0.014 [-0.034,+0.006] \\
Q$\rightarrow$L & 16 & jlens & 0.5 & -0.039 [-0.062,-0.017] & -0.033 [-0.055,-0.011] & -0.034 [-0.058,-0.013] \\
Q$\rightarrow$L & 16 & jlens & 2 & +0.002 [-0.022,+0.023] & -0.016 [-0.041,+0.008] & -0.006 [-0.030,+0.016] \\
Q$\rightarrow$L & 16 & jlens & 4 & -0.008 [-0.031,+0.014] & -0.013 [-0.036,+0.009] & -0.009 [-0.033,+0.014] \\
\bottomrule
\end{tabular}}
\caption{H1 dose panel: $\Delta$(inject $-$ none) at dose multipliers other than 1 (and every multiplier of the mix pilot).}
\label{tab:aggregation_dose}
\end{table}

\begin{table}[t]
\centering
\small
\resizebox{\textwidth}{!}{%
\begin{tabular}{llllllll}
\toprule
$N$ & dose policy & n & merged $-$ none [95\% CI] & text\_all $-$ none [95\% CI] & merged wire bytes & text bytes & zero-packet \\
\midrule
2 & fixed-norm $\times$0.5 & 298 & -0.007 [-0.030,+0.017] & +0.359 [+0.305,+0.409] & 40.9 & 364.8 & exact \\
2 & fixed-norm $\times$1 & 298 & +0.010 [-0.017,+0.037] & +0.359 [+0.305,+0.409] & 40.9 & 364.8 & exact \\
2 & fixed-norm $\times$2 & 298 & +0.023 [-0.003,+0.050] & +0.359 [+0.305,+0.409] & 40.9 & 364.8 & exact \\
2 & fixed-norm $\times$4 & 298 & -0.003 [-0.034,+0.027] & +0.359 [+0.305,+0.409] & 40.9 & 364.8 & exact \\
2 & fixed-$L_1$ & 298 & +0.023 [+0.000,+0.047] & +0.359 [+0.305,+0.409] & 40.9 & 364.8 & exact \\
4 & fixed-norm $\times$1 & 296 & +0.003 [-0.020,+0.027] & +0.291 [+0.230,+0.345] & 60.7 & 725.5 & exact \\
8 & fixed-norm $\times$0.5 & 296 & +0.017 [-0.003,+0.037] & +0.328 [+0.274,+0.378] & 90.2 & 1454.2 & exact \\
8 & fixed-norm $\times$1 & 296 & +0.010 [-0.010,+0.030] & +0.328 [+0.274,+0.378] & 90.2 & 1454.2 & exact \\
8 & fixed-norm $\times$2 & 296 & -0.020 [-0.044,+0.003] & +0.328 [+0.274,+0.378] & 90.2 & 1454.2 & exact \\
8 & fixed-norm $\times$4 & 296 & -0.027 [-0.057,+0.003] & +0.328 [+0.274,+0.378] & 90.2 & 1454.2 & exact \\
8 & fixed-$L_1$ & 296 & +0.010 [-0.010,+0.034] & +0.328 [+0.274,+0.378] & 90.2 & 1454.2 & exact \\
\bottomrule
\end{tabular}}
\caption{Fan-in (N senders merged into one packet, receiver qwen3-8b-base, J-lens dictionaries): merged $-$ none and text\_all $-$ none accuracy with cluster-bootstrap 95\% CIs per dose policy, and the merged wire bytes against the text bytes the same episodes cost. The byte half is a structural fact (bounded vs $O(N)$); the accuracy half is the fan-in boundary.}
\label{tab:fanin}
\end{table}

\begin{table}[t]
\centering
\small
\resizebox{\textwidth}{!}{%
\begin{tabular}{llllll}
\toprule
link & contrast & estimate [95\% CI] & n & entities & NI \\
\midrule
llama-3.1-8b $\rightarrow$ qwen3-8b-base & PRIMARY Delta m-permX (gap) & +2.2364 [+1.6302,+2.7742] * & 861 & 47 & -- \\
llama-3.1-8b $\rightarrow$ qwen3-8b-base & m-none (gap) & +2.1997 [+1.5911,+2.7475] * & 900 & 47 & -- \\
llama-3.1-8b $\rightarrow$ qwen3-8b-base & m-none (acc) & +0.1600 [+0.1193,+0.1977] * & 900 & 47 & -- \\
llama-3.1-8b $\rightarrow$ qwen3-8b-base & oracle-none (acc) & +0.4311 [+0.4034,+0.4600] * & 900 & 47 & -- \\
llama-3.1-8b $\rightarrow$ qwen3-8b-base & m-textpkt (gap) & -0.4326 [-0.7789,-0.0610] * & 900 & 47 & -- \\
llama-3.1-8b $\rightarrow$ qwen3-8b-base & m-textpkt (acc, NI -0.05) & -0.0233 [-0.0530,+0.0069] n.s. & 900 & 47 & MISS \\
llama-3.1-8b $\rightarrow$ qwen3-8b-base & m-shuf (gap) & +2.3047 [+1.6713,+2.8906] * & 900 & 47 & -- \\
llama-3.1-8b $\rightarrow$ qwen3-8b-base & m-cf (gap) & +4.3394 [+3.1413,+5.4041] * & 900 & 47 & -- \\
qwen3-8b-base $\rightarrow$ llama-3.1-8b & PRIMARY Delta m-permX (gap) & +0.6960 [+0.5018,+0.9159] * & 834 & 47 & -- \\
qwen3-8b-base $\rightarrow$ llama-3.1-8b & m-none (gap) & +0.6495 [+0.4822,+0.8405] * & 900 & 47 & -- \\
qwen3-8b-base $\rightarrow$ llama-3.1-8b & m-none (acc) & +0.0900 [+0.0570,+0.1213] * & 900 & 47 & -- \\
qwen3-8b-base $\rightarrow$ llama-3.1-8b & oracle-none (acc) & +0.2333 [+0.1901,+0.2760] * & 900 & 47 & -- \\
qwen3-8b-base $\rightarrow$ llama-3.1-8b & m-textpkt (gap) & -1.7941 [-2.2304,-1.2930] * & 900 & 47 & -- \\
qwen3-8b-base $\rightarrow$ llama-3.1-8b & m-textpkt (acc, NI -0.05) & -0.1067 [-0.1365,-0.0738] * & 900 & 47 & MISS \\
qwen3-8b-base $\rightarrow$ llama-3.1-8b & m-shuf (gap) & +0.6188 [+0.4466,+0.8208] * & 900 & 47 & -- \\
qwen3-8b-base $\rightarrow$ llama-3.1-8b & m-cf (gap) & +1.3272 [+0.9891,+1.7076] * & 900 & 47 & -- \\
\bottomrule
\end{tabular}}
\caption{C3$'$ on the frozen test bank (one spend): every contrast of the pinned analyze\_j3final adjudicator, recomputed with its own entity-bootstrap engine (gold-entity unit from bank provenance, nb=8000, seed 20260802, qbits 8). Non-inferiority to same-content text is adjudicated at the frozen $-0.05$ margin on the accuracy contrast.}
\label{tab:c3prime}
\end{table}

\begin{table}[t]
\centering
\small
\resizebox{\textwidth}{!}{%
\begin{tabular}{lllllll}
\toprule
bank (test) & contrast & scope & $\Delta$acc [95\% CI] & $p$ & n items & entities \\
\midrule
ac-countries & packet $-$ donor's packet & full & +0.0517 [+0.0216,+0.0823] & 0.0010 & 400 & 78 \\
ac-countries & packet $-$ donor's packet & hard slice & +0.0515 [+0.0249,+0.0813] & 0.0001 & 298 & 74 \\
ac-countries & packet $-$ random addresses & full & +0.0483 [+0.0220,+0.0747] & 0.0001 & 400 & 78 \\
ac-countries & packet $-$ random addresses & hard slice & +0.0481 [+0.0226,+0.0756] & 0.0001 & 298 & 74 \\
ac-countries & packet $-$ no comm & full & +0.0533 [+0.0052,+0.1024] & 0.0342 & 400 & 78 \\
ac-countries & packet $-$ no comm & hard slice & +0.1779 [+0.1355,+0.2226] & 0.0001 & 298 & 74 \\
ac-countries & text oracle $-$ no comm & full & +0.7400 [+0.6779,+0.8010] & 0.0001 & 400 & 78 \\
ac-countries & text oracle $-$ no comm & hard slice & +0.9933 [+0.9779,+1.0000] & 0.0001 & 298 & 74 \\
ac-tipsheets & packet $-$ donor's packet & full & -0.0025 [-0.0079,+0.0018] & 0.3806 & 400 & 80 \\
ac-tipsheets & packet $-$ donor's packet & hard slice & -0.0022 [-0.0072,+0.0000] & 0.5030 & 300 & 79 \\
ac-tipsheets & packet $-$ random addresses & full & -0.0083 [-0.0175,-0.0008] & 0.0346 & 400 & 80 \\
ac-tipsheets & packet $-$ random addresses & hard slice & -0.0056 [-0.0154,+0.0031] & 0.2710 & 300 & 79 \\
ac-tipsheets & packet $-$ no comm & full & +0.0017 [-0.0065,+0.0112] & 0.8262 & 400 & 80 \\
ac-tipsheets & packet $-$ no comm & hard slice & +0.0078 [+0.0010,+0.0183] & 0.0490 & 300 & 79 \\
ac-tipsheets & text oracle $-$ no comm & full & +0.5900 [+0.5253,+0.6556] & 0.0001 & 400 & 80 \\
ac-tipsheets & text oracle $-$ no comm & hard slice & +0.7900 [+0.7282,+0.8471] & 0.0001 & 300 & 79 \\
\bottomrule
\end{tabular}}
\caption{CONFIRM0818 CC2 on the held-out AC test banks, touched once by this gate and by no later study (sender llama-3.1-8b, receiver qwen3-32b, 3 seeds, template t0): specificity and headroom contrasts on the full set (pinned analysis) and on the hard slice (items the receiver gets wrong with no message), hierarchical entity-then-item bootstrap nb=10000, seed 20260805. Utility is not re-litigated (lock P7).}
\label{tab:cc2}
\end{table}

\begin{table}[t]
\centering
\small
\resizebox{\textwidth}{!}{%
\begin{tabular}{lllll}
\toprule
$\gamma$ & llama-3.1-8b & qwen3-8b-base & two-sender mean & zero-integrity \\
\midrule
8 & +0.0155 [-0.0386,+0.0722] & -0.0134 [-0.0861,+0.0595] & +0.0010 & yes \\
16* & +0.0473 [-0.0164,+0.1182] & -0.0023 [-0.0690,+0.0696] & +0.0225 & yes \\
24 & +0.0484 [+0.0052,+0.0953] & -0.0178 [-0.0701,+0.0395] & +0.0153 & yes \\
32 & +0.0089 [-0.0290,+0.0480] & -0.0114 [-0.0635,+0.0447] & -0.0012 & yes \\
48 & +0.0104 [-0.0210,+0.0431] & -0.0074 [-0.0383,+0.0265] & +0.0015 & yes \\
\bottomrule
\end{tabular}}
\caption{XBDOSE0902 stage 1 (dev material): cross-family address transport into glm-4-9b on the T3 relation family, swept over the frozen dose grid. Each entry is $I_{\mathrm{sender}} = (MC-MS)-(PC-PS)$ on the forced-choice logit gap with the gold-entity resampling unit (nb=8000, seed 20260802), recomputed by the pinned per-cell adjudicator from the raw rows. $\gamma^*$ (marked *) is chosen by the pre-registered rule on the two-sender mean; at $\gamma^*$ neither CI lower bound exceeds 0, so the frozen outcome is DOSE-SWEPT NULL and stage 2 was not run. The most favourable single cell (llama-3.1-8b at $\gamma$24, $p=0.0145$ unadjusted) does not survive Holm over the 10 cells ($p_{\mathrm{Holm}}=0.145$).}
\label{tab:xbdose}
\end{table}

\begin{table}[t]
\centering
\small
\resizebox{\textwidth}{!}{%
\begin{tabular}{llllll}
\toprule
link & benchmark & dim & $H$ [95\% CI] & $D_{\mathrm{cmdv}}$ [95\% CI] &  \\
\midrule
llama-3.1-8b $\rightarrow$ glm-4-9b & ac-countries & same & -0.0008 [-0.0180,+0.0164] & -0.0033 [-0.0233,+0.0167] &  \\
llama-3.1-8b $\rightarrow$ glm-4-9b & ac-tipsheets & same & +0.0000 [-0.0143,+0.0144] & -0.0058 [-0.0246,+0.0111] &  \\
llama-3.1-8b $\rightarrow$ llama-3.1-8b & ac-countries & same & +0.0000 [-0.0143,+0.0137] & -0.0100 [-0.0301,+0.0093] &  \\
llama-3.1-8b $\rightarrow$ llama-3.1-8b & ac-tipsheets & same & +0.0017 [-0.0169,+0.0199] & -0.0008 [-0.0291,+0.0250] &  \\
llama-3.1-8b $\rightarrow$ qwen3-32b & ac-countries & cross & +0.0383 [+0.0102,+0.0687]* & +0.0017 [-0.0442,+0.0480] & \textbf{primary} \\
llama-3.1-8b $\rightarrow$ qwen3-32b & ac-tipsheets & cross & -0.0108 [-0.0251,+0.0017] & +0.0167 [+0.0048,+0.0316] &  \\
llama-3.1-8b $\rightarrow$ qwen3-8b-base & ac-countries & same & -0.0067 [-0.0195,+0.0055] & +0.0008 [-0.0118,+0.0127] &  \\
llama-3.1-8b $\rightarrow$ qwen3-8b-base & ac-tipsheets & same & +0.0100 [-0.0111,+0.0319] & -0.0233 [-0.0500,+0.0017] &  \\
qwen3-8b-base $\rightarrow$ glm-4-9b & ac-countries & same & +0.0042 [-0.0182,+0.0285] & -0.0100 [-0.0255,+0.0037] &  \\
qwen3-8b-base $\rightarrow$ glm-4-9b & ac-tipsheets & same & +0.0108 [-0.0050,+0.0282] & -0.0142 [-0.0358,+0.0048] &  \\
qwen3-8b-base $\rightarrow$ llama-3.1-8b & ac-countries & same & +0.0017 [-0.0172,+0.0195] & -0.0083 [-0.0283,+0.0116] &  \\
qwen3-8b-base $\rightarrow$ llama-3.1-8b & ac-tipsheets & same & +0.0008 [-0.0182,+0.0205] & +0.0008 [-0.0249,+0.0255] &  \\
qwen3-8b-base $\rightarrow$ qwen3-32b & ac-countries & cross & +0.0433 [+0.0146,+0.0727]* & -0.0050 [-0.0513,+0.0403] &  \\
qwen3-8b-base $\rightarrow$ qwen3-32b & ac-tipsheets & cross & -0.0017 [-0.0110,+0.0087] & +0.0133 [+0.0050,+0.0239] &  \\
qwen3-8b-base $\rightarrow$ qwen3-8b-base & ac-countries & same & -0.0075 [-0.0235,+0.0078] & +0.0017 [-0.0124,+0.0170] &  \\
qwen3-8b-base $\rightarrow$ qwen3-8b-base & ac-tipsheets & same & -0.0092 [-0.0341,+0.0165] & -0.0125 [-0.0434,+0.0182] &  \\
\bottomrule
\end{tabular}}
\caption{CMDV0902: the Command-V-style trained pairwise-linear-map baseline (arm \texttt{cmdv\_graft}) at matched write pathway, band, seeded positions, receiver dose and write-norm budget. $H = $ packet\_extracted $-$ cmdv\_graft (positive = the address packet wins); $D_{\mathrm{cmdv}} = $ cmdv\_graft $-$ no\_comm. Accuracy, paired per item over 400 items and 77 gold entities, 3 seeds, hierarchical entity-clustered bootstrap nb=10000, base seed 20260805. ``dim'' is same/cross by the sender's and receiver's hidden sizes, read from the rows. Cells marked * have an $H$ interval excluding 0; they are the cross-dimension cells into qwen3-32b on ac-countries. Sign across the 16 cells: 9 positive, 7 negative.}
\label{tab:cmdv}
\end{table}

\begin{table}[t]
\centering
\small
\resizebox{\textwidth}{!}{%
\begin{tabular}{lllll}
\toprule
$\gamma$ & $I_{\mathrm{sender}}$ [95\% CI] & one-sided $p$ & CI excludes 0 & zero-write \\
\midrule
8 & +0.1199 [-0.0196,+0.2156] & 0.0430 & no & exact \\
16 & +0.1776 [+0.0247,+0.3079] & 0.0139 & yes & exact \\
24 & +0.1906 [+0.0457,+0.3298] & 0.0057 & yes & exact \\
32 & +0.1864 [+0.0298,+0.3359] & 0.0102 & yes & exact \\
48$^{*}$ & +0.2390 [+0.0781,+0.3856] & 0.0019 & yes & exact \\
\bottomrule
\end{tabular}}
\caption{OLMO0903 (development material, descriptive): address transport into olmo-3-7b, a third pretraining family, swept over the frozen dose grid with sender llama-3.1-8b. Each entry is $I_{\mathrm{sender}} = (MC-MS)-(PC-PS)$ on the forced-choice logit gap at the gold-entity resampling unit (nb=8000, seed 20260802, $n=559$ donor-matched episodes over 18 entities), recomputed by the pinned per-cell adjudicator from the raw rows. $\gamma^{*}$ (marked $^{*}$) is chosen by the rule frozen before any row; the outcome rule reads the gold-entity lower bound there and returns OPENS-A-THIRD-FAMILY. This is a development anchor: it licenses no confirmatory cross-family claim and is never pooled with Table~\ref{tab:cc1} or Table~\ref{tab:uchan}.}
\label{tab:olmo}
\end{table}

\begin{table}[t]
\centering
\small
\resizebox{\textwidth}{!}{%
\begin{tabular}{lll}
\toprule
$\gamma$ & llama-3.1-8b $I_{\mathrm{sender}}$ [95\% CI] & qwen3-8b-base $I_{\mathrm{sender}}$ [95\% CI] \\
\midrule
8 & +0.1199 [-0.0196,+0.2156] & +0.1459 [+0.0165,+0.2612]$^{\dagger}$ \\
16 & -- & +0.1605 [+0.0060,+0.3052] \\
24 & -- & +0.1567 [+0.0182,+0.3014] \\
32 & -- & +0.1620 [-0.0211,+0.3586]$^{*}$ \\
48 & +0.2390 [+0.0781,+0.3856] & +0.0621 [-0.0911,+0.2417] \\
56 & +0.2636 [+0.0955,+0.4116]$^{\dagger}$ & -- \\
64 & +0.2597 [+0.1000,+0.4023]$^{*}$ & +0.0107 [-0.1098,+0.1552]$^{\dagger}$ \\
80 & +0.1917 [+0.0545,+0.3193]$^{\dagger}$ & -- \\
96 & +0.1290 [+0.0005,+0.2299] & -- \\
\bottomrule
\end{tabular}}
\caption{OLMOEXT0903 (development material, descriptive): the two sender dose curves into olmo-3-7b, a third pretraining family, on the same attribute bank and the same receiver. $I_{\mathrm{sender}} = (MC-MS)-(PC-PS)$ on the forced-choice logit gap at the gold-entity resampling unit (nb=8000, seed 20260802; $n=559$ and $n=494$ donor-matched episodes over 18 entities), recomputed by the pinned per-cell adjudicator from the raw rows; zero-write exact in all 12 cells. $^{*}$ marks each sender's $\gamma^{*}$, the argmax over that sender's frozen doses; $^{\dagger}$ marks a band-refinement addendum cell, which the lock forbids from selecting a dose or moving a verdict. The frozen outcomes are CEILING-WAS-REAL (leg~A: the response rises past $\gamma{=}48$ and falls again, so the usable band is bracketed on both sides) and SENDER-SPECIFIC (leg~B: the second sender is positive at every dose but its argmax interval covers zero). The two $\gamma^{*}$ differ, on one receiver: dose is not a receiver constant.}
\label{tab:olmoext}
\end{table}

\begin{table}[t]
\centering
\small
\resizebox{\textwidth}{!}{%
\begin{tabular}{llllll}
\toprule
sender (bank) & $\gamma$ & $I_{\mathrm{sender}}$ [95\% CI] & one-sided $p$ & Holm & zero-write \\
\midrule
llama-3.1-8b (T3 relation) & 24 & -0.0888 [-0.2534,+0.0770] & 0.8548 & 1.000 & exact \\
llama-3.1-8b (T3 relation) & 48 & -0.0163 [-0.2029,+0.1611] & 0.5641 & 1.000 & exact \\
llama-3.1-8b (T3 relation) & 64 & -0.0450 [-0.2088,+0.1034] & 0.7078 & 1.000 & exact \\
llama-3.1-8b (T3 relation) & 96 & -0.0969 [-0.2170,+0.0199] & 0.9479 & 1.000 & exact \\
qwen3-32b (T2 attribute) & 48 & -0.1814 [-0.3133,-0.0688] & 0.9994 & -- & exact \\
qwen3-32b (T2 attribute) & 64 & -0.1550 [-0.2769,-0.0562] & 0.9995 & -- & exact \\
\bottomrule
\end{tabular}}
\caption{OLMOTASK0903 (development material, descriptive): the two things the third-family anchor does NOT do. Top block, leg~A: the relation task family into olmo-3-7b, swept over the whole dose band bracketed on attribute material ($\gamma{=}24$--$96$), so no outcome can be blamed on an untried dose --- every point estimate negative, every interval covering zero, 0 of 4 surviving Holm: RELATION-FAMILY-NULL. Bottom block, leg~B: a third sender into the same receiver on attribute material, whose intervals exclude zero on the NEGATIVE side at both doses (DOES-NOT-TRACK; one-sided $p$ tests $I_{\mathrm{sender}}>0$, so values near 1 are reliable negatives). Gold-entity resampling unit, nb=8000, seed 20260802; zero-write exact in every cell.}
\label{tab:olmotask}
\end{table}

\begin{table}[t]
\centering
\small
\resizebox{\textwidth}{!}{%
\begin{tabular}{llllll}
\toprule
sender & $\gamma$ & $MC-MS$ (matched) & $PC-PS$ (donor) & $I_{\mathrm{sender}}$ & gold-in-support \\
\midrule
llama-3.1-8b & 48 & +0.2231 & -0.0159 & +0.2390 & 0.162 \\
llama-3.1-8b & 64 & +0.2129 & -0.0469 & +0.2597 & 0.162 \\
qwen3-8b-base & 48 & +0.1629 & +0.1009 & +0.0621 & 0.108 \\
qwen3-32b & 48 & -0.2276 & -0.0462 & -0.1814 & 0.218 \\
qwen3-32b & 64 & -0.1817 & -0.0267 & -0.1550 & 0.218 \\
\bottomrule
\end{tabular}}
\caption{The CPU decomposition behind the sender spread on olmo-3-7b (attribute bank, three senders). $I_{\mathrm{sender}} = (MC-MS)-(PC-PS)$, computed on the episodes carrying all four arms so the two legs add up exactly. The reversal of the third sender is carried by the MATCHED leg, not by a broken donor control; and \texttt{frac\_gold\_word\_in\_support} --- a bank/sender property logged before any outcome existed --- is HIGHER for the negative sender than for the positive one, which is what refutes the extraction-quality explanation. No mechanism is proposed: the reversal is reported as unexplained.}
\label{tab:sender_decomp}
\end{table}

\begin{table}[t]
\centering
\small
\resizebox{\textwidth}{!}{%
\begin{tabular}{llllll}
\toprule
leg & link & $\gamma$ & $I_{\mathrm{sender}}$ [95\% CI] & one-sided $p$ & zero-write \\
\midrule
leg A (cross-family) & qwen3-32b $\rightarrow$ glm-4-9b & 8 & +0.1632 [-0.0113,+0.3654] & 0.0337 & exact \\
leg A (cross-family) & qwen3-32b $\rightarrow$ glm-4-9b & 16 & +0.2006 [-0.0291,+0.4698] & 0.0416 & exact \\
leg A (same-family, never pooled) & qwen3-32b $\rightarrow$ qwen3-8b-base & 16 & +0.1373 [-0.0813,+0.3558] & 0.1109 & exact \\
leg A (same-family, never pooled) & qwen3-32b $\rightarrow$ qwen3-8b-base & 32 & +0.0292 [-0.0680,+0.1283] & 0.2761 & exact \\
leg B (incumbent receiver) & qwen3-8b-base $\rightarrow$ glm-4-9b & 8$^{*}$ & +0.3660 [+0.1936,+0.5272] & 0.0001 & exact \\
leg B (incumbent receiver) & qwen3-8b-base $\rightarrow$ glm-4-9b & 16$^{\ddagger}$ & +0.3400 [+0.1886,+0.5696] & 0.0001 & exact \\
leg B (incumbent receiver) & qwen3-8b-base $\rightarrow$ glm-4-9b & 32 & +0.2309 [+0.1357,+0.3911] & 0.0001 & exact \\
leg B (incumbent receiver) & qwen3-8b-base $\rightarrow$ glm-4-9b & 48 & +0.1871 [+0.0982,+0.3068] & 0.0001 & exact \\
\bottomrule
\end{tabular}}
\caption{SENDERDIAG0904 (development, diagnostic). Leg~A asks whether the significant negative of Table~\ref{tab:olmotask} follows the SENDER: qwen3-32b is positive-leaning into glm-4-9b at both doses, so the reversal did not replicate on a second receiver and is bounded as pair-specific (MIXED under the frozen rule, which requires a positively significant interval for PAIR-PROPERTY). The same-family half is labelled and never pooled with it. Leg~B sweeps a SECOND sender over an incumbent receiver whose dose was adjudicated with a different one: all 4 of 4 intervals exclude zero, the curve decreases monotonically, and the argmax is $\gamma{=}8$ ($^{*}$) rather than the adjudicated $\gamma{=}16$ ($^{\ddagger}$, selected with llama-3.1-8b) --- CONVENTION-FAILS. Every published $\gamma^{*}$ is therefore conditional on the sender it was chosen with. Gold-entity resampling unit, nb=8000, seed 20260802; zero-write exact in all 8 cells.}
\label{tab:senderdiag}
\end{table}

\begin{table}[t]
\centering
\small
\resizebox{\textwidth}{!}{%
\begin{tabular}{lll}
\toprule
contrast & corrected fit & CMDV0902 (pre-fix fit) \\
\midrule
H & +0.0342 [+0.0055,+0.0646] & +0.0383 [+0.0102,+0.0687] \\
D\_cmdv & +0.0058 [-0.0410,+0.0518] & +0.0017 [-0.0442,+0.0480] \\
cmdv\_vs\_acgraft & +0.0092 [-0.0118,+0.0307] & +0.0000 [-0.0192,+0.0185] \\
\bottomrule
\end{tabular}}
\caption{The fit-protocol control. CMDV0902's primary cell (llama-3.1-8b / qwen3-32b / ac-countries) re-run in a separate output directory with the corrected disjoint-split stitch (fitted map \texttt{affine\_stitch:43bdd441fe9346ec} against CMDV0902's \texttt{affine\_stitch:751edf9aad782a12}; the builder aborts if the two are equal). $H = \texttt{packet\_extracted} - \texttt{cmdv\_graft}$ is the head-to-head. Both columns are recomputed from their own raw rows with the campaign's estimator (400 paired items, 77 gold entities, hierarchical entity-clustered bootstrap nb=10000, base seed 20260805); the right column additionally reproduces CMDV0902's adjudication JSON exactly. Each $H$ falls inside the other protocol's interval, so the corrected fit does not move H, and the asymmetric interpretation declared before the cross-dimension extension can be lifted. CMDV0902's verdict is unchanged.}
\label{tab:fitctl}
\end{table}

\begin{table}[t]
\centering
\small
\resizebox{\textwidth}{!}{%
\begin{tabular}{lllllll}
\toprule
link & $\gamma$ & $I_{\mathrm{sender}}$ [95\% CI] & one-sided $p$ & $n$ & $G$ & zero-write \\
\midrule
llama-3.1-8b $\rightarrow$ glm-4-9b & 8 & +0.3962 [+0.1364,+0.6357] & 0.0021 & 495 & 11 & exact \\
llama-3.1-8b $\rightarrow$ glm-4-9b & 16$^{*}$ & +0.4986 [+0.1538,+0.8153] & 0.0020 & 495 & 11 & exact \\
qwen3-8b-base $\rightarrow$ glm-4-9b & 8$^{*}$ & +0.5856 [+0.3881,+0.8115] & 0.0001 & 433 & 11 & exact \\
qwen3-8b-base $\rightarrow$ glm-4-9b & 16 & +0.5642 [+0.3149,+0.8595] & 0.0001 & 433 & 11 & exact \\
\bottomrule
\end{tabular}}
\caption{XFAMENT0905 (descriptive only; calibration material). The XFAM0819 senders into glm-4-9b on a further T2 bank carrying $G{=}11$ gold entities, minted from the calibration split because the T2 development universe is exhausted. The split manifest reserves that split for selection, so this table cannot upgrade the five-entity replication, enters no gate family and is never pooled with it; it reports only that the same contrast is present at each sender's better dose ($^{*}$; the frozen rule, STABLE), with 4 of 4 intervals excluding zero. The two senders' better doses differ. $n$ is the donor-matched denominator; gold-entity bootstrap, $n_b{=}8000$.}
\label{tab:xfament}
\end{table}

\begin{table}[t]
\centering
\small
\resizebox{\textwidth}{!}{%
\begin{tabular}{llllll}
\toprule
cell & gold in own slot & $I_{\mathrm{sender}}$ all ($n$) & hit episodes ($n$) & non-hit episodes ($n$) & top concept word (share) \\
\midrule
llama-3.1-8b $\rightarrow$ qwen3-14b-base & 3/592 (0.5\%) & +0.566 (551) & +0.542 (3) & +0.566 (548) & scarf (36.5\%) \\
llama-3.1-8b $\rightarrow$ qwen3-30b-a3b & 3/592 (0.5\%) & +1.443 (551) & +3.458 (3) & +1.432 (548) & scarf (36.5\%) \\
llama-3.1-8b $\rightarrow$ qwen3-32b & 3/592 (0.5\%) & +1.433 (551) & +3.354 (3) & +1.423 (548) & scarf (36.5\%) \\
llama-3.1-8b $\rightarrow$ qwen3.5-27b & 3/592 (0.5\%) & +0.827 (551) & +1.208 (3) & +0.824 (548) & scarf (36.5\%) \\
llama-3.1-8b $\rightarrow$ qwen3.5-397b & 1/296 (0.3\%) & +1.034 (261) & -0.438 (1) & +1.040 (260) & scarf (36.5\%) \\
qwen3-8b-base $\rightarrow$ qwen3.5-27b & 43/592 (7.3\%) & +0.541 (488) & +0.423 (21) & +0.546 (467) & block (23.3\%) \\
qwen3-8b-base $\rightarrow$ qwen3.5-397b & 18/296 (6.1\%) & +1.667 (237) & -0.494 (10) & +1.762 (227) & block (23.0\%) \\
\bottomrule
\end{tabular}}
\caption{What the sender's packet names (CC1 cells, MC arm). Second column: MC packets whose support at the gold alias's slot contains the gold entity's own concept word. Third to fifth: $I_{\mathrm{sender}}$ over all four-arm episodes (the adjudicated estimate), over the episodes where the packet names the gold entity, and over those where it does not (plain episode means; descriptive, no interval). Last column: the sender's most-used concept word and the share of all lane occurrences it carries. The effect is carried by the non-hit episodes; the table discloses this and interprets nothing.}
\label{tab:packet_contents}
\end{table}

\begin{table}[t]
\centering
\small
\resizebox{\textwidth}{!}{%
\begin{tabular}{llllll}
\toprule
cell & $G$ & LOO min mean (dropped) & LOO worst CI-lo (dropped) & leave-two-out worst CI-lo & every interval $>0$ \\
\midrule
llama-3.1-8b $\rightarrow$ qwen3-14b-base & 14 & +0.507 (belt) & +0.306 (belt) & +0.262 & yes \\
llama-3.1-8b $\rightarrow$ qwen3-30b-a3b & 14 & +1.173 (belt) & +0.506 (belt) & +0.376 & yes \\
llama-3.1-8b $\rightarrow$ qwen3-32b & 14 & +1.032 (fox) & +0.300 (fox) & +0.134 & yes \\
llama-3.1-8b $\rightarrow$ qwen3.5-27b & 14 & +0.692 (belt) & +0.317 (belt) & +0.247 & yes \\
llama-3.1-8b $\rightarrow$ qwen3.5-397b & 14 & +0.817 (belt) & +0.281 (belt) & +0.164 & yes \\
qwen3-8b-base $\rightarrow$ qwen3.5-27b & 14 & +0.476 (pear) & +0.295 (cocoa) & +0.252 & yes \\
qwen3-8b-base $\rightarrow$ qwen3.5-397b & 14 & +1.379 (pear) & +0.562 (pear) & +0.336 & yes \\
llama-3.1-8b $\rightarrow$ glm-4-9b (XFAM0819) & 5 & +0.244 (shoe) & -0.078 (shoe) & -- & NO \\
\bottomrule
\end{tabular}}
\caption{Leave-gold-entities-out sensitivity of $I_{\mathrm{sender}}$ (gold-entity bootstrap, $n_b{=}8000$, the primary interval's draws). Each CC1 cell is re-estimated with every one, and every pair, of its $G$ gold entities removed; the columns show the smallest resulting mean, the smallest lower 95\% bound, and whether every interval stays above zero. The XFAM0819 cross-family primary cell is shown on the last row: removing one entity gives an interval that covers zero, so that closure is conditional on one entity. The gold-entity unit does not resample foil or donor entities.}
\label{tab:cc1_loo}
\end{table}

\begin{table}[t]
\centering
\small
\resizebox{\textwidth}{!}{%
\begin{tabular}{llllll}
\toprule
cell & $n$ & $|\Delta q|\le26$ & $\|\delta\|$ within 5\% & median $|\Delta q|$ & max rel.\ norm gap \\
\midrule
llama-3.1-8b $\rightarrow$ qwen3-14b-base & 551 & 167/551 (30\%) & 519/551 (94\%) & 48 & 0.129 \\
llama-3.1-8b $\rightarrow$ qwen3-30b-a3b & 551 & 168/551 (30\%) & 518/551 (94\%) & 47 & 0.124 \\
llama-3.1-8b $\rightarrow$ qwen3-32b & 551 & 167/551 (30\%) & 508/551 (92\%) & 47 & 0.128 \\
llama-3.1-8b $\rightarrow$ qwen3.5-27b & 551 & 168/551 (30\%) & 506/551 (92\%) & 47 & 0.117 \\
llama-3.1-8b $\rightarrow$ qwen3.5-397b & 261 & 64/261 (25\%) & 249/261 (95\%) & 63 & 0.104 \\
qwen3-8b-base $\rightarrow$ qwen3.5-27b & 488 & 56/488 (11\%) & 372/488 (76\%) & 45 & 0.137 \\
qwen3-8b-base $\rightarrow$ qwen3.5-397b & 237 & 19/237 (8\%) & 234/237 (99\%) & 50 & 0.068 \\
\bottomrule
\end{tabular}}
\caption{Donor matching as executed (CC1 cells, PC arm; $n$ = donor-matched episodes). The hard admissibility clauses (support must not name the recipient's entities, lane sizes, type histogram, byte count, entity disjointness) hold on every row; the coefficient clause (max sorted-$q$ gap $\le26$ of 255) and the write-norm clause (relative gap $\le5$\%) are soft costs minimised by the assignment with an argmin fallback, pre-declared as such, and this table reports how often the realized donor sits inside each tolerance. The write norm is estimated on an un-injected forward and is exact only at the lowest band layer.}
\label{tab:donor_match}
\end{table}
\newcommand{\XSNULLJACCGAPMAX}{+0.032}   %
\newcommand{\XSNULLZEROFRACMIN}{0.57}   %

\begin{table}[t]
\centering
\small
\resizebox{\textwidth}{!}{%
\begin{tabular}{llllllll}
\toprule
pair & lineage & Jaccard median & observed mean & null mean & gap & null zero-fraction & exact-set (obs.) \\
\midrule
llama-3.1-8b vs llama-3.1-8b-instruct & same-base & 0.33 & 0.310 & 0.139 & +0.171 & 0.37 & 0.003 \\
llama-3.1-70b vs llama-3.1-8b & same-family & 0.17 & 0.213 & 0.114 & +0.099 & 0.45 & 0.000 \\
llama-3.1-70b vs llama-3.1-8b-instruct & same-family & 0.33 & 0.246 & 0.117 & +0.129 & 0.42 & 0.000 \\
qwen3-14b-base vs qwen3-30b-a3b & same-family & 0.00 & 0.076 & 0.069 & +0.008 & 0.57 & 0.000 \\
qwen3-14b-base vs qwen3-32b & same-family & 0.00 & 0.005 & 0.002 & +0.003 & 0.99 & 0.000 \\
qwen3-14b-base vs qwen3-8b-base & same-family & 0.00 & 0.015 & 0.006 & +0.009 & 0.96 & 0.000 \\
qwen3-14b-base vs r1-qwen-14b & same-family & 0.00 & 0.003 & 0.001 & +0.002 & 0.99 & 0.000 \\
qwen3-30b-a3b vs qwen3-32b & same-family & 0.00 & 0.008 & 0.005 & +0.003 & 0.96 & 0.000 \\
qwen3-30b-a3b vs qwen3-8b-base & same-family & 0.14 & 0.208 & 0.156 & +0.052 & 0.17 & 0.000 \\
qwen3-30b-a3b vs r1-qwen-14b & same-family & 0.00 & 0.047 & 0.030 & +0.017 & 0.79 & 0.000 \\
qwen3-32b vs qwen3-8b-base & same-family & 0.00 & 0.046 & 0.032 & +0.014 & 0.79 & 0.000 \\
qwen3-32b vs r1-qwen-14b & same-family & 0.00 & 0.036 & 0.022 & +0.014 & 0.85 & 0.000 \\
qwen3-8b-base vs r1-qwen-14b & same-family & 0.14 & 0.115 & 0.083 & +0.032 & 0.45 & 0.000 \\
llama-3.1-70b vs qwen3-14b-base & cross & 0.00 & 0.039 & 0.027 & +0.012 & 0.82 & 0.000 \\
llama-3.1-70b vs qwen3-30b-a3b & cross & 0.00 & 0.027 & 0.009 & +0.018 & 0.93 & 0.000 \\
llama-3.1-70b vs qwen3-32b & cross & 0.14 & 0.100 & 0.071 & +0.029 & 0.60 & 0.000 \\
llama-3.1-70b vs qwen3-8b-base & cross & 0.00 & 0.047 & 0.026 & +0.021 & 0.82 & 0.000 \\
llama-3.1-70b vs r1-qwen-14b & cross & 0.00 & 0.065 & 0.033 & +0.032 & 0.79 & 0.000 \\
llama-3.1-8b-instruct vs qwen3-14b-base & cross & 0.00 & 0.045 & 0.028 & +0.018 & 0.82 & 0.000 \\
llama-3.1-8b-instruct vs qwen3-30b-a3b & cross & 0.00 & 0.029 & 0.018 & +0.011 & 0.88 & 0.000 \\
llama-3.1-8b-instruct vs qwen3-32b & cross & 0.14 & 0.104 & 0.085 & +0.019 & 0.44 & 0.000 \\
llama-3.1-8b-instruct vs qwen3-8b-base & cross & 0.00 & 0.018 & 0.011 & +0.007 & 0.92 & 0.000 \\
llama-3.1-8b-instruct vs r1-qwen-14b & cross & 0.00 & 0.057 & 0.032 & +0.025 & 0.80 & 0.000 \\
llama-3.1-8b vs qwen3-14b-base & cross & 0.00 & 0.067 & 0.058 & +0.009 & 0.66 & 0.000 \\
llama-3.1-8b vs qwen3-30b-a3b & cross & 0.00 & 0.029 & 0.011 & +0.018 & 0.93 & 0.000 \\
llama-3.1-8b vs qwen3-32b & cross & 0.00 & 0.046 & 0.028 & +0.018 & 0.83 & 0.000 \\
llama-3.1-8b vs qwen3-8b-base & cross & 0.00 & 0.045 & 0.028 & +0.017 & 0.82 & 0.000 \\
llama-3.1-8b vs r1-qwen-14b & cross & 0.00 & 0.059 & 0.038 & +0.020 & 0.77 & 0.000 \\
\bottomrule
\end{tabular}}
\caption{Leg A against its own null (stage-2 primary pairs). Per-lane support-Jaccard between two senders' packets for the same episode (observed) and under an episode-permuted null in which the second sender's packet is drawn from a different episode with packet sizes, duplication and per-sender attractor concentration preserved. On the pairs whose observed median is zero the observed mean exceeds the null by at most \XSNULLJACCGAPMAX{} and the null itself already puts at least \XSNULLZEROFRACMIN{} of lanes at zero overlap: support disjointness across lineages is what the null predicts. The pair files store no exact-set statistic for the null; the observed exact-set agreement is shown alone.}
\label{tab:xsend_null}
\end{table}

\begin{table}[t]
\centering
\small
\resizebox{\textwidth}{!}{%
\begin{tabular}{lllllllllll}
\toprule
cell & $n$ & $n$ (donor) & none & MC & MS & PC & PS & text packet & text oracle & zero packet \\
\midrule
llama-3.1-8b $\rightarrow$ qwen3-14b-base & 592 & 551 & 48.6 & 58.6 & 36.7 & 49.2 & 42.8 & 62.7 & 100.0 & 48.6 \\
llama-3.1-8b $\rightarrow$ qwen3-30b-a3b & 592 & 551 & 49.2 & 57.6 & 41.9 & 51.0 & 47.4 & 67.1 & 100.0 & 49.2 \\
llama-3.1-8b $\rightarrow$ qwen3-32b & 592 & 551 & 49.0 & 60.3 & 37.0 & 52.6 & 44.8 & 64.9 & 99.8 & 49.0 \\
llama-3.1-8b $\rightarrow$ qwen3.5-27b & 592 & 551 & 48.3 & 62.3 & 34.5 & 53.2 & 46.3 & 71.5 & 100.0 & 48.3 \\
llama-3.1-8b $\rightarrow$ qwen3.5-397b & 296 & 261 & 44.9 & 61.5 & 35.8 & 52.9 & 51.0 & 69.6 & 100.0 & 44.9 \\
qwen3-8b-base $\rightarrow$ qwen3.5-27b & 592 & 488 & 48.3 & 56.2 & 39.7 & 46.1 & 49.2 & 76.5 & 100.0 & 48.3 \\
qwen3-8b-base $\rightarrow$ qwen3.5-397b & 296 & 237 & 44.9 & 66.9 & 30.1 & 49.8 & 43.9 & 75.0 & 100.0 & 44.9 \\
\bottomrule
\end{tabular}}
\caption{Forced-choice accuracy (percent correct on the first-wordpiece alias logit gap) per arm on the CC1 cells, beside the logit-gap contrasts of the main table. $n$ is the episode count for the sender-packet arms; the donor arms (PC, PS) run on the donor-matched subset. The zero-packet arm reproduces the no-write arm's correctness on every episode (integrity, asserted at build time).}
\label{tab:cc1_acc}
\end{table}

\begin{table}[t]
\centering
\small
\resizebox{\textwidth}{!}{%
\begin{tabular}{llllllll}
\toprule
channel & base (qwen2.5-14b) & post (R1 distill) & instruct sibling & $R_B'$ [95\% CI] & class & $D$ [95\% CI] & $R_B'-R_A$ [cons.\ CI] \\
\midrule
unembed & +0.321 & +0.402 & +0.365 & 1.253 [1.191,1.329] & \textsc{preserved} & 1.137 [1.098,1.180] & 1.159 [1.072,1.260] (vs $R_A{=}0.094$) \\
probe & +0.320 & +0.261 & +0.332 & 0.814 [0.738,0.889] & \textsc{preserved} & 1.038 [0.989,1.089] & 0.788 [0.694,0.876] (vs $R_A{=}0.026$) \\
\bottomrule
\end{tabular}}
\caption{POSTTRAINBASE0905: lineage B against its \emph{true} pretrained base. Mean transfer from the three outside senders into the pretrained qwen2.5-14b (new cells), its R1 distill and its Instruct sibling (the signed P2 / P2PROBE0902 cells, reused and never re-run) on both instruments; $R_B'{=}$post/base, $D{=}$instruct/base, concept-clustered bootstrap over 64 development concepts ($n_b{=}8000$). Classes on the imported bars (0.80 preserved / 0.50 broken); the differential's interval is the conservative combination with $R_A$'s signed interval. Verdict \emph{TRUE-BASE-PRESERVED(unembed,probe)}: on the probe channel the point sits at the bar and the interval covers it, reported as such. Development tier, two instruments; nothing here says why the Qwen lineage survives, and $R_B'$ is never pooled with the signed Instruct-sibling $R_B$.}
\label{tab:posttrainbase}
\end{table}
\begin{table}[t]
\centering
\small
\resizebox{\textwidth}{!}{%
\begin{tabular}{lllllllll}
\toprule
cell & episodes & four-arm $n$ & $G$ & $I_{sender}$ & gold-entity 95\% CI & $p$ (Holm) & verdict & worst LOO CI-lower (entity) \\
\midrule
llama-3.1-8b $\rightarrow$ qwen3-14b-base & 596 & 536 & 42 & +1.345 & [+0.995,+1.739] & 0.000875 & closes & +0.929 (dolphin) \\
llama-3.1-8b $\rightarrow$ qwen3-30b-a3b & 596 & 536 & 42 & +4.477 & [+3.240,+5.805] & 0.000875 & closes & +3.035 (mango) \\
llama-3.1-8b $\rightarrow$ qwen3-32b & 596 & 536 & 42 & +4.212 & [+3.000,+5.502] & 0.000875 & closes & +2.818 (dolphin) \\
llama-3.1-8b $\rightarrow$ qwen3.5-27b & 596 & 536 & 42 & +2.504 & [+1.648,+3.398] & 0.000875 & closes & +1.524 (dolphin) \\
llama-3.1-8b $\rightarrow$ qwen3.5-397b & 596 & 536 & 42 & +3.674 & [+2.625,+4.751] & 0.000875 & closes & +2.472 (dolphin) \\
qwen3-8b-base $\rightarrow$ qwen3.5-27b & 596 & 548 & 42 & +0.777 & [+0.462,+1.141] & 0.000875 & closes & +0.413 (potato) \\
qwen3-8b-base $\rightarrow$ qwen3.5-397b & 596 & 548 & 42 & +1.867 & [+1.220,+2.604] & 0.000875 & closes & +1.087 (mango) \\
llama-3.1-8b $\rightarrow$ glm-4-9b & 596 & 536 & 42 & +1.168 & [+0.776,+1.583] & 0.000250 & closes & +0.701 (mango) \\
qwen3-8b-base $\rightarrow$ glm-4-9b & 596 & 548 & 42 & +0.746 & [+0.470,+1.089] & 0.000250 & closes & +0.413 (mango) \\
\bottomrule
\end{tabular}}
\caption{ENTDISJ0908: the pre-registered confirmatory roster re-run on ENTITY-DISJOINT material. Upper block: the 7 CC1-roster cells under Holm at $\alpha=0.05$ with $m=7$ fixed at freeze. Lower block: the two primary-dose cross-family cells, adjudicated as their own family (each alone with no correction; Holm $m=2$ only for the joint two-sender sentence). The resampling unit is the gold entity, 42 per cell against CC1's 14. `episodes' is the loaded count; the contrast and its interval are computed on the four-arm subset, and the effective unit behind the interval is $G$, not either episode count. Zero-write reproduces no-communication exactly in every cell. Every value here is reproduced by an independent plain-python recompute that shares no code with the adjudicator.}
\label{tab:entdisj}
\end{table}

\begin{table}[t]
\centering
\small
\resizebox{\textwidth}{!}{%
\begin{tabular}{lllll}
\toprule
write dose & injected coverage & free $I_{sender}$ & 95\% CI & forced $I_{sender}$ \\
\midrule
$\gamma$16 (adjudicated) & 0.058 & not evaluable & -- & +0.566 \\
$\gamma$8 & 0.711 & +0.111 & [+0.058,+0.161] & +0.321 \\
$\gamma$4 & 0.885 & +0.078 & [+0.028,+0.134] & +0.182 \\
\bottomrule
\end{tabular}}
\caption{FREEGEN0906, DEVELOPMENT TIER, one sender into one receiver on one development bank (llama-3.1-8b $\rightarrow$ qwen3-14b-base). Lowering the write restores free-generation format compliance while the forced-choice effect falls: the write trades compliance against effect size, and there is a regime where both are non-trivial. This selects nothing --- the confirmatory cells keep $\gamma$16 --- and it is never pooled with the confirmatory tallies.}
\label{tab:freegen}
\end{table}

\begin{table}[t]
\centering
\small
\resizebox{\textwidth}{!}{%
\begin{tabular}{lllllll}
\toprule
receiver & dose & injected coverage & free $I_{sender}$ & 95\% CI & forced $I_{sender}$ & CI excludes 0 \\
\midrule
qwen3-30b-a3b & $\gamma^{*}/2$ & 0.169 & +0.0436 & [+0.0100,+0.0809] & +0.5719 & yes \\
qwen3-30b-a3b & $\gamma^{*}/4$ & 0.211 & +0.0018 & [-0.0129,+0.0207] & +0.2754 & no \\
qwen3-32b & $\gamma^{*}/2$ & 0.479 & +0.0871 & [+0.0192,+0.1574] & +1.0375 & yes \\
qwen3-32b & $\gamma^{*}/4$ & 0.454 & +0.0254 & [-0.0017,+0.0519] & +0.3033 & no \\
qwen3.5-27b & $\gamma^{*}/2$ & 0.063 & +0.0363 & [+0.0076,+0.0695] & +0.4974 & yes \\
qwen3.5-27b & $\gamma^{*}/2$ & 0.040 & +0.0328 & [+0.0068,+0.0568] & +0.2590 & yes \\
qwen3.5-27b & $\gamma^{*}/4$ & 0.014 & +0.0054 & [+0.0000,+0.0130] & +0.2002 & no \\
qwen3.5-27b & $\gamma^{*}/4$ & 0.022 & +0.0000 & -- & +0.0831 & n/a \\
\bottomrule
\end{tabular}}
\caption{Widened free-generation dose panel, DEVELOPMENT TIER and descriptive. Each receiver is run at half and a quarter of its OWN adjudicated dose, on the already-adjudicated confirmatory rows with their frozen packets and donor assignment; only the write dose changes. The grid and the commitment to report every cell were frozen before any cell ran. The panel selects nothing --- every confirmatory cell keeps its adjudicated dose --- and is pooled with nothing.}
\label{tab:fgpanel}
\end{table}

\begin{table}[t]
\centering
\small
\resizebox{\textwidth}{!}{%
\begin{tabular}{llll}
\toprule
cell & $I_{sender}$ on gap\_seq & gold-entity 95\% CI & $p$ \\
\midrule
llama-3.1-8b $\rightarrow$ qwen3-14b-base & +0.571 & [+0.356,+0.795] & 0.000125 \\
llama-3.1-8b $\rightarrow$ qwen3-30b-a3b & +1.443 & [+0.666,+2.340] & 0.000125 \\
llama-3.1-8b $\rightarrow$ qwen3-32b & +1.479 & [+0.514,+2.582] & 0.001125 \\
llama-3.1-8b $\rightarrow$ qwen3.5-27b & +0.814 & [+0.398,+1.271] & 0.000125 \\
qwen3-8b-base $\rightarrow$ qwen3.5-27b & +0.540 & [+0.348,+0.707] & 0.000125 \\
llama-3.1-8b -\textgreater{} qwen3.5-397b & VOID & replay 4/50 & 1.0 \\
qwen3-8b-base -\textgreater{} qwen3.5-397b & VOID & replay 5/223 & 1.0 \\
\bottomrule
\end{tabular}}
\caption{RESCORE\_SEQ0906: the confirmatory rows re-scored on the full alias sequence. 5 of 7 cells are re-scored and close; 2 are VOID on the pre-declared replay guard and enter Holm at $p=1$. The void cells are the largest receiver, so the sequence endpoint is unverified there.}
\label{tab:rescoreseq}
\end{table}

\begin{table}[t]
\centering
\small
\resizebox{\textwidth}{!}{%
\begin{tabular}{lllll}
\toprule
contrast & Llama-3.1-8B & 95\% CI & Qwen3-8B-Base & 95\% CI \\
\midrule
write the gold entity's coordinate & +0.682 & [+0.544,+0.831] & +1.211 & [+0.804,+1.664] \\
gold coordinate $-$ random entity's & +0.649 & [+0.521,+0.784] & +1.166 & [+0.879,+1.466] \\
write the \textbf{foil}'s coordinate & -0.759 & [-0.891,-0.622] & -1.253 & [-1.627,-0.837] \\
random entity's coordinate (null) & +0.033 & [-0.041,+0.106] & +0.045 & [-0.174,+0.284] \\
\textbf{extracted} coefficients & +0.245 & [+0.172,+0.323] & +0.236 & [+0.077,+0.423] \\
\bottomrule
\end{tabular}}
\caption{Instrument study, self links, forced-choice gap, entity-unit bootstrap. \textbf{Recomputed over all 200 clusters}: the gate's own numbers are on a frozen 100-cluster held-out half whose split is unrecoverable, so these are a full-sample recomputation and not the adjudicated values. Writing the gold entity's coordinate and writing the foil's are near mirror images, while a random entity's coordinate moves nothing --- which is what separates a content-specific edit from a non-specific push. The extracted-coefficient row is the one any end-to-end claim must carry, and it is several times smaller than the oracle rows.}
\label{tab:j0cr}
\end{table}

\begin{table}[t]
\centering
\small
\resizebox{\textwidth}{!}{%
\begin{tabular}{llllllll}
\toprule
bank & split & clusters & opposite dir. & same dir. & \textbf{both} & pairs & entities \\
\midrule
t2\_dev400 & dev & 400 & 76 & 95 & 12 & 9 & 2 \\
t2\_test & test & 300 & 127 & 46 & 10 & 5 & 1 \\
t2\_confirm & dev & 200 & 37 & 61 & 9 & 7 & 2 \\
t2\_xfament0905 & calibration & 284 & 58 & 76 & 8 & 4 & 2 \\
t2\_confirm0818 & dev & 300 & 42 & 36 & 4 & 2 & 2 \\
t2\_construction & construction & 20 & 10 & 6 & 3 & 2 & 2 \\
t2\_dev & dev & 200 & 39 & 34 & 3 & 2 & 1 \\
t2\_entdisj0908\_test & test & 300 & 53 & 64 & 0 & 0 & 0 \\
t2\_xfam0819 & dev & 107 & 12 & 0 & 0 & 0 & 0 \\
\bottomrule
\end{tabular}}
\caption{Every $t2$ bank, counted against the \emph{enumerable} query universe rather than the raw attribute table. A cluster can support a question-following test only if its entity pair carries both an opposite-direction second attribute (the answer flips) and a same-direction one (the answer does not) as legal queries in that bank's split. The `both' column is what such a study could actually use.}
\label{tab:qfollowmat}
\end{table}

\begin{table}[t]
\centering
\small
\resizebox{\textwidth}{!}{%
\begin{tabular}{lllllll}
\toprule
direction & budget $B$ & arm & $\Delta$acc vs none [95\% CI] & mean bytes billed & n & entities \\
\midrule
L$\rightarrow$Q & 12 & text\_budget & +0.022 [+0.006,+0.040] & 9.8 & 900 & 49 \\
L$\rightarrow$Q & 12 & text\_codebook & +0.087 [+0.061,+0.115] & 7.6 & 900 & 49 \\
L$\rightarrow$Q & 12 & text\_gen & +0.369 [+0.333,+0.408] & 8.3 & 900 & 49 \\
L$\rightarrow$Q & 16 & text\_budget & +0.038 [+0.017,+0.060] & 12.2 & 900 & 49 \\
L$\rightarrow$Q & 16 & text\_codebook & +0.130 [+0.092,+0.172] & 11.4 & 900 & 49 \\
L$\rightarrow$Q & 16 & text\_gen & +0.377 [+0.342,+0.416] & 9.6 & 900 & 49 \\
L$\rightarrow$Q & 22 & text\_budget & +0.046 [+0.011,+0.078] & 19.8 & 900 & 49 \\
L$\rightarrow$Q & 22 & text\_codebook & +0.158 [+0.126,+0.192] & 17.6 & 900 & 49 \\
L$\rightarrow$Q & 22 & text\_gen & +0.349 [+0.315,+0.386] & 19.1 & 900 & 49 \\
L$\rightarrow$Q & 32 & text\_budget & +0.116 [+0.079,+0.153] & 29.3 & 900 & 49 \\
L$\rightarrow$Q & 32 & text\_codebook & +0.148 [+0.111,+0.189] & 27.7 & 900 & 49 \\
L$\rightarrow$Q & 32 & text\_gen & +0.378 [+0.343,+0.416] & 28.2 & 900 & 49 \\
L$\rightarrow$Q & 48 & text\_budget & +0.250 [+0.213,+0.292] & 45.3 & 900 & 49 \\
L$\rightarrow$Q & 48 & text\_codebook & +0.148 [+0.113,+0.187] & 29.7 & 900 & 49 \\
L$\rightarrow$Q & 48 & text\_gen & +0.380 [+0.340,+0.423] & 43.4 & 900 & 49 \\
L$\rightarrow$Q & 64 & text\_budget & +0.318 [+0.276,+0.360] & 59.3 & 900 & 49 \\
L$\rightarrow$Q & 64 & text\_codebook & +0.148 [+0.113,+0.187] & 29.7 & 900 & 49 \\
L$\rightarrow$Q & 64 & text\_gen & +0.374 [+0.338,+0.414] & 59.3 & 900 & 49 \\
L$\rightarrow$Q & -- & inject\_matched (the packet) & +0.117 [+0.084,+0.153] & 18.0 & 900 & 49 \\
L$\rightarrow$Q & -- & inject\_oracle (reference) & +0.423 [+0.395,+0.456] & 12.0 & 900 & 49 \\
L$\rightarrow$Q & -- & text\_oracle (reference) & +0.450 [+0.422,+0.482] & -- & 900 & 49 \\
Q$\rightarrow$L & 12 & text\_budget & +0.038 [+0.014,+0.062] & 9.8 & 900 & 49 \\
Q$\rightarrow$L & 12 & text\_codebook & +0.119 [+0.083,+0.156] & 6.4 & 900 & 49 \\
Q$\rightarrow$L & 12 & text\_gen & +0.374 [+0.345,+0.408] & 9.7 & 900 & 49 \\
Q$\rightarrow$L & 16 & text\_budget & +0.048 [+0.024,+0.073] & 12.2 & 900 & 49 \\
Q$\rightarrow$L & 16 & text\_codebook & +0.146 [+0.109,+0.184] & 12.5 & 900 & 49 \\
Q$\rightarrow$L & 16 & text\_gen & +0.374 [+0.344,+0.408] & 9.5 & 900 & 49 \\
Q$\rightarrow$L & 22 & text\_budget & +0.083 [+0.058,+0.107] & 19.8 & 900 & 49 \\
Q$\rightarrow$L & 22 & text\_codebook & +0.167 [+0.131,+0.203] & 17.2 & 900 & 49 \\
Q$\rightarrow$L & 22 & text\_gen & +0.384 [+0.352,+0.419] & 18.8 & 900 & 49 \\
Q$\rightarrow$L & 32 & text\_budget & +0.151 [+0.122,+0.179] & 29.3 & 900 & 49 \\
Q$\rightarrow$L & 32 & text\_codebook & +0.154 [+0.118,+0.189] & 25.0 & 900 & 49 \\
Q$\rightarrow$L & 32 & text\_gen & +0.387 [+0.348,+0.425] & 29.1 & 900 & 49 \\
Q$\rightarrow$L & 48 & text\_budget & +0.304 [+0.268,+0.345] & 45.3 & 900 & 49 \\
Q$\rightarrow$L & 48 & text\_codebook & +0.153 [+0.117,+0.189] & 25.9 & 900 & 49 \\
Q$\rightarrow$L & 48 & text\_gen & +0.400 [+0.368,+0.433] & 40.1 & 900 & 49 \\
Q$\rightarrow$L & 64 & text\_budget & +0.332 [+0.295,+0.371] & 59.3 & 900 & 49 \\
Q$\rightarrow$L & 64 & text\_codebook & +0.153 [+0.117,+0.189] & 25.9 & 900 & 49 \\
Q$\rightarrow$L & 64 & text\_gen & +0.410 [+0.379,+0.443] & 59.7 & 900 & 49 \\
Q$\rightarrow$L & -- & inject\_matched (the packet) & +0.078 [+0.049,+0.108] & 17.0 & 900 & 49 \\
Q$\rightarrow$L & -- & inject\_oracle (reference) & +0.308 [+0.268,+0.344] & 12.0 & 900 & 49 \\
Q$\rightarrow$L & -- & text\_oracle (reference) & +0.452 [+0.421,+0.486] & -- & 900 & 49 \\
\bottomrule
\end{tabular}}
\caption{Rate-distortion, both directions (t1rd bank, qbits 8): accuracy over no communication per text encoding and byte budget, the packet's single measured operating point, and the oracle reference rows; gold-entity bootstrap (nb=8000, seed 20260802). Bytes are the full transmitted message (no framing subsidy). Adjacent points closer than the lock's resolution are unresolved.}
\label{tab:rd_curve}
\end{table}

\begin{table}[t]
\centering
\small
\resizebox{\textwidth}{!}{%
\begin{tabular}{llllll}
\toprule
cell & $\Delta=(MC-MS)-(RC-RS)$ & Holm $p$ & $RC-RS$ (random dict) & covers 0 & $HC-HS$ \\
\midrule
llama-3.1-8b -\textgreater{} qwen3-14b-base & +0.721 [+0.302,+1.192] & 0.00150 & +0.025 [-0.217,+0.256] & yes & +2.83 \\
llama-3.1-8b -\textgreater{} qwen3-30b-a3b & +2.272 [+0.508,+4.330] & 0.00700 & +0.098 [-0.384,+0.585] & yes & +15.93 \\
llama-3.1-8b -\textgreater{} qwen3-32b & +1.586 [+0.363,+3.060] & 0.00700 & +0.475 [+0.002,+0.959] & NO & +9.79 \\
llama-3.1-8b -\textgreater{} qwen3.5-27b & +1.265 [+0.594,+2.034] & 0.00087 & -0.037 [-0.111,+0.049] & yes & +6.60 \\
llama-3.1-8b -\textgreater{} qwen3.5-397b & +1.434 [+0.461,+2.496] & 0.00337 & +0.014 [-0.110,+0.150] & yes & +8.94 \\
qwen3-8b-base -\textgreater{} qwen3.5-27b & +0.399 [+0.194,+0.595] & 0.00087 & +0.041 [-0.065,+0.157] & yes & +6.60 \\
qwen3-8b-base -\textgreater{} qwen3.5-397b & +1.785 [+1.091,+2.329] & 0.00087 & +0.088 [+0.008,+0.163] & NO & +8.94 \\
\bottomrule
\end{tabular}}
\caption{ADDRPERM0908: the random-\emph{dictionary} control at matched coefficient mass, run in the confirmatory binding cells. $\Delta$ closes in 7 of 7 cells under Holm with the family fixed at freeze; the random arm covers zero in 5 of 7, which is exactly the pre-registered line and not more. The two cells where it does not are printed here rather than summarised away: a row-normalized random dictionary produces a small but significant slot effect on those receivers. Realized write norms are not matched here; the sitewise-matched control is Table~\ref{tab:normfix}. Replay mismatch was 0.0000 in every cell. $HC-HS$ is sender-independent by construction, so its 5 distinct values are per receiver.}
\label{tab:addrperm}
\end{table}

\begin{table}[t]
\centering
\small
\resizebox{\textwidth}{!}{%
\begin{tabular}{lllll}
\toprule
cell & $\Delta_{\mathrm{sem}}$ & 95\% CI & Holm $p$ & permuted-dict slot effect [CI] \\
\midrule
llama-3.1-8b -\textgreater{} qwen3-14b-base & +0.9443 & [+0.4637,+1.4804] & $<$0.00088 & -0.1985 [-0.4217,+0.0199] \\
llama-3.1-8b -\textgreater{} qwen3-30b-a3b & +2.5362 & [+0.6961,+4.6977] & 0.00550 & -0.1666 [-0.6669,+0.3041] \\
llama-3.1-8b -\textgreater{} qwen3-32b & +2.7622 & [+0.8328,+4.9024] & 0.00550 & -0.7013 [-1.7029,+0.3107] \\
llama-3.1-8b -\textgreater{} qwen3.5-27b & +1.3538 & [+0.5862,+2.2652] & $<$0.00088 & -0.1258 [-0.3825,+0.1058] \\
llama-3.1-8b -\textgreater{} qwen3.5-397b & +2.0513 & [+0.7171,+3.4395] & 0.00313 & -0.6030 [-1.4241,+0.1234] \\
qwen3-8b-base -\textgreater{} qwen3.5-27b & +0.1432 & [-0.0862,+0.3924] & 0.12812 & +0.2963 [+0.1000,+0.5015] \\
qwen3-8b-base -\textgreater{} qwen3.5-397b & +1.1997 & [+0.2795,+1.9710] & 0.01250 & +0.6740 [+0.1461,+1.2226] \\
\bottomrule
\end{tabular}}
\caption{ADDRSEM0909: the address-identity control, run IN the confirmatory binding cells. The packet is untouched; only the address$\to$row assignment inside the receiver's own dictionary is permuted, so the packet, the coefficient multiset and the wire are matched BY CONSTRUCTION. Amplitude is not: the permuted assignment writes a different subset of the same unit-norm rows, and unit rows fix only the diagonal of the Gram matrix, so the realised write is close but not exact (\NORMPERMMIN--\NORMPERMMAX{} of MC's plan-side norm, Table~\ref{tab:normaudit}). $\Delta_{\mathrm{sem}}$ closes in 6 of 7 cells under Holm with the family fixed at freeze; the permuted dictionary carries a slot effect of its own in 2 of 7, printed here rather than summarised away.}
\label{tab:addrsem}
\end{table}

\newcommand{\ASEMCELLS}{7}   %
\newcommand{\ASEMSIG}{6}   %
\begin{table}[t]
\centering
\small
\resizebox{\textwidth}{!}{%
\begin{tabular}{lllllllll}
\toprule
cell & $\gamma$ & free $I_{\mathrm{sender}}$ [95\% CI] & Holm $p$ & closes & forced $I_{\mathrm{sender}}$ [95\% CI] & cov.\ none/inj. & oracle (margin) & MC $-$ none, free (pp) \\
\midrule
llama-3.1-8b -\textgreater{} qwen3-14b-base & 8 & +0.2146 [+0.1439,+0.2919] & 0.00050 & yes & +0.8001 [+0.5736,+1.0617] & 0.997/0.895 & 0.990 (+0.492) & +9.90 [+6.04,+13.80] \\
llama-3.1-8b -\textgreater{} qwen3.5-27b & 4 & +0.0000 [-0.0345,+0.0337] & 1.00000 & no & +0.9534 [+0.6157,+1.3258] & 0.983/0.732 & 1.000 (+0.508) & -11.74 [-15.50,-8.22] \\
qwen3-8b-base -\textgreater{} qwen3.5-27b & 4 & -0.0164 [-0.0391,+0.0038] & 1.00000 & no & +0.2392 [+0.1204,+0.3731] & 0.983/0.903 & 1.000 (+0.508) & -4.19 [-6.29,-2.27] \\
qwen3-8b-base -\textgreater{} glm-4-9b & 2 & +0.0493 [+0.0148,+0.0862] & 0.00825 & yes & +0.1649 [+0.1036,+0.2368] & 0.678/0.609 & 0.990 (+0.651) & -1.01 [-3.50,+1.51] \\
\bottomrule
\end{tabular}}
\caption{FREEGENCONF0910: the free-generation endpoint on confirmatory material, at a development-selected reduced dose. Free $I_{\mathrm{sender}}$ is scored on free greedy generation (unanswered counts as incorrect); the forced-choice contrast beside it is computed on the SAME rows. The last column is the positive control that makes the two non-closing cells readable as nulls rather than as unmeasurable cells: the text-oracle arm answers correctly in free generation on those very episodes. The dose was selected on development material by a frozen coverage-only rule over the candidate grid $\{2,4,8\}$ (so a $\gamma^*=8$ cell may select 2); the bank is the entity-disjoint ENTDISJ0908 bank (test-split material with a stated exposure).}
\label{tab:freegenconf}
\end{table}

\newcommand{\FGCNETGAIN}{1}   %
\newcommand{\FGCNETCELLS}{4}   %
\begin{table}[t]
\centering
\small
\resizebox{\textwidth}{!}{%
\begin{tabular}{llllllll}
\toprule
group & $T$ (linear) & $T$ (flex) & pairs $>$ null & Holm $p$ & $T_{\mathrm{ctrl}}$ (fixed support) & $\Delta_{\mathrm{spec}}$ & positive control \\
\midrule
G1 (17-sender bank) & +0.723 [+0.656,+0.773] & +0.724 & 28/28 & 0.0010 & +0.661 [+0.578,+0.718] & +0.063 [+0.034,+0.095] & 0/8 \\
G2 (lineage bank) & +0.740 [+0.694,+0.782] & +0.745 & 15/15 & 0.0010 & -- & not measured & 0/6 \\
\bottomrule
\end{tabular}}
\caption{XSFUNC0910: do byte-disjoint codes from separately trained senders move the same receiver (qwen3.5-27b, $\gamma$ 8) the same way on the same episode? The pooled agreement is large and replicates on a second bank --- and the two columns on the right are why it licenses less than it looks. A \emph{fixed-support} sender, whose concept support cannot encode the episode, already reaches $T_{\mathrm{ctrl}}=0.66$, so only 9\% of the agreement is support-specific; and the positive control --- does the write move the receiver at all? --- fails in 14 of 14 admitted cells, both banks being relation material on which this paper reports transport as null. Reported as adjudicated, with both limits, and the pre-declared abstract sentence for this branch deliberately not taken.}
\label{tab:xsfunc}
\end{table}

\newcommand{\ROBTOLCELLS}{15}   %
\newcommand{\ROBTOLFALL}{14}   %
\newcommand{\ROBTOLRATIO}{0.63}   %

\begin{table}[t]
\centering
\small
\resizebox{\textwidth}{!}{%
\begin{tabular}{lllllllll}
\toprule
bank & cell & $I_{\mathrm{sender}}$ & $G$ & sign-flip $p$ (Holm ok) & joint gold+donor 95\% CI & acc.\ DiD pp [95\% CI] & MC$-$none / MS$-$none (pp) & within-tol.\ donors [CI] ($n$) \\
\midrule
CC1 & llama-3.1-8b $\rightarrow$ qwen3-14b-base & +0.566 & 14 & $<$0.0001 (yes) & [+0.305,+0.854] & +15.8 [+8.5,+22.5] & +10.3 / -11.8 & +0.516 [+0.215,+0.849] (167) \\
CC1 & llama-3.1-8b $\rightarrow$ qwen3-30b-a3b & +1.443 & 14 & 0.0009 (yes) & [+0.526,+2.578] & +13.2 [+3.7,+22.3] & +9.3 / -7.6 & +1.019 [-0.045,+2.234] (168) \\
CC1 & llama-3.1-8b $\rightarrow$ qwen3-32b & +1.433 & 14 & 0.0054 (yes) & [+0.388,+2.642] & +15.4 [+4.1,+28.4] & +11.1 / -12.2 & +0.906 [+0.120,+1.829] (167) \\
CC1 & llama-3.1-8b $\rightarrow$ qwen3.5-27b & +0.827 & 14 & 0.0004 (yes) & [+0.356,+1.423] & +21.1 [+13.3,+28.9] & +14.0 / -14.0 & +0.602 [+0.186,+1.125] (168) \\
CC1 & llama-3.1-8b $\rightarrow$ qwen3.5-397b & +1.034 & 14 & 0.0035 (yes) & [+0.126,+1.983] & +22.6 [+10.4,+34.8] & +18.4 / -6.1 & +0.404 [-0.018,+0.923] (64) \\
CC1 & qwen3-8b-base $\rightarrow$ qwen3.5-27b & +0.541 & 14 & 0.0005 (yes) & [+0.332,+0.735] & +21.5 [+10.5,+31.7] & +8.6 / -9.8 & +0.877 [+0.573,+1.214] (38) \\
CC1 & qwen3-8b-base $\rightarrow$ qwen3.5-397b & +1.667 & 14 & 0.0026 (yes) & [+0.631,+2.585] & +30.0 [+7.4,+46.6] & +23.2 / -12.7 & n too small (19) \\
ENTDISJ & llama-3.1-8b $\rightarrow$ qwen3-14b-base & +1.345 & 42 & $<$0.0001 (yes) & [+0.920,+1.802] & +28.0 [+19.5,+37.1] & +17.0 / -20.0 & +1.086 [+0.626,+1.662] (130) \\
ENTDISJ & llama-3.1-8b $\rightarrow$ qwen3-30b-a3b & +4.477 & 42 & $<$0.0001 (yes) & [+3.006,+5.984] & +35.4 [+25.9,+44.7] & +19.2 / -20.1 & +3.042 [+1.684,+4.735] (130) \\
ENTDISJ & llama-3.1-8b $\rightarrow$ qwen3-32b & +4.212 & 42 & $<$0.0001 (yes) & [+2.877,+5.614] & +42.7 [+31.6,+54.2] & +21.6 / -20.5 & +2.780 [+1.640,+4.265] (131) \\
ENTDISJ & llama-3.1-8b $\rightarrow$ qwen3.5-27b & +2.504 & 42 & $<$0.0001 (yes) & [+1.593,+3.454] & +33.8 [+21.0,+46.2] & +19.2 / -17.9 & +1.403 [+0.620,+2.433] (131) \\
ENTDISJ & llama-3.1-8b $\rightarrow$ qwen3.5-397b & +3.674 & 42 & $<$0.0001 (yes) & [+2.510,+4.852] & +45.7 [+35.6,+56.2] & +28.0 / -23.9 & +2.128 [+1.121,+3.278] (130) \\
ENTDISJ & qwen3-8b-base $\rightarrow$ qwen3.5-27b & +0.777 & 42 & $<$0.0001 (yes) & [+0.414,+1.174] & +13.7 [+8.5,+19.3] & +6.2 / -6.9 & +0.095 [-0.299,+0.522] (50) \\
ENTDISJ & qwen3-8b-base $\rightarrow$ qwen3.5-397b & +1.867 & 42 & $<$0.0001 (yes) & [+1.105,+2.717] & +30.7 [+20.1,+41.9] & +18.8 / -15.3 & +0.930 [-0.093,+2.072] (55) \\
ENTDISJ-GLM & llama-3.1-8b $\rightarrow$ glm-4-9b & +1.168 & 42 & $<$0.0001 (yes) & [+0.726,+1.620] & +25.9 [+16.8,+35.3] & +15.7 / -13.1 & +0.705 [+0.386,+1.122] (130) \\
ENTDISJ-GLM & qwen3-8b-base $\rightarrow$ glm-4-9b & +0.746 & 42 & $<$0.0001 (yes) & [+0.431,+1.105] & +21.2 [+11.8,+31.6] & +11.3 / -8.0 & +0.468 [+0.015,+1.027] (60) \\
\bottomrule
\end{tabular}}
\caption{Robust inference beside the primary table (descriptive; nothing here re-adjudicates a gate). Per confirmatory cell on the CC1 bank and the entity-disjoint bank (ENTDISJ0908, roster and GLM cells; exposure statement in \S\ref{sec:address_transport}): the one-sided gold-entity sign-flip randomization $p$ of $I_{\mathrm{sender}}$ (exhaustive over $2^G$ flips for $G\le16$, 20000 Monte-Carlo flips otherwise) with whether it clears the family's Holm floor $0.05/m$; the 95\% interval when gold entities AND donor entities are resampled jointly; the same difference-in-differences on forced-choice accuracy in percentage points with its gold-entity interval; the MC and MS accuracy contrasts against no message; and $I_{\mathrm{sender}}$ restricted to episodes whose realized donor met both soft clauses (code gap $\le26/255$, write norm $\le5\%$). Bootstraps $n_b=8000$, seed 20260912.}
\label{tab:cc1_robust}
\end{table}

\begin{table}[t]
\centering
\small
\resizebox{\textwidth}{!}{%
\begin{tabular}{llllllllll}
\toprule
bank & cell & attr.\ in gold slot & attr.\ anywhere & gold entity in own slot & any entity anywhere & $I_{\mathrm{sender}}$, attr.\ hit [CI] ($n$) & attr.\ no-hit [CI] ($n$) & gold-slot hit [CI] ($n$) & gold-slot no-hit [CI] ($n$) \\
\midrule
CC1 & llama-3.1-8b $\rightarrow$ qwen3-14b-base & 0.0\% & 0.0\% & 0.5\% & 14.9\% & n too small (0) & +0.566 [+0.347,+0.796] (551) & n too small (3) & +0.566 [+0.346,+0.797] (548) \\
CC1 & llama-3.1-8b $\rightarrow$ qwen3-30b-a3b & 0.0\% & 0.0\% & 0.5\% & 14.9\% & n too small (0) & +1.443 [+0.659,+2.348] (551) & n too small (3) & +1.432 [+0.641,+2.344] (548) \\
CC1 & llama-3.1-8b $\rightarrow$ qwen3-32b & 0.0\% & 0.0\% & 0.5\% & 14.9\% & n too small (0) & +1.433 [+0.467,+2.519] (551) & n too small (3) & +1.423 [+0.450,+2.516] (548) \\
CC1 & llama-3.1-8b $\rightarrow$ qwen3.5-27b & 0.0\% & 0.0\% & 0.5\% & 14.9\% & n too small (0) & +0.827 [+0.407,+1.300] (551) & n too small (3) & +0.824 [+0.401,+1.300] (548) \\
CC1 & llama-3.1-8b $\rightarrow$ qwen3.5-397b & 0.0\% & 0.0\% & 0.4\% & 13.8\% & n too small (0) & +1.034 [+0.404,+1.737] (261) & n too small (1) & +1.040 [+0.409,+1.744] (260) \\
CC1 & qwen3-8b-base $\rightarrow$ qwen3.5-27b & 0.0\% & 0.0\% & 4.3\% & 6.4\% & n too small (0) & +0.541 [+0.345,+0.712] (488) & +0.423 [-0.250,+0.759] (21) & +0.546 [+0.350,+0.720] (467) \\
CC1 & qwen3-8b-base $\rightarrow$ qwen3.5-397b & 0.0\% & 0.0\% & 4.2\% & 5.1\% & n too small (0) & +1.667 [+0.762,+2.446] (237) & n too small (10) & +1.762 [+0.859,+2.569] (227) \\
ENTDISJ & llama-3.1-8b $\rightarrow$ qwen3-14b-base & 2.8\% & 3.7\% & 24.8\% & 40.3\% & +0.838 [-0.250,+1.842] (20) & +1.365 [+1.002,+1.762] (516) & +2.636 [+1.803,+3.562] (133) & +0.919 [+0.642,+1.221] (403) \\
ENTDISJ & llama-3.1-8b $\rightarrow$ qwen3-30b-a3b & 2.8\% & 3.7\% & 24.8\% & 40.3\% & +2.194 [-1.012,+4.875] (20) & +4.566 [+3.282,+5.870] (516) & +8.568 [+5.753,+11.702] (133) & +3.127 [+1.854,+4.482] (403) \\
ENTDISJ & llama-3.1-8b $\rightarrow$ qwen3-32b & 2.8\% & 3.7\% & 24.8\% & 40.3\% & +2.663 [+1.170,+5.175] (20) & +4.273 [+3.031,+5.565] (516) & +7.488 [+5.165,+9.928] (133) & +3.131 [+1.896,+4.507] (403) \\
ENTDISJ & llama-3.1-8b $\rightarrow$ qwen3.5-27b & 2.8\% & 3.7\% & 24.8\% & 40.3\% & +1.994 [+0.222,+4.007] (20) & +2.524 [+1.636,+3.433] (516) & +4.742 [+3.204,+6.386] (133) & +1.765 [+0.902,+2.745] (403) \\
ENTDISJ & llama-3.1-8b $\rightarrow$ qwen3.5-397b & 2.8\% & 3.7\% & 24.8\% & 40.3\% & +3.109 [+1.632,+5.588] (20) & +3.696 [+2.620,+4.774] (516) & +6.380 [+4.584,+8.102] (133) & +2.781 [+1.647,+4.013] (403) \\
ENTDISJ & qwen3-8b-base $\rightarrow$ qwen3.5-27b & 13.1\% & 14.1\% & 3.1\% & 8.6\% & +3.461 [+2.155,+4.691] (77) & +0.339 [+0.128,+0.593] (471) & n too small (17) & +0.686 [+0.378,+1.027] (531) \\
ENTDISJ & qwen3-8b-base $\rightarrow$ qwen3.5-397b & 13.1\% & 14.1\% & 3.1\% & 8.6\% & +6.267 [+3.847,+9.009] (77) & +1.148 [+0.662,+1.706] (471) & n too small (17) & +1.755 [+1.110,+2.482] (531) \\
ENTDISJ-GLM & llama-3.1-8b $\rightarrow$ glm-4-9b & 2.8\% & 3.7\% & 24.8\% & 40.3\% & -0.072 [-1.148,+0.786] (20) & +1.216 [+0.832,+1.613] (516) & +2.295 [+1.322,+3.384] (133) & +0.796 [+0.407,+1.205] (403) \\
ENTDISJ-GLM & qwen3-8b-base $\rightarrow$ glm-4-9b & 13.1\% & 14.1\% & 3.1\% & 8.6\% & +2.731 [+1.552,+3.969] (77) & +0.421 [+0.230,+0.620] (471) & n too small (17) & +0.708 [+0.419,+1.045] (531) \\
\bottomrule
\end{tabular}}
\caption{Attribute-level packet attribution (descriptive; both splits pre-declared before they were computed). Per confirmatory cell, MC arm, four-arm episodes: the percentage of packets whose lane at the gold entity's slot contains the episode's attribute word, whose packet contains it in any lane, whose gold-slot lane contains the gold entity's own word (Table~\ref{tab:packet_contents}'s statistic), and whose packet names any entity; then $I_{\mathrm{sender}}$ on the episodes where the attribute word is present versus absent, and on those where the gold entity is named in its own slot versus not, plain episode means with gold-entity bootstrap intervals ($n_b=8000$, seed 20260912); splits with fewer than 20 hit episodes are not printed. The 11 test attribute words are members of the 221-word inventory; the development-bank attributes of CC1 are not, so the attribute split is undefined there. ENTDISJ cells carry the exposure statement of \S\ref{sec:address_transport}.}
\label{tab:packet_attr}
\end{table}

\begin{table}[t]
\centering
\small
\resizebox{\textwidth}{!}{%
\begin{tabular}{lllll}
\toprule
cell & $\Delta_{\mathrm{abl}}$ & 95\% CI & Holm $p$ & ablated-dict slot effect [CI] \\
\midrule
llama-3.1-8b -\textgreater{} qwen3-14b-base & +0.1426 & [-0.0676,+0.3858] & 0.70613 & +1.4174 [+1.1224,+1.7693] \\
llama-3.1-8b -\textgreater{} qwen3-30b-a3b & +0.2417 & [-0.2672,+0.8116] & 1.00000 & +4.7418 [+3.6972,+5.8301] \\
llama-3.1-8b -\textgreater{} qwen3-32b & +0.1618 & [-0.3405,+0.6858] & 1.00000 & +4.3060 [+3.2075,+5.4336] \\
llama-3.1-8b -\textgreater{} qwen3.5-27b & -0.0076 & [-0.2469,+0.2637] & 1.00000 & +2.6722 [+2.0740,+3.3250] \\
llama-3.1-8b -\textgreater{} qwen3.5-397b & -0.5159 & [-1.0941,+0.0749] & 1.00000 & +4.5019 [+3.5274,+5.5002] \\
qwen3-8b-base -\textgreater{} qwen3.5-27b & -0.1224 & [-0.2496,+0.0002] & 1.00000 & +0.8923 [+0.5532,+1.2751] \\
qwen3-8b-base -\textgreater{} qwen3.5-397b & -0.3085 & [-0.6371,-0.0130] & 1.00000 & +2.2452 [+1.4133,+3.1607] \\
\bottomrule
\end{tabular}}
\caption{ADDRABL0912: the single-address ABLATION control, run in the entity-disjoint roster cells (the bank of ENTDISJ0908, exposure statement as there) at their frozen doses. The packet is untouched; at the receiver the row of the sender's single most frequent address is zeroed (ZC/ZS arms), so the arm asks whether that dominant address is load-bearing. $\Delta_{\mathrm{abl}}=(\mathrm{MC}-\mathrm{MS})-(\mathrm{ZC}-\mathrm{ZS})$ closes in 0 of 7 cells under Holm with the family fixed at freeze (a non-rejection, not an equivalence bound; the raw interval excludes zero in 1 of 7, printed below); the ablated dictionary still carries a slot effect of its own in 7 of 7.}
\label{tab:addrabl}
\end{table}

\begin{table}[t]
\centering
\small
\resizebox{\textwidth}{!}{%
\begin{tabular}{llllllll}
\toprule
cell & episodes & four-arm $n$ & $G$ & $I_{sender}$ & gold-entity 95\% CI & $p$ (Holm) & verdict \\
\midrule
llama-3.1-8b $\rightarrow$ qwen3-14b-base & 596 & 536 & 42 & +0.800 & [+0.572,+1.059] & 0.000875 & closes \\
llama-3.1-8b $\rightarrow$ qwen3-30b-a3b & 596 & 536 & 42 & +4.477 & [+3.240,+5.805] & 0.000875 & closes \\
llama-3.1-8b $\rightarrow$ qwen3-32b & 596 & 536 & 42 & +2.231 & [+1.375,+3.178] & 0.000875 & closes \\
llama-3.1-8b $\rightarrow$ qwen3.5-27b & 596 & 536 & 42 & +2.504 & [+1.648,+3.398] & 0.000875 & closes \\
llama-3.1-8b $\rightarrow$ qwen3.5-397b & 596 & 536 & 42 & +1.676 & [+1.266,+2.087] & 0.000875 & closes \\
qwen3-8b-base $\rightarrow$ qwen3.5-27b & 596 & 548 & 42 & +0.777 & [+0.462,+1.141] & 0.000875 & closes \\
qwen3-8b-base $\rightarrow$ qwen3.5-397b & 596 & 548 & 42 & +0.978 & [+0.632,+1.350] & 0.000875 & closes \\
llama-3.1-8b $\rightarrow$ glm-4-9b & 596 & 536 & 42 & +0.758 & [+0.469,+1.065] & 0.000250 & closes \\
qwen3-8b-base $\rightarrow$ glm-4-9b & 596 & 548 & 42 & +0.746 & [+0.470,+1.089] & 0.000250 & closes \\
\bottomrule
\end{tabular}}
\caption{UNIDOSE0912: the entity-disjoint confirmatory roster (the bank of ENTDISJ0908, exposure statement as there) re-run at ONE roster-wide dose, $\gamma=8$ for every cell, fixed before the run in place of the per-pair adjudicated dose; cells whose adjudicated dose was already 8 are same-dose replays and reproduce ENTDISJ0908 exactly. Upper block: the 7 CC1-roster cells under Holm at $\alpha=0.05$ with $m=7$ fixed at freeze. Lower block: the two primary-dose cross-family cells, adjudicated as their own family (each alone with no correction; Holm $m=2$ only for the joint two-sender sentence). The resampling unit is the gold entity, 42 per cell against CC1's 14. `episodes' is the loaded count; the contrast and its interval are computed on the four-arm subset, and the effective unit behind the interval is $G$, not either episode count. Zero-write reproduces no-communication exactly in every cell. Every value here is reproduced by an independent plain-python recompute that shares no code with the adjudicator.}
\label{tab:unidose}
\end{table}

\begin{table}[t]
\centering
\small
\resizebox{\textwidth}{!}{%
\begin{tabular}{llllllll}
\toprule
bank / receiver & episodes/entities & $T$ linear [95\% CI] (flex) & raw perm.\ $p$ & $T_{\mathrm{ctrl}}$ [95\% CI] & $\Delta_{\mathrm{spec}}$ [95\% CI] & pos.\ ctrl & frozen \\
\midrule
CC1 / qwen3.5-27b ($\gamma$ 8) & 592/14 & +0.344 [+0.233,+0.520] (+0.335) & 0.0005 & +0.372 [+0.244,+0.509] & -0.028 [-0.124,+0.109] & 2/2 & VOID \\
CC1 / qwen3.5-397b ($\gamma$ 4) & 296/14 & +0.214 [+0.134,+0.304] (+0.218) & 0.0005 & +0.281 [+0.146,+0.420] & -0.067 [-0.197,+0.067] & 2/2 & VOID \\
ENTDISJ / qwen3.5-27b ($\gamma$ 8) & 596/42 & +0.361 [+0.253,+0.479] (+0.348) & 0.0005 & +0.266 [+0.214,+0.327] & +0.094 [-0.029,+0.221] & 2/2 & VOID \\
ENTDISJ / qwen3.5-397b ($\gamma$ 4) & 596/42 & +0.155 [+0.041,+0.273] (+0.172) & 0.0005 & +0.185 [+0.141,+0.232] & -0.030 [-0.159,+0.095] & 2/2 & VOID \\
\bottomrule
\end{tabular}}
\caption{XSFUNCB0912 (blind-panel item~9): per-episode agreement of $\mathrm{MC}-\mathrm{none}$ between llama-3.1-8b and qwen3-8b-base on the binding material, \texttt{none} partialled (linear; flexible in parentheses), cluster-permuted null, gold-entity bootstrap; $T_{\mathrm{ctrl}}$ is the same statistic between each real sender and a fixed-support sender (gemma-3-27b, packet diversity 1) and $\Delta_{\mathrm{spec}}=T-T_{\mathrm{ctrl}}$. \textbf{Frozen verdict PARTIAL:} every group is VOID under the lock's alignment guard ($\geq 4$ admitted senders, $\geq 6$ admitted pairs, copied from the relation-material lock; the design has 2 senders and 1 pair per group), so no string beyond PARTIAL is adjudicated and every number here is descriptive. On those numbers the positive control passes in 8 of 8 admitted cells and the real pair's agreement is matched by the fixed-support write in every group ($\Delta_{\mathrm{spec}}$ covers zero in 4 of 4; $T_{\mathrm{ctrl}} > T$ in 3).}
\label{tab:xsfuncb}
\end{table}

\begin{table}[t]
\centering
\small
\resizebox{\textwidth}{!}{%
\begin{tabular}{llllll}
\toprule
cell & stage & $n$ & tokens & ms median [p5, p95] & of which in-hook solve (ms) \\
\midrule
binding: llama-3.1-8b -\textgreater{} qwen3-14b-base (\$\textbackslash\{\}gamma\$ 16, probe221) & sender: packet (forward + batch-1 solve, per lane $\times$ lanes) & 253 & -- & 71.9 (2$\times$[10.1+25.8]) & -- \\
 & sender: text\_gen@B greedy decode (B median 22 B) & 125 & 11 new & 106.4 [85.7,133.1] & -- \\
 & receiver: none prefill & 128 & 100 & 18.0 [17.8,18.2] & 0 \\
 & receiver: text\_oracle prefill & 128 & 98 & 18.0 [17.9,18.2] & 0 \\
 & receiver: text\_packet prefill & 128 & 148 & 19.6 [19.1,23.2] & 0 \\
 & receiver: MC prefill & 128 & 100 & 430.6 [387.7,473.6] & 409.3 \\
 & receiver: zero\_packet prefill & 128 & 100 & 18.3 [18.2,18.4] & 0 \\
utility: llama-3.1-8b -\textgreater{} qwen3-32b (AC-Countries, acaug) & sender: packet (forward + batch-1 solve, per lane $\times$ lanes) & 61 & -- & 17.8 (1$\times$[9.9+7.9]) & -- \\
 & sender: text\_gen@B greedy decode (B median 16 B) & 61 & 8 new & 76.0 [75.5,85.5] & -- \\
 & receiver: no\_comm prefill & 64 & 43 & 30.7 [30.5,30.9] & 0 \\
 & receiver: text\_oracle prefill & 64 & 55 & 30.7 [30.4,34.2] & 0 \\
 & receiver: text\_byte\_matched prefill & 64 & 50 & 30.4 [29.9,33.7] & 0 \\
 & receiver: packet\_extracted prefill & 64 & 43 & 12163.3 [11992.5,12180.9] & 12128.5 \\
 & receiver: zero\_packet prefill & 64 & 43 & 31.0 [30.7,31.1] & 0 \\
\bottomrule
\end{tabular}}
\caption{COST0912: the compute axis, measured as implemented (batch 1, one B200, synchronised wall-clock, median [p5, p95] after three warm-up rows). Sender: one forward at the tap plus the batch-1 FISTA solve per lane makes the packet; a greedy text generation capped at the packet's own byte budget (text\_gen@B, max(8, B/2) new tokens, run to the cap as in the rate--distortion study) makes the text. Receiver: prefill wall-clock and prompt tokens per arm; the packet arm's hooks solve $a^0$ at every written position of every band layer, and that solve is the whole of its receiver-side cost. Descriptive; nothing here is gated.}
\label{tab:cost}
\end{table}

\begin{table}[t]
\centering
\small
\resizebox{\textwidth}{!}{%
\begin{tabular}{lllll}
\toprule
dose & cov.\ no message & cov.\ packet write & cov.\ text oracle & admitted? \\
\midrule
$\gamma=4$ & 0.4865 (72/148) & 0.2095 (31/148) & 1.0000 (148/148) & no \\
$\gamma=8$ & 0.4865 (72/148) & 0.0000 (0/148) & 1.0000 (148/148) & no \\
\bottomrule
\end{tabular}}
\caption{FREEGEN397B0912: why this paper reports no behavioural measurement on qwen3.5-397b. Format compliance (the fraction of episodes answered in the frozen free-generation format, 16 tokens) for the unwritten arm, the packet arm and the text oracle, on the same episodes. The gate's frozen rule admits a dose only if the unwritten AND the injected arm each reach 0.50; neither candidate does, so the gate stopped at stage~1, no stage-2 cell was launched and no behavioural field was read. Two things are visible: this receiver is already below the floor with no message at all, and the write --- not the prompt --- removes the rest, while the oracle on the same rows answers every time. Coverage only; no contrast is reported or implied here.}
\label{tab:freegen397b}
\end{table}

\begin{table}[t]
\centering
\small
\resizebox{\textwidth}{!}{%
\begin{tabular}{llllllll}
\toprule
point & bytes pkt/txt & no comm. & packet & text & packet $-$ text [95\% CI] & Holm $p$ & recv.\ prefill pkt/txt \\
\midrule
$k=1$ & 13/12 & 0.240 & 0.213 & 0.507 & -0.2933 [-0.4220,-0.1707] & 0.0006 & 12.1\,s / 31\,ms \\
$k=2$ & 16/14 & 0.240 & 0.220 & 0.527 & -0.3067 [-0.4274,-0.1989] & 0.0006 & 12.2\,s / 31\,ms \\
$k=4$ & 16/14 & 0.240 & 0.227 & 0.527 & -0.3000 [-0.4151,-0.1917] & 0.0006 & 12.2\,s / 31\,ms \\
$k=6$ & 16/14 & 0.240 & 0.253 & 0.527 & -0.2733 [-0.3882,-0.1636] & 0.0006 & 12.2\,s / 31\,ms \\
$k=8$ & 16/14 & 0.240 & 0.273 & 0.527 & -0.2533 [-0.3636,-0.1527] & 0.0006 & 12.3\,s / 31\,ms \\
\bottomrule
\end{tabular}}
\caption{RATECOMP0912: the packet's rate family against a budget-aware text sender on a confirmatory-strength pair (llama-3.1-8b $\to$ qwen3-32b) and the confirmatory channel, 150 items, 60 gold entities, paired per item and bootstrapped over entities. The text arm composes under each point's own realized wire bytes, so the comparison is byte-matched per item. \textbf{Two limits:} the precision half of the frozen design (5 of 10 points) is \emph{not run} --- the zero-write integrity control cannot exist when no coefficients are transmitted --- and the $k$ sweep realizes only 2 distinct wire sizes, because the grounded candidate set caps the support at a median of 2 addresses. Read this as five runs over 2 rates on one pair. At every point text wins (all intervals strictly below zero, worst Holm $p$ 0.0006) and the packet's own delta over no communication covers zero, with the oracle at 1.000 against a 0.240 baseline.}
\label{tab:ratecomp}
\end{table}

\begin{table}[t]
\centering
\small
\resizebox{\textwidth}{!}{%
\begin{tabular}{llllllll}
\toprule
sender family & episodes & candidates & both entity words & exactly one & words used & top-1 & null mean / p99 \\
\midrule
llama-3.1-8b & 592 & 266 & 0 (0.000) & 88 (0.149) & 41 & 0.0000 & 0.00005 / 0.00169 \\
qwen3-8b-base & 592 & 266 & 0 (0.000) & 57 (0.096) & 22 & 0.0000 & 0.00000 / 0.00000 \\
\bottomrule
\end{tabular}}
\caption{DECODE0912: can a reader recover the episode from the packet? The frozen test ranks every candidate entity pair in the bank (266 of them) by word overlap with the packet and counts an episode as retrieved only if its true pair is strictly first. Top-1 is \textbf{zero} in both sender families, below permutation nulls that are themselves near zero. The middle columns are a \emph{post-hoc} decomposition, computed after the frozen statistic to make it interpretable: the packet never carries both of the episode's entity words, carries exactly one in a tenth to a seventh of episodes, and each entity word recurs across a median of six candidate pairs, so a one-word match always ties. The frozen string is PARTIAL rather than NOT-DECODABLE only because that branch required the observed rate to fall below twice a null mean that is exactly zero in one family --- a defect in the threshold, reported rather than repaired. Nothing here speaks to whether the receiver uses the packet. Nor does it speak to what a different reader could recover: the frozen test is a LITERAL one, scoring word overlap alone, and it ignores slot, coefficient, attribute word and any association between words. What it licenses is that this literal reader cannot uniquely name the entity pair, not that the packet carries no episode-specific content --- the concentration split in the main text shows it does. (External review, 2026-09-13, which named exactly this over-reading.)}
\label{tab:decode}
\end{table}

\begin{table}[htbp]
\centering
\small
\begin{tabular}{lll}
\toprule
arm & gap & what it is \\
\midrule
none & +2.1250 & no message \\
MC & +2.6875 & the message assigned to the correct objects \\
MS & -4.2500 & the same message with object assignments exchanged \\
PC & +0.4375 & another example's message \\
PS & -0.2500 & the replacement message with assignments exchanged \\
zero\_packet & +2.1250 & zero vectors added through the same write operation \\
text\_packet & +2.4375 & the selected words read as text \\
text\_oracle & +12.3125 & both hidden entities named as text \\
\bottomrule
\end{tabular}
\caption{One episode, end to end (\texttt{t2\_dev\_003d4e77b8\_t0p0o0}, the primary confirmatory cell at $\gamma=16$; picked as the lexicographically first episode carrying every arm, not for its size). The sender reads two descriptions and never sees the question. Its packet is 24 bytes --- for NEXO: \texttt{drawer}, \texttt{scarf}, \texttt{stove}, \texttt{wooden}; for BLIX: \texttt{boat}, \texttt{ship} --- and the gold entity's own word (\texttt{pear}) is \textbf{not among} the words selected for that object. The receiver resolves those addresses through its own dictionary into 45 writes at layers 26, 32, 38. Exchanging the object assignments turns the contrast around; another example's message changes it much less; a packet that decodes to zero reproduces the no-message gap exactly.}
\label{tab:worked}
\end{table}

\begin{table}[t]
\centering
\small
\resizebox{\textwidth}{!}{%
\begin{tabular}{lllllll}
\toprule
cell & pair units & $I_{\mathrm{sender}}\mid q_0$ & $I_{\mathrm{sender}}\mid q_1$ & FOLLOW & SLOT & verdict \\
\midrule
llama-3.1-8b -\textgreater{} glm-4-9b $\gamma$16 & 26 & +0.083 [-0.416,+0.582] & +0.414 [+0.004,+0.846] & +0.249 [-0.024,+0.504] & -0.166 [-0.555,+0.209] & INCONCLUSIVE \\
llama-3.1-8b -\textgreater{} qwen3-30b-a3b $\gamma$8 & 26 & +1.329 [-0.800,+3.411] & +1.857 [+0.056,+3.852] & +1.593 [+0.784,+2.429] & -0.264 [-2.177,+1.516] & INCONCLUSIVE \\
llama-3.1-8b -\textgreater{} qwen3-32b $\gamma$16 & 26 & +1.546 [-0.123,+3.218] & +1.818 [+0.145,+3.535] & +1.682 [+0.573,+2.761] & -0.136 [-1.430,+1.082] & INCONCLUSIVE \\
llama-3.1-8b -\textgreater{} qwen3.5-27b $\gamma$8 & 26 & +0.788 [-0.055,+1.630] & +1.361 [+0.489,+2.326] & +1.075 [+0.550,+1.600] & -0.287 [-1.037,+0.379] & INCONCLUSIVE \\
llama-3.1-8b -\textgreater{} qwen3.5-397b $\gamma$4 & 26 & +1.901 [+0.499,+3.223] & +1.668 [+0.189,+3.220] & +1.784 [+0.821,+2.697] & +0.117 [-1.013,+1.161] & FOLLOWS-QUERY \\
qwen3-8b-base -\textgreater{} qwen3.5-27b $\gamma$8 & 27 & +0.470 [+0.006,+0.986] & +0.608 [+0.124,+1.127] & +0.539 [+0.140,+0.985] & -0.069 [-0.326,+0.194] & FOLLOWS-QUERY \\
qwen3-8b-base -\textgreater{} qwen3.5-397b $\gamma$4 & 27 & +0.748 [+0.062,+1.431] & +0.747 [+0.047,+1.467] & +0.748 [+0.196,+1.292] & +0.000 [-0.412,+0.425] & FOLLOWS-QUERY \\
\bottomrule
\end{tabular}}
\caption{QFOLLOW0913: does the effect follow the question? The sender extracts its packet ONCE and never sees a question; the receiver is then scored under two pre-fixed attribute questions whose correct aliases differ, with the same write. Per-cell verdicts are the pinned analyzer's under the rule frozen on 2026-09-05 for ENTDISJ0908's B2 arm --- FOLLOWS-QUERY iff $I_{\mathrm{sender}}$'s CI-lower is above zero under BOTH questions, SLOT-BIAS iff positive under one and negative under the other --- and this table only counts them. FOLLOW is the mean of the two per-query contrasts and SLOT their half-difference (the convention of the earlier verdict table; the analyzer stores the sum). The unit is the unordered entity pair. The two cells B2 measured are quoted in the caption and were neither re-run nor re-adjudicated: llama-3.1-8b -\textgreater{} qwen3-14b-base\_g16 FOLLOWS-QUERY (ENTDISJ0908 B2); qwen3-8b-base -\textgreater{} glm-4-9b\_g8 INCONCLUSIVE (ENTDISJ0908 B2).}
\label{tab:qfollow}
\end{table}

\begin{table}[t]
\centering
\small
\resizebox{\textwidth}{!}{%
\begin{tabular}{llllllll}
\toprule
cell & $n$ & $I_{sender}$, one layer [95\% CI] & $p$ (Holm) & $I_{sender}$, band (CC1) & retention & norm ratio & verdict \\
\midrule
llama-3.1-8b -\textgreater{} qwen3-14b-base & 551 & +0.439 [+0.27,+0.62] & 0.0009 & +0.566 [+0.35,+0.79] & 0.78 [0.67,0.90] & 0.73 & closes \\
llama-3.1-8b -\textgreater{} qwen3-30b-a3b & 551 & +1.247 [+0.62,+1.95] & 0.0009 & +1.443 [+0.67,+2.34] & 0.86 [0.77,1.04] & 0.79 & closes \\
llama-3.1-8b -\textgreater{} qwen3-32b & 551 & +1.344 [+0.41,+2.40] & 0.0019 & +1.433 [+0.48,+2.53] & 0.94 [0.76,1.13] & 0.63 & closes \\
llama-3.1-8b -\textgreater{} qwen3.5-27b & 551 & +0.811 [+0.39,+1.28] & 0.0009 & +0.827 [+0.40,+1.30] & 0.98 [0.80,1.20] & 0.65 & closes \\
llama-3.1-8b -\textgreater{} qwen3.5-397b & 261 & +1.064 [+0.45,+1.71] & 0.0009 & +1.034 [+0.40,+1.74] & 1.03 [0.68,1.79] & 0.35 & closes \\
qwen3-8b-base -\textgreater{} qwen3.5-27b & 488 & +0.662 [+0.43,+0.88] & 0.0009 & +0.541 [+0.35,+0.71] & 1.22 [1.03,1.44] & 0.56 & closes \\
qwen3-8b-base -\textgreater{} qwen3.5-397b & 237 & +0.777 [+0.38,+1.13] & 0.0009 & +1.667 [+0.79,+2.46] & 0.47 [0.34,0.63] & 0.39 & closes \\
\bottomrule
\end{tabular}}
\caption{BANDWIDTH0914: the seven confirmatory cells re-run with the receiver written at the band's lowest layer only (depth 0.4) instead of three layers (0.4/0.5/0.6); packets, dose $\gamma^*$, donor pairs and bank are the confirmatory run's, and the no-message arm replays it exactly. Gate string SINGLE\_LAYER\_SUFFICES: the contrast closes under Holm ($m=7$) in 7 of 7 cells. The single layer was chosen on one development cell after four configurations were compared. Retention (single-layer over band estimate on the same episodes, CI from the same entity resamples) and the norm ratio (the single write's realised total norm over the band's, recorded inside the forward) are descriptive. Because the single write carries less norm, a cell that does not close does not separate needing several layers from needing more written norm; no dose was re-fitted.}
\label{tab:bandwidth}
\end{table}
\newcommand{\ADRCELLS}{7}   %
\newcommand{\ADRDRAWS}{6}   %
\newcommand{\ADRSIG}{0}   %

\begin{table}[t]
\centering
\small
\resizebox{\textwidth}{!}{%
\begin{tabular}{lllllll}
\toprule
cell & draws & worst-draw CI lower & $p$ (worst draw) & frozen draw's CI lower & draw SD\,/\,SE & verdict \\
\midrule
llama-3.1-8b -\textgreater{} qwen3-14b-base & 6 & +0.090 & 0.0075 & +0.465 & 0.76 & misses \\
llama-3.1-8b -\textgreater{} qwen3-30b-a3b & 6 & -0.479 & 0.0759 & +0.725 & 0.66 & misses \\
llama-3.1-8b -\textgreater{} qwen3-32b & 6 & -0.651 & 0.1065 & +0.862 & 0.76 & misses \\
llama-3.1-8b -\textgreater{} qwen3.5-27b & 6 & +0.133 & 0.0090 & +0.578 & 0.54 & misses \\
llama-3.1-8b -\textgreater{} qwen3.5-397b & 6 & -1.653 & 0.4913 & +0.750 & 1.29 & misses \\
qwen3-8b-base -\textgreater{} qwen3.5-27b & 6 & -0.135 & 0.3680 & -0.086 & 1.99 & misses \\
qwen3-8b-base -\textgreater{} qwen3.5-397b & 6 & -0.932 & 0.2190 & +0.261 & 1.86 & misses \\
\bottomrule
\end{tabular}}
\caption{ADDRDRAW0913: the address-identity control repeated over 6 independent fixed-point-free permutations, the one ADDRSEM0909 drew included. Each cell is adjudicated on its \emph{worst} draw --- the minimum gold-entity CI lower bound and the maximum one-sided $p$ over draws --- under Holm with $m=7$ fixed at freeze, which is strictly harder than the single-draw bar ADDRSEM0909 met. Gate string DRAW-DEPENDENT: 0 of 7 cells survive their worst draw, against 6 that the frozen draw alone clears. The frozen draw enters by re-reading ADDRSEM0909's existing rows, whose no-message, MC and MS gaps are bit-identical to the new runs' on every shared episode; that is checked, not assumed. Descriptive column: the standard deviation of the draws' point estimates over the median within-draw standard error (95\% interval width / 3.92). The draw moves the estimate by a median 0.76 of a standard error (0.54--1.99), by more than one in 3 of 7 cells.}
\label{tab:addrdraw}
\end{table}

\begin{table}[t]
\centering
\small
\resizebox{\textwidth}{!}{%
\begin{tabular}{llllll}
\toprule
cell & $n$ & $\Delta^{*}$ [95\% CI] & $p$ & matched random slot effect & verdict \\
\midrule
llama-3.1-8b -\textgreater{} qwen3-14b-base & 592 & +0.689 [+0.28,+1.15] & 0.0002 & +0.056 [-0.15,+0.29] & SURVIVES-CELL \\
llama-3.1-8b -\textgreater{} qwen3-30b-a3b & 592 & +2.238 [+0.51,+4.25] & 0.0032 & +0.132 [-0.30,+0.56] & SURVIVES-CELL \\
llama-3.1-8b -\textgreater{} qwen3-32b & 592 & +1.740 [+0.58,+3.12] & 0.0006 & +0.321 [+0.02,+0.64] & SURVIVES-CELL \\
llama-3.1-8b -\textgreater{} qwen3.5-27b & 592 & +1.263 [+0.59,+2.04] & 0.0001 & -0.035 [-0.08,+0.02] & SURVIVES-CELL \\
llama-3.1-8b -\textgreater{} qwen3.5-397b & 296 & +1.449 [+0.41,+2.54] & 0.0026 & -0.001 [-0.22,+0.23] & SURVIVES-CELL \\
qwen3-8b-base -\textgreater{} qwen3.5-27b & 592 & +0.352 [+0.11,+0.57] & 0.0025 & +0.088 [-0.06,+0.25] & SURVIVES-CELL \\
qwen3-8b-base -\textgreater{} qwen3.5-397b & 296 & +1.791 [+1.07,+2.36] & 0.0001 & +0.083 [-0.02,+0.20] & SURVIVES-CELL \\
\bottomrule
\end{tabular}}
\caption{NORMFIX0913: the same contrast through a random dictionary matched in runtime edit-vector norm, site by site, to the receiver's own dictionary. Norms are recorded in float32 before bf16 conversion and addition; final rounded state differences are not matched. $\Delta^{*}=(\mathrm{MC}-\mathrm{MS})-(\mathrm{RC}^{*}-\mathrm{RS}^{*})$; the correct-slot control matches MC and the exchanged-slot control matches MS, so both legs of the control interaction stand against the leg they replace. Holm at $\alpha=0.05$ with $m=7$ fixed at freeze; gate string SURVIVES ($\Delta^{*}>0$ in 7 of 7, matched random covering zero in 6). The amplitude guard is measured, not asserted: a cell is void unless the matched arms reproduce their reference norms to $10^{-6}$ on at least 99\% of episodes.}
\label{tab:normfix}
\end{table}

\begin{table}[t]
\centering
\small
\resizebox{\textwidth}{!}{%
\begin{tabular}{lllll}
\toprule
cell & unmatched realised & unmatched plan-side & rescale factor & matched fraction \\
\midrule
llama-3.1-8b -\textgreater{} qwen3-14b-base & 1.161 & 0.844 & 0.939 & 1.0000 \\
llama-3.1-8b -\textgreater{} qwen3-30b-a3b & 1.071 & 0.846 & 1.049 & 1.0000 \\
llama-3.1-8b -\textgreater{} qwen3-32b & 1.058 & 0.851 & 1.081 & 1.0000 \\
llama-3.1-8b -\textgreater{} qwen3.5-27b & 1.139 & 0.848 & 0.937 & 1.0000 \\
llama-3.1-8b -\textgreater{} qwen3.5-397b & 0.551 & 0.837 & 3.131 & 1.0000 \\
qwen3-8b-base -\textgreater{} qwen3.5-27b & 0.825 & 0.680 & 1.516 & 1.0000 \\
qwen3-8b-base -\textgreater{} qwen3.5-397b & 0.480 & 0.625 & 3.193 & 1.0000 \\
\bottomrule
\end{tabular}}
\caption{DESCRIPTIVE amplitude audit (it may correct a factual statement and may promote no claim). `realised' is the runtime edit-vector norm before bf16 addition, summed over the written band; `plan-side' is computed against the un-intervened forward, which is exact only at the lowest written layer. They differ because the real dictionary's first write moves the residual toward its own target, leaving the upper layers little to do, while a random target does not self-cancel that way. `matched fraction' is the guard: the proportion of episodes on which the matched control reproduced its reference's realised norm to $10^{-6}$.}
\label{tab:normfix_amp}
\end{table}
\begin{table}[t]
\centering
\small
\resizebox{\textwidth}{!}{%
\begin{tabular}{lllll}
\toprule
control & cell & episodes & median $\lVert\delta\rVert$ ratio to MC & [min, max] \\
\midrule
Gaussian rows & llama-3.1-8b -\textgreater{} qwen3-14b-base & 592 & 0.843 & [0.637, 1.008] \\
Gaussian rows & llama-3.1-8b -\textgreater{} qwen3-30b-a3b & 592 & 0.848 & [0.642, 1.009] \\
Gaussian rows & llama-3.1-8b -\textgreater{} qwen3-32b & 592 & 0.851 & [0.652, 0.998] \\
Gaussian rows & llama-3.1-8b -\textgreater{} qwen3.5-27b & 592 & 0.846 & [0.654, 1.004] \\
Gaussian rows & llama-3.1-8b -\textgreater{} qwen3.5-397b & 296 & 0.839 & [0.631, 1.000] \\
Gaussian rows & qwen3-8b-base -\textgreater{} qwen3.5-27b & 592 & 0.681 & [0.623, 0.758] \\
Gaussian rows & qwen3-8b-base -\textgreater{} qwen3.5-397b & 296 & 0.643 & [0.613, 0.707] \\
the receiver's own rows, permuted & llama-3.1-8b -\textgreater{} qwen3-14b-base & 592 & 0.994 & [0.919, 1.046] \\
the receiver's own rows, permuted & llama-3.1-8b -\textgreater{} qwen3-30b-a3b & 592 & 0.987 & [0.914, 1.036] \\
the receiver's own rows, permuted & llama-3.1-8b -\textgreater{} qwen3-32b & 592 & 0.990 & [0.936, 1.035] \\
the receiver's own rows, permuted & llama-3.1-8b -\textgreater{} qwen3.5-27b & 592 & 0.994 & [0.915, 1.058] \\
the receiver's own rows, permuted & llama-3.1-8b -\textgreater{} qwen3.5-397b & 296 & 0.994 & [0.943, 1.025] \\
the receiver's own rows, permuted & qwen3-8b-base -\textgreater{} qwen3.5-27b & 592 & 1.086 & [0.981, 1.286] \\
the receiver's own rows, permuted & qwen3-8b-base -\textgreater{} qwen3.5-397b & 296 & 1.042 & [0.974, 1.145] \\
\bottomrule
\end{tabular}}
\caption{What the dictionary-substitution controls actually write. The write is $\delta=(\hat a_S-a^0_S)^{\top}D_S$, so unit-norm rows fix only the diagonal of $D_SD_S^{\top}$: the angles between the written rows, and the receiver's own $a^0$ (which changes with the dictionary), set the realized magnitude. Per episode at the lowest written layer, the Gaussian arm writes 0.64--0.85 of MC's norm and the permuted-real arm 0.99--1.09. An earlier version of this appendix called the Gaussian amplitude matched \emph{exactly}; it is not, and the correction is this table. The permuted-real arm --- the stronger of the two controls, since it moves only address identity --- is matched to about a percent, so the address-identity result does not rest on the mismatched arm. Higher layers are logged on the un-intervened forward and are proxies, so only the lowest written layer is quoted.}
\label{tab:normaudit}
\end{table}

\begin{table}[t]
\centering
\small
\resizebox{\textwidth}{!}{%
\begin{tabular}{llllll}
\toprule
arm & $L_{26}$ & $L_{32}$ & $L_{38}$ & share of total & band mean \\
\midrule
the receiver's own rows (MC) & 240.7 & 55.3 & 78.3 & 64\% / 15\% / 21\% & 125.0 \\
Gaussian rows & 206.9 & 83.1 & 109.1 & 52\% / 21\% / 27\% & 134.2 \\
ratio, Gaussian / own rows & 0.859 & 1.502 & 1.394 &  &  \\
\bottomrule
\end{tabular}}
\caption{What the three-layer band actually writes, measured inside the forward on a development cell (llama-3.1-8b $\rightarrow$ qwen3-32b, $\gamma^*{=}16$; descriptive, no gate). A three-layer write is not three equal writes: the receiver's own rows put most of the norm at the first written layer, because that write already moves the residual toward the target and leaves the layers above it little to do. A random dictionary does not self-cancel that way, so it writes less than the real dictionary at the first layer and more above it. The plan-side norm of Table~\ref{tab:normaudit} equals the first-layer figure here exactly, which is what makes it exact at that layer and a proxy above it.}
\label{tab:bandprofile}
\end{table}

\begin{table}[t]
\centering
\small
\resizebox{\textwidth}{!}{%
\begin{tabular}{lllll}
\toprule
cell & episodes & layers & median $\lVert\delta\rVert$ per layer & share of total \\
\midrule
llama-3.1-8b -\textgreater{} qwen3-14b-base & 592 & 16/20/24 & 108.45 / 27.14 / 13.46 & 73\% / 18\% / 9\% \\
llama-3.1-8b -\textgreater{} qwen3-30b-a3b & 592 & 19/24/29 & 10.72 / 1.97 / 0.91 & 79\% / 15\% / 7\% \\
llama-3.1-8b -\textgreater{} qwen3-32b & 592 & 26/32/38 & 210.09 / 52.39 / 71.23 & 63\% / 16\% / 21\% \\
llama-3.1-8b -\textgreater{} qwen3.5-27b & 592 & 26/32/38 & 52.96 / 16.15 / 11.76 & 65\% / 20\% / 15\% \\
llama-3.1-8b -\textgreater{} qwen3.5-397b & 296 & 24/30/36 & 2.21 / 2.16 / 2.02 & 35\% / 34\% / 32\% \\
qwen3-8b-base -\textgreater{} qwen3.5-27b & 592 & 26/32/38 & 44.40 / 19.36 / 15.94 & 56\% / 24\% / 20\% \\
qwen3-8b-base -\textgreater{} qwen3.5-397b & 296 & 24/30/36 & 1.94 / 1.63 / 1.34 & 39\% / 33\% / 27\% \\
\bottomrule
\end{tabular}}
\caption{The same measurement on every confirmatory cell, for the receiver's own dictionary. A multi-layer write is front-loaded: the first written layer carries 35--79\% of the band's total written norm. The flattest cell is llama-3.1-8b -\textgreater{} qwen3.5-397b, which is also the smallest write and the lowest dose in the roster; that is consistent with the mechanism, since front-loading is a consequence of the first write actually moving the residual toward the target, and a write too small to move it leaves the layers above with as much to do as the first. The ordering was read off the same table, not predicted in advance. Norms are per-episode medians of the mean over that layer's written sites. This is write geometry, not a contrast, and it does not depend on any gate's outcome.}
\label{tab:bandprofile_cells}
\end{table}

\newcommand{\NORMPERMMIN}{0.99}   %
\newcommand{\NORMPERMMAX}{1.09}   %

\title{Share the Symbol, Not the Vector:\texorpdfstring{\\}{ } Training-Free Fact Communication Across Language Models}
\makeatletter
\def\@author{Xin Li\textsuperscript{1}, Mengbing Liu\textsuperscript{1}, Hao Jiang\textsuperscript{1}, Xin Gao\textsuperscript{2}, Annan Wang\textsuperscript{1}, Xingwei Qu\textsuperscript{3}, Yichi Zhang, Chau Yuen\textsuperscript{1}\par\vskip8pt
{\fontsize{8}{10}\selectfont\itshape \textsuperscript{1}Nanyang Technological University\quad \textsuperscript{2}Yale University\quad \textsuperscript{3}University of Manchester\par}\vskip4pt
{\fontsize{9}{12}\selectfont\color{rhoaccent}%
\href{https://lixin.ai/SYNC/}{\BeginAccSupp{ActualText={}}\faGlobe\EndAccSupp{}\enspace Project}\quad\BeginAccSupp{method=hex,unicode,ActualText=00B7}$\cdot$\EndAccSupp{}\quad
\href{https://github.com/LiXin97/SYNC}{\BeginAccSupp{ActualText={}}\faGithub\EndAccSupp{}\enspace Code}\quad\BeginAccSupp{method=hex,unicode,ActualText=00B7}$\cdot$\EndAccSupp{}\quad
\href{https://huggingface.co/datasets/XINLI1997/SYNC-data}{\BeginAccSupp{ActualText={}}\faDatabase\EndAccSupp{}\enspace Data}\quad\BeginAccSupp{method=hex,unicode,ActualText=00B7}$\cdot$\EndAccSupp{}\quad
\href{https://huggingface.co/XINLI1997/SYNC-directions}{\BeginAccSupp{ActualText={}}\faCube\EndAccSupp{}\enspace Model}\par}}
\makeatother
\begin{abstract}
Multi-agent systems allow language models to solve tasks collaboratively by
sharing complementary information. These systems typically communicate
through text, while recent approaches also explore exchanging internal model
states. However, sharing these states across architectures often requires
learned translators or assumptions about model compatibility. We introduce SYNC
(\textbf{Sy}mbol-Indexed \textbf{N}eural \textbf{C}ommunication), a word-indexed interface that connects
heterogeneous models without fine-tuning or a learned translator. From descriptions only it can see, the sender
selects words from a shared list and sends their indices for each object,
without seeing the questions. The receiver writes the described facts into its hidden
states using its own word directions, vectors fitted once from its
activations. We test selective use by swapping one attribute between two
objects and requiring correct answers about both changed and unchanged
attributes before and after the swap, four answers in total. We choose an early write layer on four development receivers and then fix it.
With three different senders and nine newer receivers from five families,
four-answer success on new scenes reaches 72.0--98.3\%, up to 96 percentage
points above middle-layer writing, and the two largest receivers match plain
text. A task-free probe predicted in advance which depths each new receiver
can use (45 of 45 cases). Without
retuning, messages listing every fact keep 89.6--100\% success under new layouts, and sender messages reach
23.9--64.0\% on a landmark task with multi-token symbols, below text but from
0\% without a message. These results show
that shared word indices and receiver-local directions can support selective
fact use across model families without additional training.
\end{abstract}

\begin{document}
\maketitle
\thispagestyle{firststyle}
\section{Introduction}
\label{sec:intro}
\begin{figure}[!b]
\centering
\includegraphics[width=0.98\linewidth]{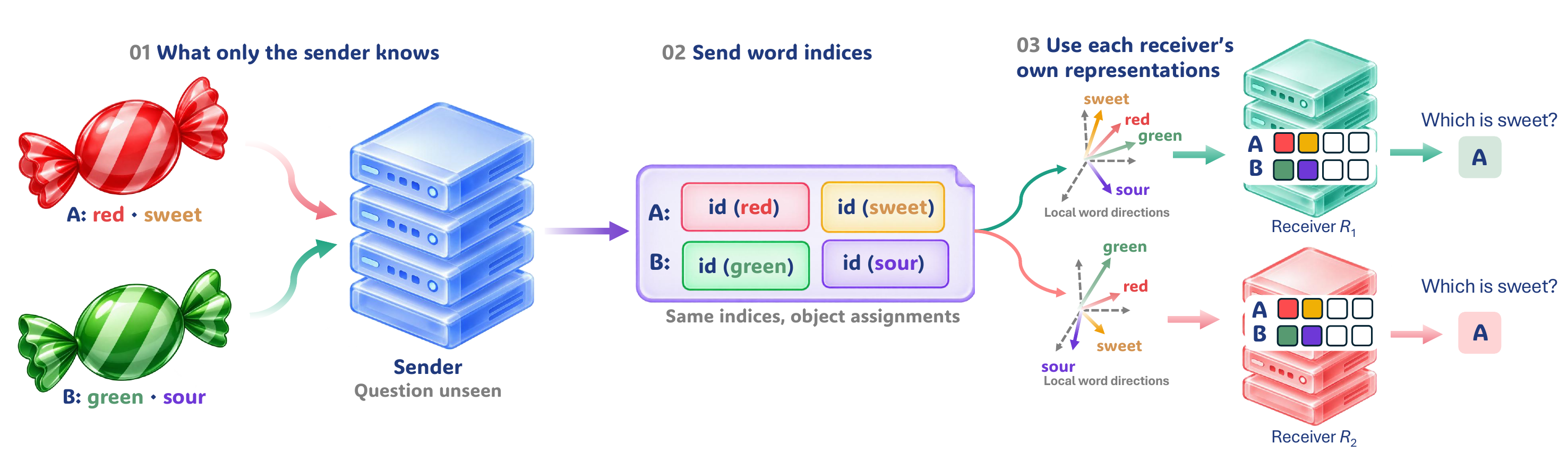}
\caption{\textbf{Shared words, each receiver's own directions.}
The sender reads private facts and sends the indices of selected words,
grouped by object, without seeing the question. Each receiver writes these
words into its hidden states with its own word directions, then answers.
Colors mark the same word across models; models and vectors are schematic,
and answers are task ground truth.}
\label{fig:teaser}\label{fig:lineage}
\end{figure}

Multi-agent systems extend language models from individual problem solving
to collaborative workflows, with applications in mathematical reasoning,
question answering, and software development
\citep{wu2024autogen,du2024debate,hong2024metagpt}. By assigning agents
complementary roles, these systems can distribute subtasks, combine
information from different sources, and refine intermediate solutions
through interaction. Such collaboration depends on communication: evidence
gathered or conclusions reached by one agent must become available for
others to use. Natural language typically provides this interface, allowing
agents built on different models to exchange observations, reasoning, and
instructions.

Recent approaches explore communication through internal model states,
whose interpretation depends on the model that produces them.
Cache-to-Cache addresses this compatibility problem by learning projection
and fusion modules between models \citep{cachetocache2510}. LatentMAS
enables training-free state sharing under structural compatibility
assumptions and discusses trainable adapters for heterogeneous agents
\citep{zou2025latentmas}. Sharing internal states between different
architectures therefore requires either a learned translator or models built
to be compatible. We seek an interface that supports useful fact transfer
across different model architectures without fine-tuning either model or
training a cross-model translator.

Our key insight is that words provide shared references even when their
internal representations differ across models. Recent work on J-space, a
set of word-associated internal representations, shows that a represented
concept can support multiple reasoning queries within a model
\mbox{\citep{verbalizable2607}}. Inspired by this finding, we introduce SYNC
(\textbf{Sy}mbol-Indexed \textbf{N}eural \textbf{C}ommunication), a word-indexed interface: a prompted sender selects words from private
evidence and transmits their indices and object assignments; the receiver
interprets these indices using its own word directions, one vector per word
in its hidden space (Figure~\ref{fig:teaser}). Each direction averages how
the receiver's hidden state changes when a neutral filler in generic inputs
is replaced by the word, and is written at positions reserved for the
relevant object. Both models remain frozen.

We evaluate whether the receiver uses the transmitted facts selectively.
Swapping one attribute between two objects while keeping the other creates
two versions of a scene, called worlds. For example, exchanging two candies'
colors should change the receiver's answer to a color question but not its
answer to a taste question. The sender never sees the questions, and the
receiver answers both from the same message. Success requires all four
answers, two per world, to be correct. Controls separate the sender's
selection from the receiver's use: plain text, complete facts, the
receiver's own states for those facts, and words pointed to the wrong
directions.

We find that this interface enables automatic fact transfer across model
families, with performance strongly dependent on where the receiver writes
the facts. We choose an early write layer on development data from four
receivers and then fix it. Without retuning, messages from three different
senders reach nine newer receivers from five families with 72.0--98.3\%
four-answer success on \RECScenes{} new scenes, while pointing each word to
another word's direction leaves at most 4.4\%. Early writing exceeds our
original writing at middle layers (depths 0.4--0.6) by 65.0--96.2 percentage points on seven of these
receivers, so a failure at one write location can substantially underestimate
what a receiver can use. The two largest receivers match plain text, and
two receivers tell which of two weighted values is stronger more reliably from
write strength than from printed numeric weights (\S\ref{sec:graded}).

Our contributions are:
\begin{itemize}
\item \textbf{A training-free interface that carries to new models.}
SYNC connects heterogeneous models through shared word indices and
receiver-local directions, without fine-tuning, a translator or paired data;
a write setting chosen on four development receivers carries unchanged to
three new senders and nine newer receivers, where all 25 cross-model pairs
recover the facts (\S\ref{sec:latest}).
\item \textbf{Where the receiver writes decides what it can use, and a probe predicts it.}
Early writing recovers delivery where middle-layer writing fails, by up to 96
points on newer receivers; a task-free probe labeled all 45 depths of nine new
receivers correctly before measurement, where a depth-only rule mislabels
seven (\S\S\ref{sec:discussion}--\ref{sec:predict}).
\item \textbf{The fixed setting carries beyond the development task.}
With complete facts, success stays at 89.6--100\% under new layouts; messages
reach 23.9--64.0\% on a landmark task with multi-token symbols, below text
(\S\ref{sec:beyond}).
\end{itemize}

\section{Method: symbol-indexed communication}
\label{sec:channel}\label{sec:formalism}\label{sec:addresses}\label{sec:write}\label{sec:identification}\label{sec:utility_gate}\label{sec:leakage}

\begin{figure}[htb]
\centering
\includegraphics[width=0.98\linewidth]{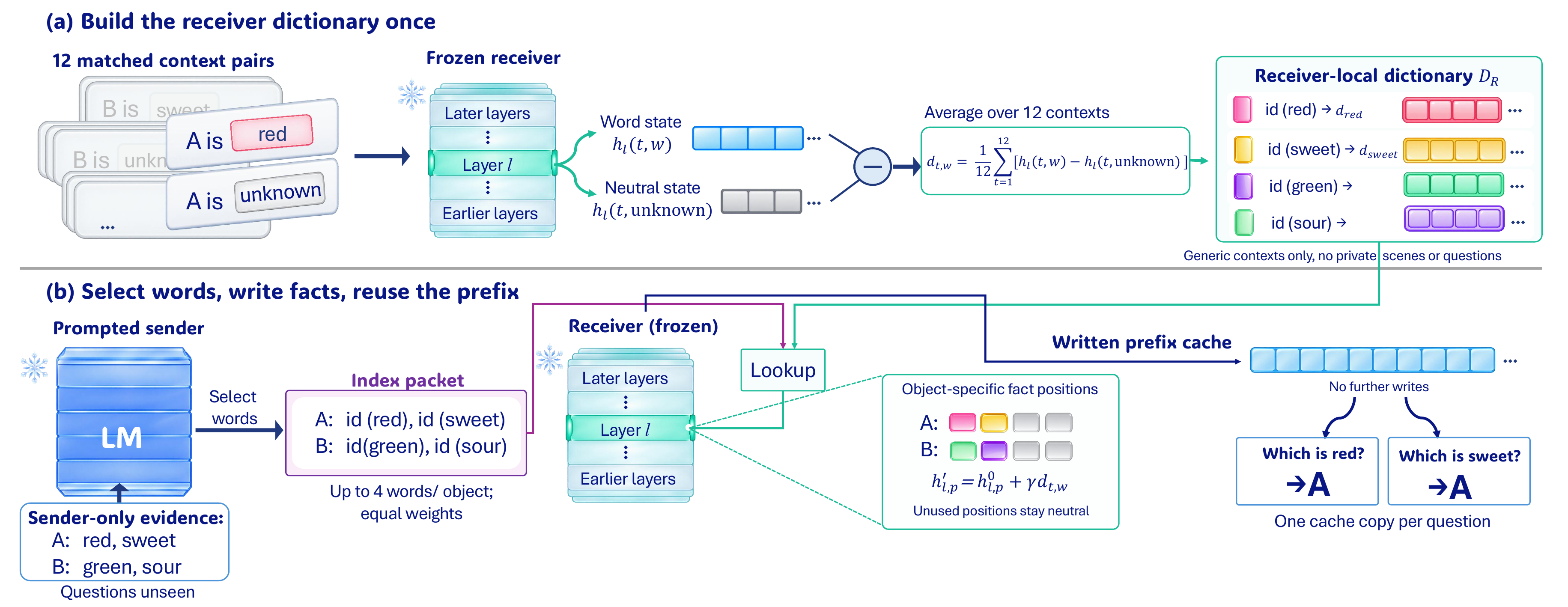}
\caption{\textbf{Construct directions once, then write and reuse facts.}
(a) Each symbol's direction averages the receiver's hidden-state difference
from a neutral filler over twelve generic contexts. (b) Retrieved directions
are added once to saved neutral states at relative depth 0.2 ($\gamma=1$); the
resulting prefix cache is copied for each question. Answers shown are task
ground truth.}
\label{fig:method}
\end{figure}

Consider a sender that reads ``$A$ is red and sweet''. It sends the indices
of \emph{red} and \emph{sweet}, both assigned to object $A$. The receiver
looks up its own vectors for these indices, writes them into positions
reserved for $A$, and answers questions from the resulting hidden states
(Figure~\ref{fig:method}). Neither model is trained: building the directions
and choosing where to write them use forward passes only.

\subsection{Private evidence and shared messages}
\label{sec:problem_setup}\label{sec:dictionary_construction}
Only the sender reads the private descriptions. The receiver sees arbitrary object names and a question about one attribute;
the sender sees neither the question nor its answer. The two models share a
fixed list (\emph{inventory}) of 221 \emph{symbols}, each an English word, and a
message format that assigns symbols to objects. Symbol indices refer to this
list, not to either model's tokenizer. Each receiver keeps a \emph{dictionary}:
a lookup table from each symbol index to a direction in its hidden space.

The sender is prompted to choose, for each object, up to four inventory
symbols that the description supports, and to answer in JSON, with no retries or exclusions. The \emph{message} (or packet) lists the selected
indices under each object in ascending order. The format also carries an
8-bit weight per symbol; automatic messages, written by a prompted sender, set every weight to 255, so they
test symbol selection and assignment, not graded weights
(Appendix~\ref{app:final_sender}).

\subsection{Fitting receiver-local symbol directions}
\label{sec:receiver_intervention}
To obtain the receiver's direction for \emph{red}, we run it on two inputs
that differ at one position, which holds either \emph{red} or the neutral
filler \texttt{unknown}, and subtract the two hidden states (residual-stream
vectors) at that position. Averaging this difference over twelve generic
contexts gives the direction; no parameters are trained.

For receiver $R$ and layer $\ell$, dictionary row $d_{\ell,w}^{(R)}$ is
\[
d_{\ell,w}^{(R)}=\frac{1}{12}\sum_{t=1}^{12}
\left[h_\ell^{(R)}(t,w)-h_\ell^{(R)}(t,\text{unknown})\right].
\]
Here $t$ indexes contexts and $h$ is the hidden vector at the final token of
the symbol or its filler; a multi-token symbol is replaced by one \texttt{unknown}
per token, preserving input length. The contexts are public message layouts
that vary object and symbol positions; they contain no private scene, question
or answer, and no activations from another model. Each layer's dictionary is
computed once and reused for every message (Appendix~\ref{app:final_fitting}).
These are not Jacobian-lens directions
(Appendix~\ref{app:dictionary_compatibility}).

\subsection{Writing facts and reusing the message}
\textbf{Prepare neutral positions.}
The receiver's prompt reserves four \texttt{unknown} slots per object,
$A{:}[\square][\square][\square][\square]$, between visible object labels,
separators and fixed weight annotations. We run this neutral prefix once and
save its hidden states $h^0_{\ell,p}$ at each layer $\ell$ and position $p$;
these are the receiver's own states, not zeros.

\textbf{Write the selected facts.}
We then run the prefix again, replacing the state at each used slot with its
saved neutral state plus the symbol's direction. In the example, two of $A$'s
slots receive the directions for \emph{red} and \emph{sweet}, in message
order; unused slots stay neutral. For symbol $w$ with transmitted weight $q_w$, this
\emph{fact-position writer} sets
\[
h'_{\ell,p}=h^0_{\ell,p}
+\gamma\frac{q_w}{q_{\max}}d_{\ell,w}^{(R)},
\]
where $q_{\max}$ is the largest weight for that object and $\gamma$ is the
write strength. The write modifies these positions at the output of each
chosen layer; later layers compute as usual.

\textbf{Answer from the written prefix.}
We save the processed prefix in the receiver's key--value cache. Each
question is appended to its own copy of this cache, with no further writes. Both questions therefore read the same message. In the \emph{text} route,
the receiver instead reads the message's symbols and weights as ordinary
tokens in the same slots. Both routes carry the same symbols and object
assignments, although their byte counts differ.

\section{Experimental setup}
\label{sec:experimental_setup}

\begin{figure}[H]
\centering
\includegraphics[width=0.85\linewidth]{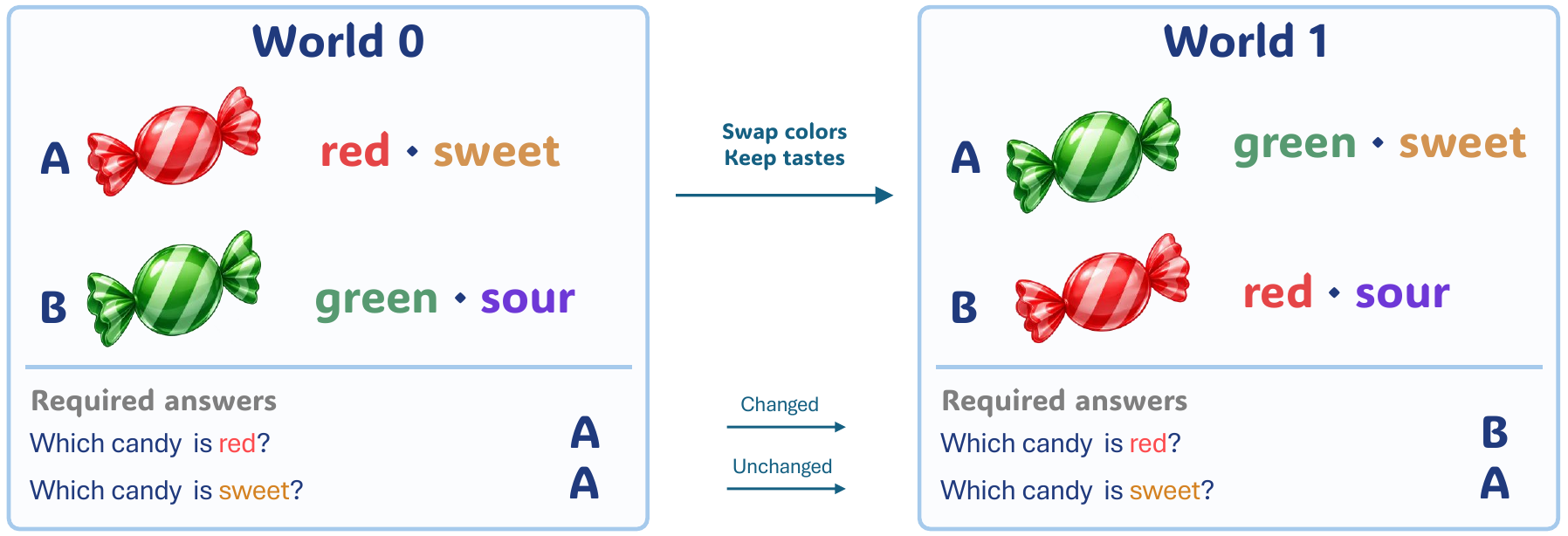}
\caption{\textbf{One changed attribute, four required answers.}
Only the sender sees the descriptions. Four-answer success requires both questions correct in both worlds. Shown answers are task ground truth, not model outputs.}
\label{fig:fact_test}
\end{figure}

\textbf{A test of selective fact use.}
Each scene consists of two worlds that describe the same two objects with two
attributes: candy color/taste, block paint/material, or coat fabric/button
material. The second world exchanges one attribute between the objects and
keeps the other. In each world, the sender reads the descriptions and sends one message, and
the receiver answers one question about each attribute from that message
(Figure~\ref{fig:fact_test}). A scene is a \emph{four-answer success} only if all four answers are correct; invalid or truncated
answers count as errors. \emph{Coverage} is the fraction of answers in a
usable format, right or wrong. Without a message, four-answer success is zero by
construction: the receiver sees the same input in both worlds, so it cannot
give the two different answers that the changed question requires.

\textbf{Models and delivery routes.}
We select the write setting with one sender, Llama-3.1-8B-Instruct, and four
development receivers: Qwen3-32B, Qwen3-14B-Base, GLM-4-9B and Gemma-3-27B-IT.
The main evaluation then fixes this setting and uses three different senders,
Gemma-4-26B-A4B, Qwen3.6-27B and gpt-oss-20b, with newer receivers from five
families (Table~\ref{tab:latest_main}). Each message is delivered in three ways: an \emph{early} write at relative depth
0.2 with $\gamma=1$; our original \emph{middle} writes at depths 0.4, 0.5 and
0.6; and plain text. \emph{Relative depth} is the written layer's index
divided by the number of layers; each written layer uses its own dictionary.

\textbf{Controls.}
\emph{Symbol--direction reassignment} points each symbol to another symbol's
direction, using three fixed permutations, while keeping the message, the
positions and the \emph{edit norm}, the size of the actual hidden-state change
at each layer. If the receiver reads the symbols rather than reacting to any
edit of this size, reassignment should remove success. \emph{Complete-fact} messages
list every required fact, removing sender errors. The \emph{native-state
reference} copies the receiver's own hidden states for these facts into the
same positions; it is an empirical comparison, not an upper bound.

\textbf{Development and confirmation.}
We choose depth and strength on \RECDevScenes{} development scenes with
complete facts, using a grid shared by all receivers and requiring 90\%
coverage, and fix the choice before \RECScenes{} new confirmation scenes with
automatic messages. New scenes change object names and descriptions but
reuse the symbols, question forms and some attribute combinations.

\textbf{Inference and uncertainty.}
All routes use FP32, batch-one greedy decoding and 64-token answers.
Estimates average over scene families (36 in the confirmation bank, 12 in the development bank). Paired bootstrap intervals
resample either the families or the twelve parent templates, keeping each
scene intact; we report the interval enclosing both, with Holm correction
across each sender's receivers, or across the prespecified tests of each
later experiment. Main-text tables give point estimates; intervals and all
contrasts are in the appendices.

\section{Results}
\label{sec:results}

\subsection{Does a fixed interface carry to new senders and receivers?}
\label{sec:latest}\label{sec:counterfactual}

We fix every choice on the development receivers (\S\ref{sec:discussion}): a
single write at relative depth 0.2 with $\gamma=1$ and twelve-context
directions. We then apply it unchanged to three different senders and to
newer receivers that played no part in any choice. Each receiver fits its own
directions, and each sender passed a prespecified message-quality check
(Appendix~\ref{app:latest}). Table~\ref{tab:latest_main} reports all 572
confirmation scenes.

\begin{table}[t]
\caption{\textbf{A fixed interface carries to new senders and receivers.}}
\label{tab:latest_main}
\centering\small
\setlength{\tabcolsep}{3.5pt}\renewcommand{\arraystretch}{1.12}
\begin{tabular*}{\linewidth}{@{\extracolsep{\fill}}llrrrrrrr@{}}
\toprule
 & & \multicolumn{2}{c}{Gemma-4 sender} & \multicolumn{2}{c}{Qwen3.6 sender} & \multicolumn{2}{c}{gpt-oss sender} & Middle \\\cmidrule(lr){3-4}\cmidrule(lr){5-6}\cmidrule(lr){7-8}
Receiver & Size & \textbf{Internal} & Text & \textbf{Internal} & Text & \textbf{Internal} & Text & (Gemma-4) \\
\midrule
Qwen3.5-2B & 2B & \textbf{81.2} & 86.7 & \textbf{79.3} & 83.8 & \textbf{77.5} & 81.5 & 4.7 \\
Qwen3.5-4B & 4B & \textbf{96.0} & 96.7 & \textbf{93.4} & 94.3 & \textbf{89.9} & 88.0 & 3.1 \\
Qwen3.5-9B & 9B & \textbf{93.4} & 94.7 & \textbf{93.6} & 96.2 & \textbf{86.6} & 88.5 & 4.9 \\
Qwen3.5-27B & 27B & \cellcolor{parity}\textbf{98.3} & \cellcolor{parity}98.6 & \cellcolor{parity}\textbf{96.4} & \cellcolor{parity}96.7 & \cellcolor{parity}\textbf{88.8} & \cellcolor{parity}89.1 & 77.8 \\
Qwen3.6-27B & 27B & \cellcolor{parity}\textbf{97.7} & \cellcolor{parity}98.2 & \textit{96.0} & \textit{96.0} & \cellcolor{parity}\textbf{90.6} & \cellcolor{parity}89.5 & 82.5 \\
\midrule
Granite-4.1 & 8B & \textbf{92.0} & 94.2 & \textbf{90.6} & 94.4 & \textbf{83.1} & 86.0 & 14.8 \\
OLMo-3 & 7B & \textbf{76.0} & 87.8 & \textbf{72.0} & 86.1 & \textbf{77.1} & 84.6 & 6.3 \\
gpt-oss & 20B & \textbf{85.3} & 97.7 & \textbf{83.8} & 95.8 & \textit{77.5} & \textit{90.9} & 4.5 \\
Mistral-Small-3.2 & 24B & \textbf{96.9} & 96.7 & \textbf{94.1} & 96.0 & \cellcolor{parity}\textbf{89.0} & \cellcolor{parity}88.8 & 0.7 \\
\bottomrule
\end{tabular*}
\par\vspace{3pt}
{\footnotesize\raggedright Four-answer success (\%) on 572 confirmation scenes; internal delivery writes at the early layer fixed on the development receivers, without retuning. Shaded: internal and text within $\pm3$ points at both clustering levels. Italic: sender and receiver are the same model, excluded from all counts. Any single symbol--direction reassignment leaves at most 4.4\% in every cell. Intervals: Appendix~\ref{app:latest}.\par}
\end{table}

\begin{findingbox}{Without retuning, three new senders reach nine newer receivers.}
Internal delivery reaches 72.0--98.3\% four-answer success in all 25
cross-model sender--receiver pairs. On the seven receivers below 27B, it
exceeds middle-layer writing by 65.0--96.2 percentage points for every sender.
\end{findingbox}

Any single reassignment of symbols to other symbols' directions, with bytes
and edit size unchanged, leaves at most 4.4\% in every cell (significant after Holm
correction for every pair), also on the four receivers that keep at least 95\%
valid answers under reassignment. Two
mixture-of-experts receivers, Qwen3.6-35B-A3B and Qwen3.5-122B-A10B, failed the
prespecified cached-versus-complete check and have no outcomes
(Appendix~\ref{app:latest}).

\subsection{Where must the receiver write the facts?}
\label{sec:discussion}

\textbf{Middle-layer writing fails on most receivers.}
Writing at depths 0.4, 0.5 and 0.6 works only on the 27B receivers
(Table~\ref{tab:latest_main}). We analyze depth on the four
development receivers, where the early write was chosen and confirmed at
\RECGLMRate{}--\RECBaseRate{}\% with the Llama sender
(Appendix~\ref{app:development_confirmation}).

\begin{figure}[t]
\centering
\includegraphics[width=\linewidth]{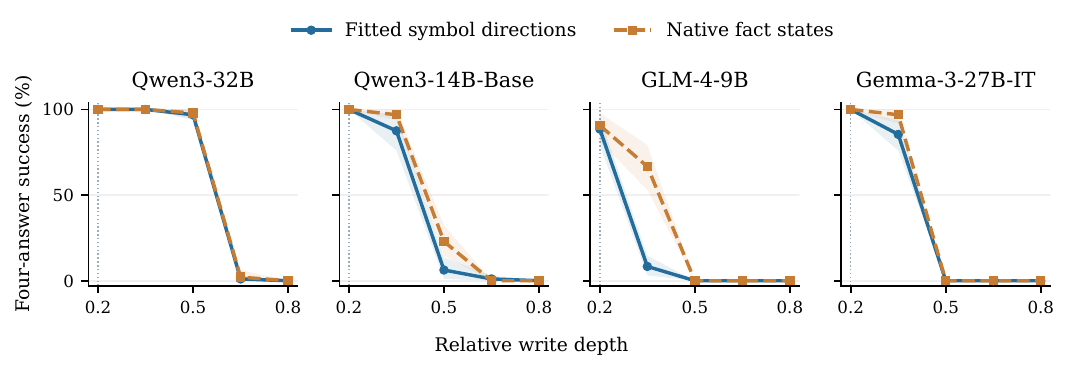}
\caption{\textbf{Usable depth ranges differ across receivers.}
Complete-fact four-answer success on \RECDevScenes{} development scenes, one
write per depth; fitted directions use $\gamma=1$, native states come from the
receiver reading the facts. Shading: 95\% family-bootstrap intervals.}
\label{fig:recovery_depth}
\end{figure}

\textbf{A depth scan with complete facts.}
Complete facts remove sender errors, isolating delivery. We write once at
relative depth 0.2, 0.35, 0.5, 0.65 or 0.8 with that layer's directions or the
native-state reference; both fail at late layers, and the usable range
differs by receiver (Figure~\ref{fig:recovery_depth}).

\begin{findingbox}{Where facts are written decides whether a receiver can use them.}
At depth 0.2, fitted directions give 100\% development success on both Qwen
receivers and Gemma and 88.5\% on GLM; only Qwen3-32B also works at 0.5.
\end{findingbox}

\begin{figure}[t]
\centering
\includegraphics[width=\linewidth]{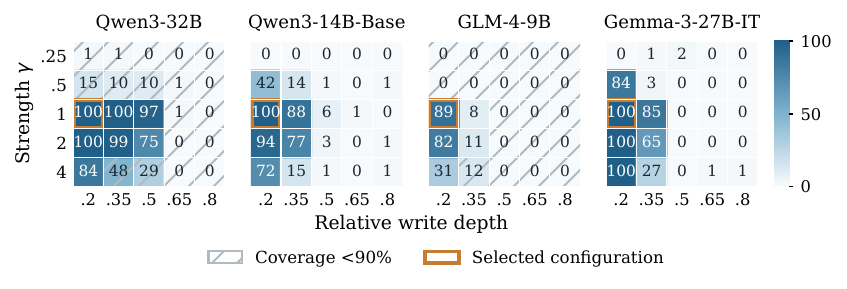}
\caption{\textbf{Depth and strength jointly determine usable writes.}
Complete-fact four-answer success (\%, rounded) on 92 development scenes.
Hatching: coverage below 90\% (scenes stay in the denominator). Orange
outlines: the configuration selected for confirmation.}
\label{fig:recovery_strength}
\end{figure}

\textbf{From development to a fixed write setting.}
Over these depths and strengths $\gamma\in\{0.25,0.5,1,2,4\}$, a fixed rule
that maximizes development success subject to 90\% valid answers selects depth
0.2 and $\gamma=1$ for all four receivers (Figure~\ref{fig:recovery_strength};
Appendix~\ref{app:receiver_recovery}). At depth 0.2, doubling the strength
keeps success above 82\% on every receiver, whereas no strength rescues
depths 0.65 or 0.8.

\textbf{Why does late writing fail?}
A prespecified diagnostic writes early, then restores the neutral keys and
values of every layer up to a cut depth. Cutting at 0.5 removes the effect on GLM
and Gemma (88.5\% and 100\% to at most 2.1\%) and lowers Qwen3-14B-Base from
100\% to 59.4\%; cutting at 0.65 leaves at most 6.2\% on all four receivers:
later tokens need fact-bearing keys and values in lower layers, which a later
write leaves neutral. Processing between layers also matters: on GLM, writing at 0.35
reaches 8.3\%, but writing at 0.2 and cutting at 0.35 reaches 72.9\%
(Appendix~\ref{app:diag}).

\textbf{Can reusable directions replace the receiver's own states?}
On the 572 confirmation scenes with complete facts, the native-state
reference is no better at the early layer: both routes exceed 99\% on the
Qwen receivers and Gemma, and GLM reaches 76.5\% with directions and 75.6\%
with native states (+0.9 [$-$1.2, 3.2]), below text at 93.0\%
(Appendix~\ref{app:development_confirmation}, Table~\ref{tab:recovery_native}). On these
receivers, directions fitted once thus come within one point of states computed
per scene (within 4.5 points on the newer receivers; Appendix~\ref{app:latest}),
without showing that the representations are identical. Simpler constructions
also work: two of three single contexts approach the twelve-context average,
and adding input-embedding differences at the input layer, which here equals
reading the symbols as text, matches the fitted directions on three receivers
and exceeds them on GLM
(Appendix~\ref{app:wordif_representations}).

\subsection{Can a receiver's usable depths be predicted in advance?}
\label{sec:predict}

Can a probe without task scenes or outcomes predict the depth scan? The
\emph{binding probe}
takes a synthetic two-object prompt with neutral slots, writes two inventory
symbols into the first slot of each object at one depth, and asks which
object carries each symbol; its value is the fraction of 110 fixed symbol
pairs for which both answers are correct. On the 20 development
receiver--depth cells, a threshold of 0.786 separates usable depths, those with
at least 50\% complete-fact success. We committed the resulting predictions for
the nine newer receivers before running their depth scans
(Table~\ref{tab:predict}); the depth-0.2 outcomes and the middle-band writes
(depths 0.4--0.6) were already known from \S\ref{sec:latest}.

\begin{table}[t]
\caption{\textbf{Usable write depths of new receivers, predicted before measurement.}}
\label{tab:predict}
\centering\small
\setlength{\tabcolsep}{3pt}\renewcommand{\arraystretch}{1.1}
\begin{tabular*}{\linewidth}{@{\extracolsep{\fill}}lrrrrrrrrr@{}}
\toprule
Relative & \multicolumn{4}{c}{Qwen3.5} & Qwen3.6 & Granite-4.1 & OLMo-3 & gpt-oss & Mistral-3.2 \\\cmidrule(lr){2-5}
depth & 2B & 4B & 9B & 27B & 27B & 8B & 7B & 20B & 24B \\
\midrule
0.2 & \cellcolor{parity}\textbf{89.6} & \cellcolor{parity}\textbf{100.0} & \cellcolor{parity}\textbf{100.0} & \cellcolor{parity}\textbf{100.0} & \cellcolor{parity}\textbf{100.0} & \cellcolor{parity}\textbf{100.0} & \cellcolor{parity}\textbf{100.0} & \cellcolor{parity}\textbf{89.6} & \cellcolor{parity}\textbf{99.0} \\
0.35 & 35.4 & 7.3 & 34.4 & \cellcolor{parity}\textbf{100.0} & \cellcolor{parity}\textbf{100.0} & \cellcolor{parity}\textbf{99.0} & 20.8 & 26.0 & 10.4 \\
0.5 & 0.0 & 0.0 & 0.0 & 18.8 & \cellcolor{parity}\textbf{85.4} & 21.9 & 2.1 & 2.1 & 0.0 \\
0.65 & 2.1 & 0.0 & 3.1 & 0.0 & 11.5 & 0.0 & 2.1 & 1.0 & 0.0 \\
0.8 & 0.0 & 0.0 & 0.0 & 0.0 & 0.0 & 0.0 & 0.0 & 0.0 & 0.0 \\
\bottomrule
\end{tabular*}
\par\vspace{3pt}
{\footnotesize\raggedright Complete-fact four-answer success (\%) on 92 development scenes, one write per depth. Shaded: predicted usable, committed before these outcomes; bold: observed usable ($\ge$50\%). A depth-only rule fitted on the development cells mislabels 7 cells. Probe values: Appendix~\ref{app:predict}.\par}
\end{table}

\begin{findingbox}{A task-free probe predicts where each new receiver can write.}
All 45 held-out predictions are correct, including the 36 beyond depth 0.2,
and the rank correlation with success is 0.89 [0.81, 0.96]. A rule that uses depth alone,
fitted on the same development cells, reaches a balanced accuracy of 0.87,
0.13 [0.08, 0.19] below the probe.
\end{findingbox}

The probe separates cases that depth alone cannot: depth 0.35 is usable on
both 27B receivers and Granite but not on the other six, and Qwen3.6-27B
remains usable at 0.5 while Qwen3.5-27B does not. We replaced the probe once, before any
held-out value: a first probe, which asked the receiver to repeat the written
symbol, scored near 1.0 on GLM at every depth, including depths where GLM
fails (Appendix~\ref{app:predict}).

\subsection{Does the fixed setting carry beyond the development task?}
\label{sec:beyond}

\textbf{New layouts.} Keeping directions, write layer, strength and parser
frozen, we use complete facts on 144 prespecified confirmation scenes and move each object's facts to its last two slots in reverse order
(\emph{slots}); replace the header, separators, weight notation and object
order (\emph{template}); or add a third object whose facts answer no question
(\emph{three objects}). A layout transfers if internal success stays within 5
points of the standard layout, by a non-inferiority test Holm-corrected over
18 receiver--layout cells.

\textbf{A landmark task.} Countries, adapted from
\citet{ramesh2025communicating}, needs the receiver's world knowledge. The
sender privately reads where two people are, such as ``QUVO is at Meteora'',
and the receiver answers ``Which object is in Greece?'' and the same for the
other country; the second world swaps the landmarks. The inventory becomes the
task's 275 landmark and country names, and every landmark spans two to nine
tokens on the Qwen receivers. The Gemma-4 sender selects the landmarks, and
each receiver fits directions for the new symbols with the frozen recipe
(Appendix~\ref{app:countries}).

\begin{table}[t]
\caption{\textbf{The fixed setting beyond the development task.}}
\label{tab:beyond}
\centering\small
\setlength{\tabcolsep}{3pt}\renewcommand{\arraystretch}{1.1}
\begin{tabular*}{\linewidth}{@{\extracolsep{\fill}}lrrrrrrrr@{}}
\toprule
 & \multicolumn{4}{c}{New layouts (complete facts)} & \multicolumn{4}{c}{Countries (Gemma-4 sender)} \\\cmidrule(lr){2-5}\cmidrule(l){6-9}
Receiver & Standard & Slots & Template & Three obj. & \textbf{Internal} & Middle & Reass. & Text \\
\midrule
Qwen3-32B & 100.0 & \cellcolor{parity}100.0 & \cellcolor{parity}98.6 & \cellcolor{parity}99.3 & \textbf{46.5} & 18.7 & 1.7 & 92.9 \\
Qwen3.5-27B & 100.0 & \cellcolor{parity}100.0 & \cellcolor{parity}100.0 & \cellcolor{parity}100.0 & \textbf{64.0} & 5.8 & 5.0 & 95.7 \\
Qwen3.6-27B & 100.0 & \cellcolor{parity}100.0 & \cellcolor{parity}100.0 & \cellcolor{parity}100.0 & \textbf{47.3} & 12.3 & 3.9 & 97.2 \\
Mistral-3.2-24B & 97.9 & \cellcolor{parity}97.2 & 89.6 & 95.1 & \textbf{23.9} & 0.0 & 0.6 & 87.1 \\
Qwen3.5-9B & 93.8 & \cellcolor{parity}99.3 & 93.8 & \cellcolor{parity}95.1 &  &  &  &  \\
Granite-4.1-8B & 96.5 & \cellcolor{parity}99.3 & 91.7 & \cellcolor{parity}97.9 &  &  &  &  \\
\bottomrule
\end{tabular*}
\par\vspace{3pt}
{\footnotesize\raggedright Four-answer success (\%). Left: internal delivery, complete facts, 144 confirmation scenes; shaded: within 5 points of Standard (non-inferiority, Holm over 18 tests); unshaded: inconclusive. Right: 172 scenes on the four larger receivers, 0\% without a message by construction; Reass.: mean of three symbol--direction reassignments. Intervals: Appendix~\ref{app:beyond}.\par}
\end{table}

\begin{findingbox}{Without retuning, the setting carries to new layouts and a landmark task.}
Across six receivers, internal success stays at 89.6--100\% in every new
layout; 14 of 18 tests pass and four are inconclusive. On Countries, four
receivers reach 23.9--64.0\%, against 0\% without a message and at most 5.0\%
on average after symbol reassignment.
\end{findingbox}

The largest layout loss is 8.3 points (Mistral, new template;
Table~\ref{tab:beyond}), while text stays at 99.3--100\% in every layout. On
Countries, the early write again exceeds the original middle-layer writes, by
23.9--58.2 points, but trails text by 31.7--63.2.

\subsection{Do receivers use weights encoded as write strength?}
\label{sec:graded}

\begin{figure}[t]
\centering
\includegraphics[width=.9\linewidth]{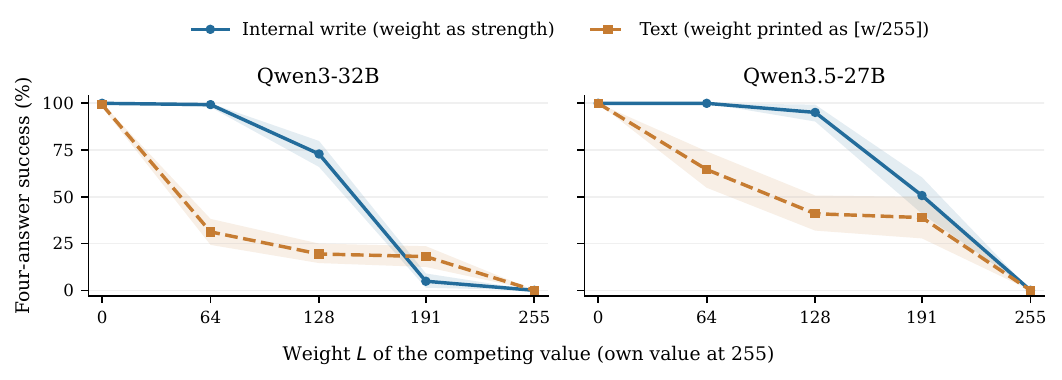}
\caption{\textbf{Graded weights: write strength versus printed weights.}
Four-answer success with complete facts as the competing value's weight $L$
rises. At $L=255$ both worlds receive identical messages, so success is zero
by construction. Shading: 95\% family-bootstrap intervals.}
\label{fig:graded}
\end{figure}

Messages can also carry an 8-bit weight per symbol. Each object gets its own
value of one attribute at weight 255 and the other object's value at a lower
weight $L$; both worlds then contain the same symbols and
differ only in which value is stronger. The internal route writes each symbol
at $L/255$ of its direction's strength, and text prints the weights, as in
``red [128/255]''. The test uses complete facts, no sender and 144
prespecified confirmation scenes (Figure~\ref{fig:graded}).

\begin{findingbox}{Two receivers follow write strength more than printed weights.}
Averaged over $L=64$, 128 and 191, internal delivery exceeds printed weights
by 36.1 [31.0, 41.4] points on Qwen3-32B and 33.8 [20.8, 48.1] on
Qwen3.5-27B, after Holm correction.
\end{findingbox}

Against five other text formats (Appendix~\ref{app:graded}), internal
delivery stays ahead on Qwen3-32B by 15.5--31.9 points; on
Qwen3.5-27B it beats four, while confidence words such as ``probably'' exceed
it by 8.8 [2.8, 14.4] points.
Close weights remain hard: at $L=191$, Qwen3-32B reaches 4.9\% internally and
18.1\% with text. Symbol reassignment at $L=128$ leaves at most 2.1\%. Of
the other two prespecified receivers, gpt-oss-20b misses the $L=0$ control and
Qwen3.5-122B-A10B fails the implementation check (Appendix~\ref{app:graded}).

\section{Related work}
\label{sec:related}

\textbf{Word directions and steering.}
J-space identifies word-associated directions involved in within-model
reasoning \citep{verbalizable2607}, motivating studies of reasoning
\citep{jcot2607}, steering \citep{reading2607} and multi-token concepts
\citep{firsttoken2608}. SYNC takes this as motivation, but its symbol directions are activation
contrasts, as in ActAdd and CAA \citep{turner2023actadd,rimsky2024caa}.
Evidence that models converge on similar representations
\citep{huh2024platonic}, or that one model can be steered, does not by itself
show that one model can send facts to another.

\textbf{Using a model to interpret hidden states.}
Patchscopes uses interventions to inspect hidden representations through
model-generated explanations \citep{ghandeharioun2024patchscopes};
SelfIE likewise interprets internal embeddings in natural language
\citep{chen2024selfie}. Where these methods read hidden states out, we write facts in, and test
whether reusable symbol directions support unseen questions and follow
controlled fact changes.

\textbf{Communication and representation transfer.}
Activation communication combines hidden states
\citep{ramesh2025communicating}; Cache-to-Cache learns to project and combine
attention caches \citep{cachetocache2510}; XKV learns receiver-aware cache
translation \citep{xkv2608}. Relative representations align models through shared anchor samples
\citep{relative2209}. CIPHER forms input embeddings from vocabulary-wide
probabilities without weight updates, assuming a shared tokenizer
\citep{cipher2310}. XBridge combines word anchors with a learned contextual bridge
\citep{xbridge2608}. SYNC shares only symbol indices, which each
receiver maps to its own directions, so it needs neither a shared tokenizer
nor a learned map.
Learned maps also transfer adapters and features between models
\citep{wang2025commandv,stitching2512,crest2608}.

\section{Discussion, limitations and conclusion}
\label{sec:limitations}\label{sec:interpretation}\label{sec:boundary}\label{sec:evaluation}\label{sec:eval:saturation}
\label{sec:utility}\label{sec:rate}

\textbf{Relation to text.}
For facts with equal weights, internal writing shows no consistent advantage: text is
similar or up to 14.1 points better on the smaller receivers, the two largest
reach parity (Table~\ref{tab:latest_main}), internal delivery takes
1.19--1.28 times as long (Appendix~\ref{app:recovery_cost}), and on Countries
text leads by 31.7--63.2 points. The clearest advantage is with graded
weights, where write strength beats printed numbers and most other text
formats (\S\ref{sec:graded}); the text still contains the weights, so this
compares encodings under our formats (Appendix~\ref{app:graded}).
\label{sec:published_channel}

\textbf{Scope.}
Four prompted senders communicate fixed inventories of 221 attribute symbols
or 275 place names to thirteen receivers from seven model families; two
mixture-of-experts receivers failed our implementation check. New scenes reuse
question forms and some attribute combinations; open-vocabulary messages,
relational facts and the TipSheets task of \citet{ramesh2025communicating} remain untested, and senders that read symbols
from their own hidden states performed poorly (Appendix~\ref{app:discussion}). 

\textbf{Interpretation.}
Moving from middle to early writing changes the layer, the number of writes
and the edit size together. Input embeddings perform at least as well as fitted
directions on single-token symbols; their comparison on multi-token symbols,
early-layer J-lens writing and the depth probe beyond nine receivers remain
untested, and latency was measured only with complete facts.

\textbf{Conclusion.}
In SYNC, shared symbol indices and each receiver's own symbol directions let heterogeneous
language models communicate facts without fine-tuning or a learned
translator. Early writing recovers delivery where middle-layer writing fails,
a task-free probe predicts where each new receiver can write, and the fixed
setting carries to newer models, new layouts and a landmark task without
retuning. What crosses between the models is the symbol; each receiver
supplies its own vector for it.
\label{sec:maintext_end}

\clearpage %
\section*{AI use statement}\label{sec:aiuse}
While preparing this paper, we relied on large-language-model assistants for
three kinds of support. For the writing, they helped draft the structure and
parts of the text, tighten wording, and format references. For the literature,
they helped search for and summarize related work. For the research itself,
working under our supervision, they helped refine hypotheses and the
experimental design, generated the synthetic task materials, implemented the
methods, evaluation and analysis code, ran and monitored the experiments, and
helped interpret the results. 
We did not use them to formulate mathematical claims or proofs, or for qualitative or thematic data analysis; these tasks do not apply to this work.
Every contribution was treated as a draft: we
reviewed it, approved the task materials, and kept a result only after it was
recounted from the stored outputs. The research question and its direction are
our own, and we are responsible for the experiments, results, interpretations
and everything that appears in this paper.

\section*{Reproducibility statement}
Appendices~\ref{app:latest}--\ref{app:graded} give the protocols of the
newer-model, prediction, transfer and graded-weight experiments, and
Appendix~\ref{app:final_interface} specifies the development sender, fitting
contexts, public layout, intervention and cache reuse. Appendix~\ref{app:final_evidence}
reports data overlap, development selection, confirmation, all controls,
uncertainty estimates and numerical checks. Tables and quantitative figures
are generated from source data with recorded hashes; independent recounts
reconstruct answers and statistics. The earlier sparse-extraction
implementation and its limits are described separately in
Appendix~\ref{app:protocol}. Code, the
preregistration locks, scene banks, sender messages, per-row records and the
analysis outputs behind the main text and Appendices~\ref{app:latest}--\ref{app:word_interface_followup},
with commands that regenerate their tables, will be released at \url{https://github.com/LiXin97/SYNC}.

\section*{Ethics statement}
The controlled tasks use mechanically generated descriptions and template
sentences over a fixed concept list; the Countries and TipSheets tasks adapted
from \citet{ramesh2025communicating} are generated mechanically as well. All experimental models are open-weight
checkpoints run locally. No human subjects or personal data were involved.

\clearpage
\appendix
\makeatletter\let\@addtocurcol\jc@defaultaddtocurcol\makeatother
\clearpage
\section*{Appendix contents}
\phantomsection\label{app:overflow}
Appendices~\ref{app:latest}--\ref{app:word_interface_followup} specify the
current method and its evidence. Appendices~\ref{app:earlier_interface}--\ref{app:utility}
summarize earlier extraction and writing comparisons that explain the design
choices and their limits. These studies use distinct configurations and
evaluation banks; their outcomes are not pooled.

\begingroup
\setlength{\parindent}{0pt}
\setlength{\parskip}{2pt}
\newcommand{\jcappentry}[3]{%
  \noindent\hspace*{#1}\hyperref[#2]{\ref*{#2}\quad #3}%
  \nobreak\leaders\hbox{\kern.25em.\kern.25em}\hfill
  \nobreak\hyperref[#2]{\pageref*{#2}}\par}
\jcappentry{0pt}{app:latest}{\textbf{Newer senders and receivers}}
\jcappentry{0pt}{app:predict}{\textbf{Predicting usable write depths}}
\jcappentry{0pt}{app:beyond}{\textbf{Beyond the development task}}
\jcappentry{1em}{app:layout}{New layouts}
\jcappentry{1em}{app:countries}{Countries}
\jcappentry{0pt}{app:graded}{\textbf{Graded weights}}
\jcappentry{0pt}{app:discussion}{\textbf{Further discussion}}
\jcappentry{0pt}{app:development_confirmation}{\textbf{Development receivers: confirmation with the Llama sender}}
\jcappentry{0pt}{app:diag}{\textbf{Why late writing fails: diagnostics}}
\smallskip
\jcappentry{0pt}{app:final_interface}{\textbf{Final interface: implementation and reproduction}}
\jcappentry{1em}{app:final_sender}{Sender, inventory and messages}
\jcappentry{1em}{app:final_fitting}{The twelve fitting contexts and public layout}
\jcappentry{1em}{app:final_writer}{Intervention, cache reuse and reference conditions}
\jcappentry{1em}{app:literal_prompts}{Literal prompts and inference settings}
\jcappentry{1em}{app:reproduction_sequence}{Reproduction sequence}
\smallskip
\jcappentry{0pt}{app:current_tasks}{\textbf{Task construction and evaluation units}}
\jcappentry{1em}{app:current_task_design}{Two worlds that differ in one fact assignment}
\jcappentry{1em}{app:current_task_split}{Development, confirmation and overlap}
\smallskip
\jcappentry{0pt}{app:final_evidence}{\textbf{Final-interface confirmation and receiver diagnostics}}
\jcappentry{1em}{app:receiver_recovery}{Write-setting selection and automatic confirmation}
\jcappentry{1em}{app:current_norm_checks}{Update magnitude and the native-state reference}
\jcappentry{1em}{app:gemma_depth_rule}{A depth rule fixed before Gemma's fitted outcomes}
\jcappentry{1em}{app:recovery_four}{Four-object transfer with the receiver fixed}
\jcappentry{1em}{app:current_error_analysis}{Message content and agreement between delivery routes}
\jcappentry{1em}{app:recovery_cost}{Timing internal and text delivery}
\smallskip
\jcappentry{0pt}{app:word_interface_followup}{\textbf{Four-object questions and simpler representations}}
\jcappentry{1em}{app:wordif_conjunction}{Four-object conjunction queries}
\jcappentry{1em}{app:wordif_representations}{Single-context directions and input embeddings}
\smallskip
\jcappentry{0pt}{app:earlier_interface}{\textbf{Earlier extraction: implementation and outcomes}}
\jcappentry{1em}{app:protocol}{Sparse messages and coefficient writing}
\jcappentry{1em}{app:earlier_binding}{Binding effects and generated answers}
\jcappentry{1em}{app:counterfactual}{Single-fact confirmation}
\smallskip
\jcappentry{0pt}{app:mechanism_diagnosis}{\textbf{Message selection and direction construction}}
\jcappentry{1em}{app:automatic_development_matrix}{Same-checkpoint message sources}
\jcappentry{1em}{app:receiver_ablation}{Each-layer directions and the update rule}
\jcappentry{1em}{app:dictionary_compatibility}{Symbol contrasts and J-lens at one layer}
\smallskip
\jcappentry{0pt}{app:utility}{\textbf{Scope of the earlier utility results}}
\endgroup
\medskip

\section{Newer senders and receivers}
\label{app:latest}

\paragraph{Protocol.}
The write setting is the one selected on the development receivers
(Appendix~\ref{app:receiver_recovery}): one write at relative depth 0.2 with
$\gamma=1$, raw twelve-context directions that each receiver fits with the
frozen recipe (Appendix~\ref{app:final_fitting}), four slots per object, and
the frozen prompts and parser. The zero-based write layers are 5 for
Qwen3.5-2B and gpt-oss-20b, 6 for Qwen3.5-4B, Qwen3.5-9B and OLMo-3-7B, 8 for
Granite-4.1-8B and Mistral-Small-3.2-24B, and 13 for both 27B receivers. For
every sender and receiver we run no message, an identity write
(implementation-check scenes only), the original middle band at depths 0.4,
0.5 and 0.6, the early write, the same message as text, three
symbol--direction reassignments with matched edit norm, and the other world's
message; complete facts are run once per receiver. All cells use the
572-scene confirmation bank after excluding the four implementation-check
scenes, FP32, eager attention, batch one and greedy decoding. A thin
compatibility layer adapts model loading, chat rendering and cache handling to
these architectures; the evaluator itself is unchanged. The senders,
receivers, thresholds and decision rules were fixed before any outcome.

\paragraph{Senders.}
Each new sender uses the prompts, inventory and parser of
Appendix~\ref{app:final_sender} and had to pass a message-quality gate on the
development scenes before generating confirmation messages
(Table~\ref{tab:latest_quality}).

\begin{table}[!htbp]
\caption{\textbf{Message quality of the three new senders.}}
\label{tab:latest_quality}
\centering\small
\begin{tabular*}{\linewidth}{@{\extracolsep{\fill}}lrrrrrrr@{}}
\toprule
 & \multicolumn{3}{c}{Development gate (96 scenes)} & \multicolumn{4}{c}{Confirmation messages (576 scenes)} \\\cmidrule(lr){2-4}\cmidrule(l){5-8}
Sender & Parsed & Recall & Complete & Parsed & Recall & Complete obj. & Bytes \\
\midrule
Gemma-4 & 100.0 & 100.0 & 100.0 & 100.0 & 98.2 & 96.4 & 21.9 \\
Qwen3.6 & 100.0 & 100.0 & 100.0 & 100.0 & 97.8 & 95.6 & 22.5 \\
gpt-oss & 99.5 & 95.4 & 90.1 & 98.9 & 94.1 & 89.2 & 20.3 \\
\bottomrule
\end{tabular*}
\par\vspace{3pt}
{\footnotesize\raggedright Percentages. Gate thresholds fixed before any output: parsed $\ge$99, recall $\ge$95, complete messages $\ge$90; all three senders pass. Recall counts required attribute symbols assigned to the correct object. Prompts, parser and inventory are those of Appendix~\ref{app:final_sender}; Gemma-4 and Qwen3.6 use their chat templates with thinking disabled, gpt-oss its lowest reasoning setting.\par}
\end{table}

\paragraph{Implementation check.}
Each receiver must reproduce the no-message computation exactly under an
identity write and agree between cached and complete forward passes (maximum
logit difference at most 0.05, alias-gap difference at most 0.1, equal parsed
answers at both budgets). All nine reported receivers pass, with logit
differences at most 0.004 and alias-gap differences at most 0.0004.
Qwen3.6-35B-A3B failed (logit difference 0.136 on one reassignment row, with
all parsed answers equal). Because expert routing in mixture-of-experts models
can flip on near-ties, we dropped the logit criterion for Qwen3.5-122B-A10B
before its first row; it still failed the alias-gap criterion (0.168 on one
reassignment row). Neither receiver has outcome rows. Seed-OSS-36B-Instruct,
also prespecified, is not reported: even with its thinking budget set to zero it
often starts a visible reasoning trace instead of an answer, leaving 28--34\%
of its internal-delivery answers unscored. Its results are included in the
released analysis outputs.

\paragraph{Results.}
Table~\ref{tab:latest_full} lists every reported cell. Every reported
cross-model pair meets the prespecified recovery rule. Within Qwen3.5, internal $-$ text narrows from 2B
to 27B by 5.2 [1.9, 8.9], 4.2 [0.9, 7.8] and 3.7 [0.5, 7.0] points for the
Gemma-4, Qwen3.6 and gpt-oss senders. This is descriptive: the prespecified
scale contrast used Qwen3.5-122B-A10B. With complete facts
(Table~\ref{tab:latest_literal}), the early write stays within 4.5 points of
the receiver's own states on every receiver; on gpt-oss-20b both fall short
of text.

\begin{table}[!htbp]
\caption{\textbf{Newer senders and receivers: all cells.}}
\label{tab:latest_full}
\centering\scriptsize
\setlength{\tabcolsep}{2pt}
\begin{tabular*}{\linewidth}{@{\extracolsep{\fill}}lrrrrrrl@{}}
\toprule
Receiver & Internal [95\% CI] & Text & Middle & Reass.\ J/valid & Valid & Int.\ $-$ text [95\% CI] & Verdict \\
\midrule
\addlinespace[2pt]\multicolumn{8}{@{}l}{\textit{Gemma-4 sender}} \\
Qwen3.5-2B & 81.2 [76.7, 85.4] & 86.7 & 4.7 & 2.5 / 99.4 & 99.9 & $-$5.6 [-9.1, -2.3] & text better \\
Qwen3.5-4B & 96.0 [93.9, 97.9] & 96.7 & 3.1 & 1.8 / 97.8 & 99.8 & $-$0.7 [-3.7, 2.1] & unresolved \\
Qwen3.5-9B & 93.4 [89.7, 96.5] & 94.7 & 4.9 & 2.9 / 93.9 & 99.3 & $-$1.4 [-3.1, 0.4] & unresolved \\
Qwen3.5-27B & 98.3 [96.2, 99.8] & 98.6 & 77.8 & 2.2 / 86.3 & 99.5 & $-$0.3 [-1.6, 0.5] & similar \\
Qwen3.6-27B & 97.7 [96.0, 99.1] & 98.2 & 82.5 & 3.0 / 93.9 & 99.9 & $-$0.5 [-1.2, 0.0] & similar \\
Granite-4.1-8B & 92.0 [87.0, 96.4] & 94.2 & 14.8 & 0.1 / 12.4 & 97.5 & $-$2.3 [-4.7, -0.3] & unresolved \\
OLMo-3-7B & 76.0 [68.6, 82.8] & 87.8 & 6.3 & 3.1 / 98.5 & 93.4 & $-$11.8 [-18.4, -5.5] & text better \\
gpt-oss-20b & 85.3 [81.5, 89.0] & 97.7 & 4.5 & 1.7 / 99.2 & 99.9 & $-$12.4 [-16.1, -8.7] & text better \\
Mistral-Small-3.2 & 96.9 [94.3, 99.0] & 96.7 & 0.7 & 1.9 / 89.4 & 100.0 & +0.2 [-3.8, 4.3] & unresolved \\
\addlinespace[2pt]\multicolumn{8}{@{}l}{\textit{Qwen3.6 sender}} \\
Qwen3.5-2B & 79.3 [74.4, 83.9] & 83.8 & 4.0 & 2.2 / 99.0 & 99.7 & $-$4.5 [-8.0, -1.6] & text better \\
Qwen3.5-4B & 93.4 [88.7, 97.2] & 94.3 & 3.1 & 1.6 / 97.7 & 99.8 & $-$0.9 [-4.0, 1.7] & unresolved \\
Qwen3.5-9B & 93.6 [88.7, 97.6] & 96.2 & 4.9 & 2.9 / 93.8 & 99.7 & $-$2.6 [-4.2, -1.2] & text better \\
Qwen3.5-27B & 96.4 [92.2, 99.8] & 96.7 & 75.7 & 2.0 / 85.9 & 99.9 & $-$0.3 [-1.2, 0.3] & similar \\
Qwen3.6-27B & 96.0 [91.3, 99.8] & 96.0 & 77.1 & 2.8 / 93.2 & 100.0 & +0.0 & \textit{same model} \\
Granite-4.1-8B & 90.6 [84.0, 96.0] & 94.4 & 14.7 & 0.0 / 12.9 & 97.6 & $-$3.8 [-8.6, -0.5] & text better \\
OLMo-3-7B & 72.0 [64.8, 78.7] & 86.1 & 7.0 & 3.3 / 98.4 & 93.2 & $-$14.1 [-19.7, -8.6] & text better \\
gpt-oss-20b & 83.8 [79.5, 87.9] & 95.8 & 4.7 & 1.6 / 99.1 & 99.9 & $-$12.0 [-16.4, -7.7] & text better \\
Mistral-Small-3.2 & 94.1 [88.9, 98.3] & 96.0 & 0.9 & 1.8 / 86.5 & 100.0 & $-$1.9 [-4.7, 0.0] & unresolved \\
\addlinespace[2pt]\multicolumn{8}{@{}l}{\textit{gpt-oss sender}} \\
Qwen3.5-2B & 77.5 [70.9, 83.5] & 81.5 & 4.2 & 2.3 / 99.8 & 100.0 & $-$4.0 [-7.1, -1.1] & text better \\
Qwen3.5-4B & 89.9 [83.9, 94.8] & 88.0 & 2.8 & 1.6 / 98.5 & 99.5 & +1.9 [-0.9, 5.3] & unresolved \\
Qwen3.5-9B & 86.6 [79.6, 92.4] & 88.5 & 4.9 & 2.8 / 94.1 & 97.9 & $-$1.9 [-4.3, 0.0] & unresolved \\
Qwen3.5-27B & 88.8 [81.6, 94.7] & 89.1 & 67.4 & 1.8 / 87.4 & 98.3 & $-$0.3 [-0.9, 0.0] & similar \\
Qwen3.6-27B & 90.6 [84.1, 95.7] & 89.5 & 74.2 & 2.9 / 94.4 & 99.2 & +1.1 [-0.2, 2.6] & similar \\
Granite-4.1-8B & 83.1 [75.0, 90.1] & 86.0 & 12.6 & 0.1 / 11.9 & 94.3 & $-$3.0 [-6.5, -0.0] & unresolved \\
OLMo-3-7B & 77.1 [69.8, 83.6] & 84.6 & 5.1 & 2.7 / 98.6 & 96.6 & $-$7.6 [-14.9, -1.8] & text better \\
gpt-oss-20b & 77.5 [71.2, 83.1] & 90.9 & 5.1 & 1.7 / 99.0 & 99.8 & $-$13.4 & \textit{same model} \\
Mistral-Small-3.2 & 89.0 [82.6, 94.2] & 88.8 & 0.7 & 2.0 / 92.6 & 99.5 & +0.2 [-2.9, 3.0] & similar \\
\bottomrule
\end{tabular*}
\par\vspace{3pt}
{\footnotesize\raggedright Four-answer success (\%) on 572 confirmation scenes; brackets give the interval enclosing family- and template-bootstrap intervals over three seeds (family interval for single estimates). Valid: fraction of answers that parse. Reassigned: mean over three edit-norm-matched symbol--direction reassignments. Every reported cross-model pair meets the prespecified recovery rule (internal $\ge$50\%, valid $\ge$90\%, internal above mean reassignment at both clustering levels after Holm correction). Comparison: similar = both intervals within $\pm3$ points; text better = upper limits below zero after Holm correction across each sender's receivers; otherwise unresolved.\par}
\end{table}

\begin{table}[!htbp]
\caption{\textbf{Complete facts on the newer receivers.}}
\label{tab:latest_literal}
\centering\small
\begin{tabular*}{\linewidth}{@{\extracolsep{\fill}}lrrrrr@{}}
\toprule
Receiver & Internal & Native states & Text & Full evidence & Internal $-$ text [95\% CI] \\
\midrule
Qwen3.5-2B & 86.3 & 85.9 & 89.6 & 97.2 & $-$3.3 [-5.6, -1.0] \\
Qwen3.5-4B & 99.0 & 99.7 & 98.4 & 100.0 & +0.5 [-0.7, 1.7] \\
Qwen3.5-9B & 96.2 & 99.8 & 100.0 & 100.0 & $-$3.8 [-7.0, -1.2] \\
Qwen3.5-27B & 100.0 & 100.0 & 100.0 & 100.0 & +0.0 [0.0, 0.0] \\
Qwen3.6-27B & 100.0 & 100.0 & 100.0 & 100.0 & +0.0 [0.0, 0.0] \\
Granite-4.1-8B & 98.1 & 97.6 & 100.0 & 99.8 & $-$1.9 [-3.1, -0.9] \\
OLMo-3-7B & 91.6 & 90.4 & 94.6 & 96.0 & $-$3.0 [-5.3, -0.6] \\
gpt-oss-20b & 86.1 & 90.6 & 99.1 & 100.0 & $-$13.1 [-17.3, -8.9] \\
Mistral-Small-3.2 & 97.6 & 98.4 & 100.0 & 100.0 & $-$2.4 [-5.0, -0.5] \\
\bottomrule
\end{tabular*}
\par\vspace{3pt}
{\footnotesize\raggedright Four-answer success (\%) on 572 confirmation scenes with complete facts and no model sender, at the fixed early layer. Native states: the receiver's own states for the facts, copied into the same positions.\par}
\end{table}

\FloatBarrier

\section{Predicting usable write depths}
\label{app:predict}

\paragraph{Probe.}
For each receiver and depth, the binding probe uses the receiver's frozen
evaluation prompt with two fixed object names, XOPA and YUME, and every slot
holding the neutral filler. The 221 inventory symbols are paired by one fixed
random permutation, giving 110 pairs $(s_a, s_b)$. For each pair, the probe
adds the fitted direction of $s_a$ at the first slot of XOPA and that of $s_b$
at the first slot of YUME ($\gamma=1$, one write at the layer of that depth),
then asks ``Which object is $s_a$?'' and ``Which object is $s_b$?'' from the
cached prefix, with 16-token answers read by the frozen parser. The probe
value is the fraction of pairs with both answers correct. No scene, task
question, answer or outcome enters the probe.

\paragraph{Rule and blind protocol.}
The outcome is complete-fact four-answer success on the 92 development scenes
with one write at relative depth 0.2, 0.35, 0.5, 0.65 or 0.8; a depth is
usable if success is at least 50\%. On the 20 development receiver--depth
cells, the threshold is the midpoint between consecutive probe values that
maximizes balanced accuracy, giving 0.786 (development balanced accuracy
1.00). The comparator uses depth alone, fitted the same way (usable if depth
$\le0.425$; development balanced accuracy 0.90). Receivers, depths, rule,
scoring and the success criterion (balanced accuracy at least 0.80 with a
lower limit above 0.50) were locked before any calibration value, and the
predictions for the nine held-out receivers were committed before any of their
depth-scan rows existed.

\paragraph{Amendment.}
The locked primary probe first asked the receiver which word is listed for
XOPA after writing one symbol there, scored as the AUC of the symbol's own
first-token logit gain against the other symbols. On one development receiver, GLM-4-9B, it scored 0.999--1.000 at every
depth while GLM's success fell from 88.5\% to 0\%, so it could not separate
usable depths. Before any held-out calibration value or outcome, we replaced it
by the binding probe, the only metric tried afterwards; this was recorded as
a dated amendment, and all other parts of the protocol are unchanged. The
repeat probe was still computed as a secondary measure on all 13 receivers:
its value is at least 0.998 at every depth, so a written symbol remains
identifiable where it can no longer be used.

\paragraph{Results.}
Table~\ref{tab:predict_full} lists every cell. The rank correlation gives tied
values their average rank; the first scoring broke ties by input order and
reported 0.90 [0.77, 0.95], a secondary statistic that we corrected after
scoring. All 45 held-out predictions
are correct, and the deepest usable depth is predicted exactly for all nine
receivers. The closest held-out case is Granite-4.1-8B at depth 0.5, with a
probe value of 0.76 and 21.9\% success. As a check that is not blind, the
prediction at depth 0.5 agrees with the middle-band result of
Table~\ref{tab:latest_main} on eight of nine receivers; the exception,
Qwen3.5-27B, succeeds with writes at 0.4, 0.5 and 0.6 but fails with a single
write at 0.5 (18.8\%), as predicted.

\begin{table}[!htbp]
\caption{\textbf{Binding probe and complete-fact success at every tested depth.}}
\label{tab:predict_full}
\centering\small
\setlength{\tabcolsep}{3pt}
\begin{tabular*}{\linewidth}{@{\extracolsep{\fill}}lrrrrr@{}}
\toprule
Receiver & 0.2 & 0.35 & 0.5 & 0.65 & 0.8 \\
\midrule
\addlinespace[2pt]\multicolumn{6}{@{}l}{\textit{Development receivers (threshold fitted here)}} \\
Qwen3-32B & 1.00 / 100.0 & 1.00 / 100.0 & 1.00 / 96.9 & 0.40 / 1.0 & 0.08 / 0.0 \\
Qwen3-14B-Base & 1.00 / 100.0 & 0.98 / 87.5 & 0.04 / 6.2 & 0.01 / 1.0 & 0.03 / 0.0 \\
GLM-4-9B & 0.89 / 88.5 & 0.68 / 8.3 & 0.09 / 0.0 & 0.12 / 0.0 & 0.12 / 0.0 \\
Gemma-3-27B-IT & 1.00 / 100.0 & 1.00 / 85.4 & 0.09 / 0.0 & 0.04 / 0.0 & 0.05 / 0.0 \\
\addlinespace[2pt]\multicolumn{6}{@{}l}{\textit{Held-out receivers (predictions committed before these outcomes)}} \\
Qwen3.5-2B & 0.95 / 89.6 & 0.21 / 35.4 & 0.02 / 0.0 & 0.02 / 2.1 & 0.02 / 0.0 \\
Qwen3.5-4B & 0.98 / 100.0 & 0.05 / 7.3 & 0.00 / 0.0 & 0.00 / 0.0 & 0.00 / 0.0 \\
Qwen3.5-9B & 0.99 / 100.0 & 0.28 / 34.4 & 0.00 / 0.0 & 0.00 / 3.1 & 0.00 / 0.0 \\
Qwen3.5-27B & 1.00 / 100.0 & 1.00 / 100.0 & 0.31 / 18.8 & 0.02 / 0.0 & 0.01 / 0.0 \\
Qwen3.6-27B & 1.00 / 100.0 & 1.00 / 100.0 & 0.93 / 85.4 & 0.18 / 11.5 & 0.01 / 0.0 \\
Granite-4.1-8B & 1.00 / 100.0 & 0.99 / 99.0 & 0.76 / 21.9 & 0.01 / 0.0 & 0.00 / 0.0 \\
OLMo-3-7B & 0.98 / 100.0 & 0.41 / 20.8 & 0.11 / 2.1 & 0.10 / 2.1 & 0.12 / 0.0 \\
gpt-oss-20B & 0.99 / 89.6 & 0.41 / 26.0 & 0.00 / 2.1 & 0.00 / 1.0 & 0.00 / 0.0 \\
Mistral-3.2-24B & 0.97 / 99.0 & 0.36 / 10.4 & 0.02 / 0.0 & 0.02 / 0.0 & 0.02 / 0.0 \\
\bottomrule
\end{tabular*}
\par\vspace{3pt}
{\footnotesize\raggedright Each cell: probe value (fraction of 110 symbol pairs with both binding answers correct) / complete-fact four-answer success (\%) on 92 development scenes. Rule: usable if the probe value is at least 0.786, the midpoint between the closest development cells on either side (GLM-4-9B at depths 0.2 and 0.35). Held-out balanced accuracy 1.00; Spearman correlation between probe and success 0.89 [0.81, 0.96]; depth-only rule (usable if depth $\le$ 0.425) balanced accuracy 0.87 [0.81, 0.92]; difference +0.13 [+0.08, +0.19]. Intervals resample the nine held-out receivers (20,000 draws, three seeds, enclosing interval).\par}
\end{table}

\FloatBarrier

\section{Beyond the development task}
\label{app:beyond}

\subsection{New layouts}
\label{app:layout}

{\raggedright The six receivers keep their directions, write layers, $\gamma=1$, parser,
questions and budgets. Material: the 144 prespecified confirmation scenes
(four per material family, balanced on every design factor) with complete
facts and no sender. The standard layout is
\texttt{A: w1 [255/255], w2 [255/255], unknown [255/255], unknown [255/255]; B: \dots}.
\emph{Slots} keeps this template but places each object's facts in its last
two slots in reverse order (\texttt{A: unknown, unknown, w2, w1}, weights
omitted here). \emph{Template} uses
\texttt{Received concepts, each followed by its strength out of 255 | B -> w1 (255) ; w2 (255) ; unknown (255) ; unknown (255) | A -> \dots\ |}.
\emph{Three objects} adds an object C with twelve slots in total; C's facts
are identical in both worlds and never match a queried value. Text delivery
shows each layout with the same facts. Standard rows reproduce the
complete-fact rows of Appendix~\ref{app:latest} exactly in tokens and logits
(1,152 of 1,152 per receiver), and every receiver passes the implementation
check in every layout (maximum logit difference $4.1\times10^{-4}$).\par}

The primary contrast is internal success in each new layout minus the
standard layout, paired by scene. A layout transfers if the one-sided test of
a loss of 5 points or more rejects after Holm correction over the 18
receiver--layout cells, fails if the upper limit lies below $-5$, and is
otherwise inconclusive. All three layouts keep each object's facts together;
layouts that interleave objects, and relational roles, were not tested with
the final interface.

\begin{table}[!htbp]
\caption{\textbf{New layouts: all cells.}}
\label{tab:layout_full}
\centering\scriptsize
\setlength{\tabcolsep}{2.5pt}
\begin{tabular*}{\linewidth}{@{\extracolsep{\fill}}llrrrrrl@{}}
\toprule
Receiver & Layout & Internal [95\% CI] & Valid & $-$ Standard [95\% CI] & Text & Reass. & Verdict \\
\midrule
Qwen3-32B & Standard & 100.0 [100.0, 100.0] & 100.0 &  & 99.3 & 1.4 & reference \\
 & Slots & 100.0 [100.0, 100.0] & 100.0 & +0.0 [0.0, 0.0] & 100.0 & 1.4 & transfers \\
 & Template & 98.6 [96.5, 100.0] & 100.0 & $-$1.4 [-3.5, 0.0] & 99.3 & 0.0 & transfers \\
 & Three obj. & 99.3 [97.9, 100.0] & 100.0 & $-$0.7 [-2.1, 0.0] & 99.3 & 0.0 & transfers \\
\addlinespace[2pt]
Qwen3.5-27B & Standard & 100.0 [100.0, 100.0] & 100.0 &  & 100.0 & 2.1 & reference \\
 & Slots & 100.0 [100.0, 100.0] & 100.0 & +0.0 [0.0, 0.0] & 100.0 & 3.5 & transfers \\
 & Template & 100.0 [100.0, 100.0] & 100.0 & +0.0 [0.0, 0.0] & 100.0 & 2.1 & transfers \\
 & Three obj. & 100.0 [100.0, 100.0] & 100.0 & +0.0 [0.0, 0.0] & 100.0 & 1.4 & transfers \\
\addlinespace[2pt]
Qwen3.6-27B & Standard & 100.0 [100.0, 100.0] & 100.0 &  & 100.0 & 4.2 & reference \\
 & Slots & 100.0 [100.0, 100.0] & 100.0 & +0.0 [0.0, 0.0] & 100.0 & 2.8 & transfers \\
 & Template & 100.0 [100.0, 100.0] & 100.0 & +0.0 [0.0, 0.0] & 100.0 & 5.6 & transfers \\
 & Three obj. & 100.0 [100.0, 100.0] & 100.0 & +0.0 [0.0, 0.0] & 100.0 & 3.5 & transfers \\
\addlinespace[2pt]
Mistral-3.2-24B & Standard & 97.9 [95.1, 100.0] & 100.0 &  & 100.0 & 2.8 & reference \\
 & Slots & 97.2 [94.4, 99.3] & 100.0 & $-$0.7 [-3.5, 2.1] & 100.0 & 2.1 & transfers \\
 & Template & 89.6 [82.6, 95.8] & 100.0 & $-$8.3 [-15.3, -2.1] & 100.0 & 2.1 & inconclusive \\
 & Three obj. & 95.1 [91.0, 98.6] & 99.1 & $-$2.8 [-8.3, 2.1] & 100.0 & 2.1 & inconclusive \\
\addlinespace[2pt]
Qwen3.5-9B & Standard & 93.8 [88.9, 97.9] & 99.8 &  & 100.0 & 0.7 & reference \\
 & Slots & 99.3 [97.9, 100.0] & 100.0 & +5.6 [0.7, 11.8] & 100.0 & 1.4 & transfers \\
 & Template & 93.8 [88.2, 97.9] & 100.0 & +0.0 [-4.9, 4.9] & 100.0 & 1.4 & inconclusive \\
 & Three obj. & 95.1 [91.7, 97.9] & 98.8 & +1.4 [-2.8, 6.2] & 100.0 & 1.4 & transfers \\
\addlinespace[2pt]
Granite-4.1-8B & Standard & 96.5 [93.8, 99.3] & 99.1 &  & 100.0 & 0.0 & reference \\
 & Slots & 99.3 [97.9, 100.0] & 99.8 & +2.8 [0.0, 6.2] & 100.0 & 0.0 & transfers \\
 & Template & 91.7 [87.5, 95.1] & 99.5 & $-$4.9 [-9.7, 0.0] & 99.3 & 0.0 & inconclusive \\
 & Three obj. & 97.9 [95.1, 100.0] & 99.5 & +1.4 [-2.1, 4.9] & 100.0 & 0.0 & transfers \\
\bottomrule
\end{tabular*}
\par\vspace{3pt}
{\footnotesize\raggedright Four-answer success (\%) on 144 confirmation scenes with complete facts; Valid: answers in a usable format; Reass.: the first symbol--direction reassignment with matched edit norm. Intervals enclose family- and template-bootstrap intervals over three seeds. Transfers: the one-sided test of a loss of at least 5 points against Standard rejects after Holm correction over 18 tests; a failure would need the upper limit below $-5$.\par}
\end{table}

\FloatBarrier

\subsection{Countries}
\label{app:countries}

\paragraph{Task.}
We use only the 400 development items of the Countries task of
\citet{ramesh2025communicating}, rebuilt mechanically; the test split stays
untouched. Pairs of items give 24 sender-gate scenes, four
implementation-check scenes and 172 evaluation scenes from 163 country pairs
and 14 continent pairs. Each scene has two objects with synthetic names at
landmarks in different countries; the second world swaps the landmarks. The
receiver answers ``Which object is in $C_1$?'' and ``Which object is in
$C_2$?'' in both worlds, so it must place each landmark in its country itself.
Without a message, four-answer success is zero by construction.

\paragraph{Inventory and directions.}
The inventory lists the task builder's 195 landmarks and 80 countries in
lower case, 275 symbols taken from the task definition only. Every landmark
spans two to nine tokens on the Qwen tokenizers (median three to four). For a
multi-token symbol, the fitting reference replaces each of its tokens with the
filler, and the direction is read at its last token; nothing else in the
fitting recipe changes.

\paragraph{Sender and controls.}
Gemma-4-26B-A4B uses the frozen two-object prompt with the new inventory. It
passed a prespecified gate on the 24 gate scenes (all messages parsed,
landmark recall 100\%, no fact assigned to the wrong object); on the
evaluation scenes it sends exactly the landmark for each object. Arms match
Appendix~\ref{app:latest}: no message, the middle band, the early write, the
same message as text, the sender's evidence as text, three symbol--direction
reassignments and the other world's message. A receiver is positive if, after
Holm correction across the four receivers, the early write exceeds both no
message and mean reassignment with positive lower limits. All four receivers
pass the implementation check and meet this rule
(Table~\ref{tab:countries_full}). An earlier hidden-state method of ours
failed on a multiple-choice version of this task (Appendix~\ref{app:utility}).

\begin{table}[!htbp]
\caption{\textbf{Countries: all receivers.}}
\label{tab:countries_full}
\centering\scriptsize
\setlength{\tabcolsep}{2pt}
\begin{tabular*}{\linewidth}{@{\extracolsep{\fill}}lrrrrrrrr@{}}
\toprule
Receiver & Internal [95\% CI] & Valid & Middle & Reass. & Text & Evidence & Int.\ $-$ reass. [95\% CI] & Int.\ $-$ text [95\% CI] \\
\midrule
Qwen3-32B & 46.5 [38.9, 56.7] & 97.9 & 18.7 & 1.7 & 92.9 & 97.5 & +44.8 [37.1, 54.6] & $-$46.4 [-54.8, -35.8] \\
Qwen3.5-27B & 64.0 [56.3, 76.0] & 100.0 & 5.8 & 5.0 & 95.7 & 96.3 & +59.0 [51.4, 69.3] & $-$31.7 [-39.6, -20.7] \\
Qwen3.6-27B & 47.3 [39.7, 58.0] & 100.0 & 12.3 & 3.9 & 97.2 & 99.4 & +43.4 [35.5, 53.1] & $-$49.9 [-57.5, -40.7] \\
Mistral-3.2-24B & 23.9 [15.1, 31.5] & 97.5 & 0.0 & 0.6 & 87.1 & 100.0 & +23.3 [14.7, 30.8] & $-$63.2 [-73.2, -52.4] \\
\bottomrule
\end{tabular*}
\par\vspace{3pt}
{\footnotesize\raggedright Four-answer success (\%) on 172 scenes, Gemma-4 sender; 0\% without a message on every receiver. Evidence: the sender's descriptions given to the receiver as text. Intervals enclose bootstrap intervals over 163 country pairs and over 14 continent pairs, three seeds each. All four receivers meet the prespecified rule (internal above no message and above mean reassignment, Holm-corrected).\par}
\end{table}

\FloatBarrier

\section{Graded weights}
\label{app:graded}

\paragraph{Design.}
Messages carry complete facts and no sender is involved. For each world,
object $o$ carries its true value of the swapped attribute $X$ at weight 255,
the other object's value of $X$ at weight $L$ (omitted at $L=0$), and its value
of the unchanged attribute $Y$ at weight 255, in ascending inventory order.
For $0<L<255$ the two worlds contain the same symbols in the same slots and
differ only in which $X$ value has the larger weight; in every world exactly
one object lists its stronger $X$ value first, so order carries no
information. At $L=255$ the two worlds are identical, so four-answer success
is zero by construction. The internal route shows the neutral layout, every
slot displayed as \texttt{unknown [255/255]}, and writes each symbol at
$(q_w/q_{\max})\,d_w$ with $q_{\max}=255$. The text route prints the same
symbols with their actual weights, \texttt{[w/255]}. A reassignment arm at
$L=128$ uses the first reassignment of Appendix~\ref{app:receiver_recovery}
with matched edit norm. The 144 scenes are a prespecified subset of the
confirmation bank, four per material family and balanced on every design
factor.

\paragraph{Controls and decision rule.}
Each route must reach at least 90\% four-answer success and coverage at
$L=0$, and outputs at $L=255$ must be identical across worlds. The primary
contrast is internal $-$ text averaged over $L=64$, 128 and 191, with Holm
correction across four prespecified receivers; ``similar'' requires both
intervals to lie within $\pm3$ points. All thresholds were fixed before any
output.

\paragraph{Results.}
Table~\ref{tab:graded_full} lists the three evaluated receivers. Both
interpretable receivers favor internal delivery, also after subtracting the
contrast at $L=0$. gpt-oss-20b favors internal delivery by a larger margin but
is not interpretable, because its internal route reaches only 84.0\% at $L=0$.
Qwen3.5-122B-A10B failed the implementation check. On Qwen3-32B, every $L=0$
answer reproduces the corresponding complete-fact confirmation row exactly
(576 of 576 prompts, tokens and logits).

Averaged over $L=64$, 128 and 191, internal delivery exceeds text with printed
weights by 36.1 [31.0, 41.4] points on Qwen3-32B and 33.8 [20.8, 48.1] on
Qwen3.5-27B. Pointing symbols to wrong directions at $L=128$ leaves at most
2.1\%. Resolution is limited when the weights are close: at $L=191$,
Qwen3-32B reaches 4.9\% internally versus 18.1\% with text. The text does
contain the weights, and keeping each object's highest-weighted values would
restore the $L=0$ text condition; the result compares two encodings under our
format, the weights carry no calibrated probability, and whether senders can
produce useful graded weights is untested.

\begin{table}[!htbp]
\caption{\textbf{Graded weights: all evaluated receivers.}}
\label{tab:graded_full}
\centering\small
\begin{tabular*}{\linewidth}{@{\extracolsep{\fill}}llrrrrr@{}}
\toprule
 & & \multicolumn{5}{c}{Weight $L$ of the competing value} \\\cmidrule(l){3-7}
Receiver & Route & 0 & 64 & 128 & 191 & 255 \\
\midrule
Qwen3-32B & Internal & 100.0 & 99.3 & 72.9 & 4.9 & 0.0 \\
 & Text & 99.3 & 31.2 & 19.4 & 18.1 & 0.0 \\
Qwen3.5-27B & Internal & 100.0 & 100.0 & 95.1 & 50.7 & 0.0 \\
 & Text & 100.0 & 64.6 & 41.0 & 38.9 & 0.0 \\
gpt-oss-20b & Internal & 84.0 & 76.4 & 63.2 & 34.0 & 0.0 \\
 & Text & 99.3 & 15.3 & 11.1 & 9.0 & 0.0 \\
\bottomrule
\end{tabular*}
\par\vspace{6pt}
{\footnotesize\setlength{\tabcolsep}{3pt}
\begin{tabular*}{\linewidth}{@{\extracolsep{\fill}}lrrrl@{}}
\toprule
Receiver & Internal $-$ text & Minus $L=0$ contrast & $-$ reassigned, $L=128$ & Interpretable \\
\midrule
Qwen3-32B & +36.1 [31.0, 41.4] & +35.4 [29.6, 41.2] & +72.9 [66.0, 79.9] & yes \\
Qwen3.5-27B & +33.8 [20.8, 48.1] & +33.8 [20.8, 48.1] & +93.1 [84.7, 99.3] & yes \\
gpt-oss-20b & +46.1 [37.3, 54.6] & +61.3 [54.2, 68.3] & +63.2 [56.2, 70.1] & no ($L=0$ control) \\
\bottomrule
\end{tabular*}}
\par\vspace{3pt}
{\footnotesize\raggedright Four-answer success (\%) on 144 scenes (top). Bottom: contrasts in points averaged over $L=64,128,191$ unless stated, with intervals enclosing family- and template-bootstrap intervals over three seeds. Holm correction runs over four prespecified receivers; Qwen3.5-122B-A10B failed the implementation check and enters with $p=1$.\par}
\end{table}

\paragraph{Other text formats.}
A follow-up, locked before its first row, printed the same weights in five
other ways on the two interpretable receivers: as percentages; as verbal labels
from strong to weak; as \texttt{[w/255]} with an added instruction that the
higher-weighted value is the true one (also given to the internal route); with
each object's values sorted by weight; and as a qualifier before each symbol
(definitely, probably, possibly, unlikely). On Qwen3-32B, internal delivery
exceeds every format by 15.5--31.9 points. On Qwen3.5-27B it exceeds four
formats by 13.0--38.4 points, whereas the qualifier format exceeds it by 8.8
[2.8, 14.4] points (Holm correction over 16 contrasts). Which encoding
conveys graded weights best therefore depends on the text format and the
receiver.
\FloatBarrier

\section{Further discussion}
\label{app:discussion}

\paragraph{A shared interface with local representations.}
Shared symbol indices say which facts hold and which object each belongs to;
each receiver supplies its own vector for every symbol, separating what the
models share from how each represents it. At the early layer, directions
fitted once on generic contexts come within one point of the receiver's own
states (\S\ref{sec:discussion}). The object assignment is not tied to one
layout: internal success stays high when facts move to other slots, the
template changes or a third object is added (\S\ref{sec:beyond}). On a
separate four-object bank, three of the four development receivers combine
two attributes sent separately and keep each attached to the right object
(Appendix~\ref{app:wordif_conjunction}).

\paragraph{Where fixed directions and the receiver's own states differ.}
On GLM at depth 0.35, fixed directions reach 8.3\%, a direction computed from
each scene's own neutral prompt with only that fact in its slot 40.6\%,
native states 66.7\% and text 99.0\%. Layout and position, co-occurring facts
and the write path thus each contribute (Appendix~\ref{app:diag}).

\paragraph{Reading a symbol is not using it.}
A written symbol stays identifiable at every tested depth: it raises the
receiver's logit for its own first token more than nearly any other symbol
does (AUC at least 0.998 on all 13 receivers), even where four-answer success
is zero. What late writes lose is the link between symbol and object, which
the binding probe measures and which predicts success (\S\ref{sec:predict}).
For a new receiver, this gives a check without task data: fit its
directions, run the probe at a few depths, and write where it passes.

\paragraph{Reading symbols from hidden states remains hard.}
Our sender chooses symbols by prompting. An earlier sender that read symbol
weights from its own hidden states never met the four-answer criterion and
improved generated-answer accuracy in only \FGCNETGAIN{} of \FGCNETCELLS{}
evaluable model pairs; its messages also conveyed less than prompted
selections as text, pointing to extraction as the limitation
(Appendices~\ref{app:earlier_binding}, \ref{app:automatic_development_matrix}).
J-lens directions underperform ours at a single middle layer; early-layer
J-lens writing remains untested (Appendix~\ref{app:dictionary_compatibility}).

\paragraph{Scale.}
Within Qwen3.5, the gap between internal delivery and text narrows from 2B to
27B by 5.2 [1.9, 8.9] points with the Gemma-4 sender. This trend is
descriptive, since our prespecified scale test required Qwen3.5-122B-A10B,
which failed the implementation check (Appendix~\ref{app:latest}).

\section{Development receivers: confirmation with the Llama sender}
\label{app:development_confirmation}

The write setting was selected on the development scenes of these four
receivers (Appendix~\ref{app:receiver_recovery}) and first confirmed with the
Llama-3.1-8B-Instruct sender on the same 572 confirmation scenes used in
\S\ref{sec:latest}. Table~\ref{tab:recovery_main} compares internal and text
delivery of identical messages.

\begin{table}[H]
\caption{\textbf{Automatic fact communication across receiver families.}}
\label{tab:recovery_main}
\centering\normalsize
\setlength{\tabcolsep}{3pt}\renewcommand{\arraystretch}{1.16}
\begin{tabular*}{\linewidth}{@{\extracolsep{\fill}}lrrrr@{}}
\toprule
Receiver & Middle & \textbf{Early [95\% CI]} & Text & \shortstack[r]{Reassigned\\Success / Valid} \\
\midrule
Qwen3-32B & 94.4 & \textbf{97.0}\enspace {\footnotesize [93.2, 99.5]} & 97.0 & 1.2 / 62.2 \\
Qwen3-14B-Base & 31.0 & \textbf{97.7}\enspace {\footnotesize [95.1, 99.8]} & 97.7 & 2.3 / 99.2 \\
GLM-4-9B & 0.5 & \textbf{74.3}\enspace {\footnotesize [66.9, 81.3]} & 87.8 & 0.6 / 49.8 \\
Gemma-3-27B-IT & 30.5 & \textbf{97.7}\enspace {\footnotesize [94.8, 100.0]} & 97.9 & 3.0 / 89.8 \\
\bottomrule
\end{tabular*}
\par\vspace{3pt}
{\footnotesize\raggedright Four-answer success (\%) on 572 scenes, averaged over 36 material families. One Llama-3.1-8B-Instruct sender supplies the same messages to all routes. Brackets give family-bootstrap intervals. Reassigned: success / valid answers, averaged over three edit-norm-matched reassignments. Invalid answers count as errors; without a message, success is zero by construction. Full results: Appendix~\ref{app:receiver_recovery}.\par}
\end{table}

All four receivers reach \RECGLMRate{}--\RECBaseRate{}\% four-answer success
using their own symbol directions. Writing at an early layer improves three
receivers by 66.7--73.8 percentage points over middle-layer writing on the
same messages and scenes.

\textbf{Does the receiver use the selected symbols?}
Pointing each symbol to another symbol's direction keeps message bytes and
actual edit size, yet cuts mean four-answer success to 0.6--3.0\%.
Qwen3-14B-Base keeps about 99\% valid answers, so its drop reflects wrong answers, not formatting; the other receivers also
give more invalid answers. Given the other world's message, receivers answer
according to that world's facts, a consistency check within the same scenes.

The six planned comparisons cover the three receivers that fail with
middle-layer writing; Qwen3-32B, which already succeeds there, serves as a
reference. For each of the three, the early write beats both middle-layer
writing and mean reassignment after Holm correction, with positive lower
bounds under both family and template resampling.

\textbf{How does internal delivery compare with reading the same message?}
On both Qwen receivers and Gemma, internal and text delivery reach similar
success, although this does not establish equivalence. GLM does 13.5 points
better with text. A post-hoc breakdown by domain locates the gap: GLM's
internal success is 91.1\% on blocks, 85.3\% on candy and 46.6\% on coats,
versus 97.4\%, 99.5\% and 66.5\% with text, so coats remain hard for both
routes. Section~\ref{sec:discussion} separates delivery from the sender's
symbol choice by supplying complete facts.

Table~\ref{tab:recovery_native} compares the fitted directions with the
receiver's own states at the early layer on the confirmation scenes.

\begin{table}[!htbp]
\caption{\textbf{Reusable symbol directions and the receiver's own states.}}
\label{tab:recovery_native}
\centering\small
\setlength{\tabcolsep}{3.5pt}\renewcommand{\arraystretch}{1.12}
\begin{tabular*}{\linewidth}{@{\extracolsep{\fill}}lrrrr@{}}
\toprule
Receiver & \textbf{Directions} & Native states & Difference [95\% CI] & Text \\
\midrule
Qwen3-32B & \textbf{99.8} & 100.0 & -0.2 [-0.5, +0.0] & 99.8 \\
Qwen3-14B-Base & \textbf{99.5} & 99.8 & -0.3 [-0.9, +0.0] & 100.0 \\
GLM-4-9B & \textbf{76.5} & 75.6 & +0.9 [-1.2, +3.2] & 93.0 \\
Gemma-3-27B-IT & \textbf{100.0} & 100.0 & +0.0 [+0.0, +0.0] & 100.0 \\
\bottomrule
\end{tabular*}
\par\vspace{3pt}
{\footnotesize\raggedright Complete-fact four-answer success (\%) on 572 confirmation scenes, with no model sender. Native states come from the receiver reading each scene\textquotesingle s facts; both routes write at the same early layer and fact positions. Differences are directions minus native states (percentage points), with paired family-bootstrap intervals. Text reads the same facts. Closeness does not establish equivalence.\par}
\end{table}

\textbf{How much fitting is needed?}
Table~\ref{tab:wordif_representations_main} compares representations using
complete facts on a separate 576-scene bank. Two of three prespecified single
contexts approach the twelve-context result; the first is worse on Qwen3-32B
and fails on GLM. Averaging is unnecessary for high success, but context
choice matters. The simplest alternative adds the symbol-minus-filler input-embedding
difference at the input layer. At unit strength this equals reading the
symbol as text, as every symbol in this bank is a single token; it matches the fitted directions on three
receivers and exceeds them on GLM (Appendix~\ref{app:wordif_representations}).

\begin{table}[!t]
\caption{\textbf{Local symbol directions and simpler representations.}}
\label{tab:wordif_representations_main}
\centering\normalsize
\setlength{\tabcolsep}{3pt}\renewcommand{\arraystretch}{1.12}
\begin{tabular*}{\linewidth}{@{\extracolsep{\fill}}lrrrrr@{}}
\toprule
Receiver & \shortstack[r]{Twelve\\contexts} & \multicolumn{3}{c}{One fitting context} & \shortstack[r]{Embedding\\(= text)} \\\cmidrule(lr){3-5}
 & & First & Middle & Last & \\
\midrule
Qwen3-32B & 98.8 & 94.6 & 98.8 & 98.6 & 99.3 \\
Qwen3-14B-Base & 99.1 & 97.2 & 99.8 & 99.0 & 100.0 \\
GLM-4-9B & 76.4 & 8.9 & 75.5 & 76.7 & 92.7 \\
Gemma-3-27B-IT & 100.0 & 99.8 & 100.0 & 100.0 & 100.0 \\
\bottomrule
\end{tabular*}
\par\vspace{3pt}
{\footnotesize\raggedright Complete-fact four-answer success (\%) on 576 new scenes. Direction strengths use the common development grid: $\gamma=1$ except GLM First ($\gamma=4$). Unit-strength input embeddings equal text for these single-token symbols. Intervals and automatic messages: Appendix~\ref{app:wordif_representations}.\par}
\end{table}

\FloatBarrier

\textbf{What the depth comparisons establish.}
Single-layer writes rule out repeated writing as the cause of late-layer
failure, yet native states and symbol directions can differ: on GLM at depth
0.35 they reach 66.7\% and 8.3\%. At depth 0.2, matching the size of the two
edits changes success by at most about one point, leaving the role of edit
size at failed depths open.

For Gemma, a rule fixed before any fitted-direction result on
this receiver, which chose the depth from native states alone, also reaches
100\% development success (Appendix~\ref{app:gemma_depth_rule}).

\section{Why late writing fails: diagnostics}
\label{app:diag}

Both diagnostics use the 92 development scenes with complete facts, the
existing twelve-context directions and $\gamma=1$, and were fixed before any
output. They localize where facts must be available; they do not identify a
unique mechanism.

\paragraph{Where written facts must be readable.}
Three conditions compare depths 0.35, 0.5 and 0.65 with the early write at
0.2: a single write at the cut depth; and the early write followed by
replacing the prefix keys and values of every layer up to the cut with those
of the neutral prefix, while later layers keep the written values. Built-in
checks pass on all four receivers (Table~\ref{tab:diag_d1}). Under the
prespecified rule, a drop of at least 20 points from the early write with a
positive lower limit means that answers need to read the facts in layers at
or below the cut. This holds at 0.5 and 0.65 for Qwen3-14B-Base, GLM and
Gemma, and at 0.65 for Qwen3-32B; native states behave the same way. At 0.35,
restoring costs at most 2.1 points on the Qwen receivers and Gemma. On GLM,
the early write followed by restoring up to 0.35 reaches 72.9\% while writing
at 0.35 reaches 8.3\%, so the layers between 0.2 and 0.35 also transform the
written directions into a usable form.

\begin{table}[!htbp]
\caption{\textbf{Where written facts must be readable.}}
\label{tab:diag_d1}
\centering\small
\setlength{\tabcolsep}{3pt}\renewcommand{\arraystretch}{1.1}
\begin{tabular*}{\linewidth}{@{\extracolsep{\fill}}llrrrrrrr@{}}
\toprule
 & & Early write & \multicolumn{2}{c}{Cut at 0.35} & \multicolumn{2}{c}{Cut at 0.5} & \multicolumn{2}{c}{Cut at 0.65} \\\cmidrule(lr){4-5}\cmidrule(lr){6-7}\cmidrule(l){8-9}
Receiver & Source & (0.2) & Write & \textbf{Restore} & Write & \textbf{Restore} & Write & \textbf{Restore} \\
\midrule
Qwen3-14B-Base & Fitted & 100.0 & 87.5 & \textbf{97.9} & 6.2 & \textbf{59.4} & 1.0 & \textbf{0.0} \\
 & Native & 100.0 & 96.9 & \textbf{97.9} & 22.9 & \textbf{52.1} & 0.0 & \textbf{0.0} \\
\addlinespace[2pt]
GLM-4-9B & Fitted & 88.5 & 8.3 & \textbf{72.9} & 0.0 & \textbf{0.0} & 0.0 & \textbf{0.0} \\
 & Native & 90.6 & 66.7 & \textbf{75.0} & 0.0 & \textbf{0.0} & 0.0 & \textbf{0.0} \\
\addlinespace[2pt]
Gemma-3-27B-IT & Fitted & 100.0 & 85.4 & \textbf{97.9} & 0.0 & \textbf{2.1} & 0.0 & \textbf{0.0} \\
 & Native & 100.0 & 96.9 & \textbf{97.9} & 0.0 & \textbf{2.1} & 0.0 & \textbf{1.0} \\
\addlinespace[2pt]
Qwen3-32B & Fitted & 100.0 & 100.0 & \textbf{100.0} & 96.9 & \textbf{100.0} & 1.0 & \textbf{6.2} \\
 & Native & 100.0 & 100.0 & \textbf{100.0} & 97.9 & \textbf{100.0} & 2.1 & \textbf{7.3} \\
\bottomrule
\end{tabular*}
\par\vspace{6pt}
{\footnotesize\setlength{\tabcolsep}{3pt}
\begin{tabular*}{\linewidth}{@{\extracolsep{\fill}}llrr@{}}
\toprule
Receiver & Source & Early $-$ restore, cut 0.5 & Early $-$ restore, cut 0.65 \\
\midrule
Qwen3-14B-Base & Fitted & +40.6 [29.2, 52.1] & +100.0 [100.0, 100.0] \\
 & Native & +47.9 [31.2, 63.5] & +100.0 [100.0, 100.0] \\
\addlinespace[2pt]
GLM-4-9B & Fitted & +88.5 [79.2, 95.8] & +88.5 [79.2, 95.8] \\
 & Native & +90.6 [81.2, 97.9] & +90.6 [81.2, 97.9] \\
\addlinespace[2pt]
Gemma-3-27B-IT & Fitted & +97.9 [94.8, 100.0] & +100.0 [100.0, 100.0] \\
 & Native & +97.9 [94.8, 100.0] & +99.0 [96.9, 100.0] \\
\addlinespace[2pt]
Qwen3-32B & Fitted & +0.0 [0.0, 0.0] & +93.8 [87.5, 99.0] \\
 & Native & +0.0 [0.0, 0.0] & +92.7 [87.5, 96.9] \\
\bottomrule
\end{tabular*}}
\par\vspace{3pt}
{\footnotesize\raggedright Complete-fact four-answer success (\%) on 92 development scenes (top) and differences in points with 95\% paired family-bootstrap intervals enclosing three seeds (bottom). Fitted: twelve-context directions; Native: the receiver's own states for the facts. Write: one write at the cut depth. Restore: the early write, after which the prefix keys and values of every layer up to the cut are replaced by those of the neutral prefix. Built-in checks pass on all receivers: restoring up to the write layer leaves the cache identical to the early write, and restoring every layer reproduces the no-message outputs.\par}
\end{table}

\paragraph{Where the gap to the receiver's own states arises.}
On GLM at depth 0.35, fitted directions reach 8.3\% and the receiver's own
states 66.7\%. A scene-matched direction, computed from each scene's own
neutral prompt with only one fact in its slot, closes about half of that gap;
the other facts present in the native states account for most of the rest
(Table~\ref{tab:diag_d2}).

\begin{table}[!htbp]
\caption{\textbf{GLM-4-9B: where the gap to the receiver's own states arises.}}
\label{tab:diag_d2}
\centering\footnotesize
\setlength{\tabcolsep}{2.5pt}
\begin{tabular*}{\linewidth}{@{\extracolsep{\fill}}lrrrrrrr@{}}
\toprule
Depth & Fitted & Scene-matched & Native & Text & Matched $-$ fitted & Native $-$ matched & Text $-$ native \\
\midrule
0.2 & 88.5 & 87.5 & 90.6 & 99.0 & $-$1.0 [-5.2, 3.1] & +3.1 [-2.1, 8.3] & +8.3 [1.0, 17.7] \\
0.35 & 8.3 & 40.6 & 66.7 & 99.0 & +32.3 [24.0, 39.6] & +26.0 [11.5, 39.6] & +32.3 [20.8, 45.8] \\
0.5 & 0.0 & 0.0 & 0.0 & 99.0 & +0.0 [0.0, 0.0] & +0.0 [0.0, 0.0] & +99.0 [96.9, 100.0] \\
\bottomrule
\end{tabular*}
\par\vspace{3pt}
{\footnotesize\raggedright Complete-fact four-answer success (\%) on 92 development scenes. Scene-matched: a direction computed from the scene's own neutral prompt with only that fact's symbol in its slot (a diagnostic, not a reusable method). Differences in points with enclosing 95\% intervals.\par}
\end{table}

\FloatBarrier

\section{Final interface: implementation and reproduction}
\label{app:final_interface}

This section specifies the prompted sender and receiver-local symbol directions
used in Tables~\ref{tab:recovery_main}--\ref{tab:recovery_four}.
Appendix~\ref{app:receiver_recovery} gives development, confirmation and
all control outcomes. Our earlier interface, which extracted symbols from the sender's hidden states,
used a different dictionary construction and update rule
(Appendix~\ref{app:protocol}).

\subsection{Sender, inventory and messages}
\label{app:final_sender}
The sender is Llama-3.1-8B-Instruct, with its native chat template, BF16
inference, greedy decoding and at most 256 new tokens. It receives the
ordered 221-symbol inventory, object names and one world's descriptions.
It is instructed to choose, for each object, up to four inventory symbols that
the description explicitly supports, spell them exactly as listed, invent no facts, and return
only a JSON object mapping names to symbol lists. The sender receives neither
question, answer nor counterfactual world. The four-object extension changes
the number of names and JSON fields, retaining the selection rule.

The parser reads the first JSON object. Missing or non-list fields become
empty; list entries are stripped and lowercased, non-inventory entries are
removed, duplicates are removed, and the first four valid symbols are retained.
There is no retry. The selected indices are sorted within each object and written into the
message with their object assignment and an 8-bit weight of 255 each. These
indices refer to inventory symbols, independently of either model's tokenizer. In the
internal route the receiver never sees the sender's generated text; it
receives only the indices. The text route spells out the same selected symbols
in the public message format below. The two routes carry the same content,
not the same number of bytes.
Table~\ref{tab:current_inventory} gives the complete ordered inventory;
Appendix~\ref{app:literal_prompts} gives the literal prompts.

\subsection{The twelve fitting contexts and public layout}
\label{app:final_fitting}
The message begins with \emph{Selected words and exact relative weights
(integers divided by 255):}. Each object then contributes a field such as
\begin{quote}\small\ttfamily\raggedright
XOPA: unknown [255/255], unknown [255/255],\newline
unknown [255/255], unknown [255/255]
\end{quote}
Fields are separated by a semicolon and the message ends with a period.
For fitting, the fixed object names are XOPA and YUME. With two symbol positions
per object, the target symbol occupies each of the four possible positions in
turn; with four per object, it occupies each of eight positions. All other
positions contain \texttt{unknown}. These $4+8=12$ public-layout contexts
contain no evaluated scene facts or questions. The directions are therefore
reusable across scenes, and the fitting contexts already use the message
format of the evaluation.

For each context we tokenize the actual prefix, locate every token overlapping
the target symbol, and record the output of the chosen layer at the symbol's final token. The neutral comparison replaces every token in that span by the
single token for \texttt{ unknown}, preserving the prefix length. We subtract
the neutral state, average the twelve contrasts in FP64 and store the result
in FP32. Directions are not rescaled in the selected configuration.
The same process constructs separate directions at every tested layer.
It uses only receiver activations, with no sender states, paired activations,
evaluation questions, target answers or parameter updates. A symbol split into several tokens still yields a single direction, taken at
its final token and written at one position.

\subsection{Intervention, cache reuse and reference conditions}
\label{app:final_writer}
The receiver instructions request exactly one object name on one line.
Two fixed demonstrations use DEMOX/DEMOY with large/small and smooth/rough,
answering ``Which object is large?'' and ``Which object is rough?''.
These same demonstrations precede both fitting and evaluation prefixes.
The neutral receiver prefix contains four symbol positions per object for both
automatic and complete-fact messages. It includes the shared instructions,
message fields and object declarations; the question is appended later.
The neutral prefix depends on object names and public layout, not on either
world's facts. Chat models use their native user-message template with thinking
disabled; the Base model uses the plain prompt. Its hidden states are captured
once. A second prefix forward
replaces each active position by its saved neutral state plus $\gamma d_w$;
unused positions receive no update. At depth 0.2, the selected zero-based
block indices are 13, 8, 8 and 12 for Qwen3-32B, Qwen3-14B-Base, GLM-4-9B
and Gemma-3-27B-IT, respectively. All use $\gamma=1$.

The write changes the output of the chosen layer at the active positions.
Later layers process the changed prefix; the cache of earlier layers still
comes from the neutral input. Each question receives
its own copy of the completed cache, with hooks removed and no further writes.
The no-message and active identity-write computations must agree exactly.
Before evaluation, engineering checks compare logits and parsed answers
between the cached computation and a complete forward pass; the scenes used
for these checks are excluded from reported estimates.
This implementation does not replace every layer's cache with one computed
from a full textual message.

For the native-state reference, the receiver reads the supplied symbols in the
same public object format. We copy each symbol's final-token states into the
matching object and slot. Textual symbols can span several tokens, so
source and neutral prefixes can have different lengths; alignment is semantic
by field, not an assumption of equal absolute token indices. Symbol--direction reassignments keep the message bytes and positions, and
rescale the replacement directions to match the actual edit norm across
active positions at each layer.
They can affect both answer correctness and output format; both are reported.

\subsection{Literal prompts and inference settings}
\label{app:literal_prompts}
The following templates are instantiated independently for each world.
\texttt{\{vocab\}} is the comma-separated inventory in
Table~\ref{tab:current_inventory}, \texttt{\{n\}} is 221, and
\texttt{\{description\}} joins that world's object descriptions with single spaces.
The braces around alias names mark substitutions, not additional message fields.

\paragraph{Sender system message.}
\begin{quote}\small\ttfamily\raggedright
You compress object descriptions into a fixed public vocabulary. You will be given a vocabulary of words and a description of two objects. For each object, choose up to four vocabulary words that the description explicitly supports. Use only words from the vocabulary, spelled exactly as given. Do not invent facts. Answer with one JSON object and nothing else.
\end{quote}
\paragraph{Sender user message.}
\begin{quote}\small\ttfamily\raggedright
Vocabulary (\{n\} words, in order):\par
\{vocab\}\par
\strut\par
Objects: \{alias0\}, \{alias1\}\par
Description:\par
\{description\}\par
\strut\par
Reply with exactly this JSON shape, listing up to four vocabulary words per object that the description explicitly supports (fewer is fine, an empty list if nothing applies):\par
\{"\{alias0\}": ["word", "word"], "\{alias1\}": ["word"]\}
\end{quote}

For four objects the system message changes ``two objects'' to ``four objects''.
The user message lists four aliases and adds the corresponding third and fourth
JSON fields, each illustrated by \texttt{["word"]}. The symbol-selection
instruction, inventory and generation limit are unchanged.

\paragraph{Receiver body before the native chat wrapper.}
\begin{quote}\small\ttfamily\raggedright
Answer with exactly one object name on one line.\par
DEMOX is large. DEMOY is small.\par
Q: Which object is large?\par
A: DEMOX\par
\strut\par
DEMOX is smooth. DEMOY is rough.\par
Q: Which object is rough?\par
A: DEMOY\par
\strut\par
Message:\par
\{message\}\par
Objects:\par
\{object\_declarations\}\par
Q: \{question\}\par
A:
\end{quote}

In this body, \texttt{\{object\_declarations\}} concatenates
\texttt{\{alias\} is an object.} for every alias. The message uses the exact
header and fields in Appendix~\ref{app:final_fitting}. Internal delivery
fills these fields with \texttt{unknown}; text delivery spells out each
selected symbol and retains its \texttt{[255/255]} weight. Missing symbols leave
the remaining positions neutral. The question is identical across worlds.

Receiver inference uses FP32, eager attention, batch size one and disabled
TF32. Generation is greedy with a primary budget of 64 new tokens and a
256-token sensitivity run. The GLM checkpoint is GLM-4-9B-0414, abbreviated
GLM-4-9B in the tables. Sender inference uses BF16 and the native
Llama-3.1-8B-Instruct chat template; receiver chat wrappers and layer indices
are specified in Appendix~\ref{app:final_writer}.

\subsection{Reproduction sequence}
\label{app:reproduction_sequence}
The dictionary is computed once for a receiver and reused for every message.
The sender does not need the receiver checkpoint or its hidden-state dimension.
The online computation is:
\begin{enumerate}\setlength{\itemsep}{2pt}
\item Parse the sender's selections and serialize sorted symbol indices with
      object assignments and equal weights. Reuse the message across questions.
\item Render the neutral prefix and save its block-output states at the four
      placeholder positions per object.
\item At the selected block, replace each populated position by its saved
      neutral state plus the scaled receiver-local symbol direction.
\item Complete the prefix forward pass, remove hooks, and copy the resulting
      cache separately for each question. Generate without further writes.
\item Parse and score all four answers. Run text and control routes with the same message and questions.
\end{enumerate}
The identity-write check, multi-token rule and field alignment in
Appendices~\ref{app:final_fitting}--\ref{app:final_writer} apply throughout.

\begin{table}[!htbp]
\caption{\textbf{The shared inventory in index order.}}
\label{tab:current_inventory}
\centering\small
\setlength{\tabcolsep}{3pt}\renewcommand{\arraystretch}{1.08}
\begin{tabular*}{\linewidth}{@{\extracolsep{\fill}}lp{0.84\linewidth}@{}}
\toprule
Indices & Symbols \\
\midrule
0--11 & accordion, ambulance, axe, bakery, banana, barn, bed, beer, beetle, belt, bench, bicycle \\
12--23 & blender, boat, boot, bowl, broccoli, broth, buffalo, bus, butterfly, cabbage, cabinet, camel \\
24--35 & car, carpet, carrot, celery, cherry, chicken, church, cider, clam, clamp, coat, cocoa \\
36--47 & coconut, cod, copper, cotton, cow, crab, crow, cucumber, curtain, deer, desk, dolphin \\
48--59 & drawer, drill, elephant, espresso, factory, flea, flute, fork, fox, fridge, garlic, glass \\
60--71 & glove, goat, gold, goose, grape, greenhouse, hat, hawk, helicopter, hoe, hospital, hotel \\
72--83 & iron, jacket, jar, juice, kettle, leather, leopard, lettuce, library, lime, lobster, mango \\
84--95 & mattress, mosquito, moth, mug, nail, needle, olive, onion, oven, pan, panda, peach \\
96--107 & pear, pharmacy, pig, pigeon, pillow, pineapple, plane, plastic, plate, plum, potato, pumpkin \\
108--119 & rake, rat, rubber, salmon, scarf, school, scissors, sheep, shelf, ship, shirt, shoe \\
120--131 & shovel, shrimp, silk, silver, skirt, sock, soda, sofa, spinach, spoon, squid, squirrel \\
132--143 & stadium, steel, stool, stove, strawberry, submarine, sweater, theater, toaster, tomato, tractor, train \\
144--155 & truck, trumpet, tuna, turtle, violin, wardrobe, whisk, wool, wrench, alive, bitter, climb \\
156--167 & crawl, edible, fly, furry, green, hop, hot, hunt, metallic, orange, red, run \\
168--179 & salty, shell, sour, sweet, swim, transparent, walk, wooden, worn, yellow, above, after \\
180--191 & before, behind, beside, bite, block, carry, chase, cover, eat, enter, fear, follow \\
192--203 & guard, hit, hold, inside, near, own, protect, pull, push, touch, under, watch \\
204--215 & all, always, both, either, every, if, maybe, most, must, neither, never, none \\
216--220 & not, only, or, some, unless \\
\bottomrule
\end{tabular*}
\par\vspace{3pt}
{\footnotesize\raggedright Indices start at zero; symbols within each row continue from left to right. The sender may choose any inventory symbol. The task-specific attribute subset is listed in Table~\ref{tab:current_task_values}. Inventory indices are independent of tokenizer token IDs.\par}
\end{table}

\FloatBarrier
\section{Task construction and evaluation units}
\label{app:current_tasks}

\subsection{Two worlds that differ in one fact assignment}
\label{app:current_task_design}
Each scene has two worlds describing the same two objects of one type, each
with two attributes (Table~\ref{tab:current_task_values}). World 0 is the
original; world 1 exchanges one attribute's values between the objects and
keeps the other attribute fixed. One question asks
about the exchanged attribute; the other asks about the unchanged attribute.
The correct answer to the first must change between worlds and the second must
stay fixed. Both answers are unique. This design prevents a receiver from
succeeding by repeating one object's name or by changing every answer whenever
the message changes.

\begin{table}[!htbp]
\caption{\textbf{Attribute space of the two-object confirmation.}}
\label{tab:current_task_values}
\centering\small
\setlength{\tabcolsep}{3pt}\renewcommand{\arraystretch}{1.08}
\begin{tabular*}{\linewidth}{@{\extracolsep{\fill}}lll@{}}
\toprule
Objects & Attribute & Values \\
\midrule
Painted blocks & Color & green, orange, red, yellow \\
 & Material & gold, iron, silver, steel \\
Candy & Color & green, orange, red, yellow \\
 & Taste & bitter, salty, sour, sweet \\
Coats & Fibre & cotton, leather, silk, wool \\
 & Buttons & copper, glass, plastic, wooden \\
\bottomrule
\end{tabular*}
\par\vspace{3pt}
{\footnotesize\raggedright Each domain contains 192 generated scenes before engineering exclusions. Each object has exactly one value of each of its two attributes. Coat fibre and button material are distinct attributes; block paint and body material are distinct.\par}
\end{table}

Each domain uses four evidence templates and twelve cells, each a pair of
values for the exchanged attribute. Three cells are assigned to each template, giving twelve material groups per
domain and 36 overall. Within each group, sixteen scenes balance the exchanged
attribute, each attribute's object assignment, and which candidate value is
queried: $2\times2\times2\times2=16$. Object names and their order are assigned
from a new pool of 24 aliases. Thus the generated bank contains
$36\times16=576$ scenes. Four implementation-check scenes are fixed before receiver evaluation and
excluded, leaving 572 scenes for reported estimates.

For blocks, the questions ask which object has a specified surface paint color
or is made of a specified material. For candy, they ask about color or taste.
For coats, they ask about fibre or button material. The changed-attribute
question is scored first, regardless of which attribute changes. The sender
receives only the two description sentences from one world. The receiver
receives names, the public message structure and one question, with no original
descriptions in the internal route.

\paragraph{A complete example.}
In world 0 the sender reads ``Workshop tag on BANQ: sprayed green over a body cast in gold.'' and ``Workshop tag on CIRV: sprayed red over a body cast in steel.''.
World 1 exchanges the paint colors and keeps the body materials fixed. Table~\ref{tab:current_example} gives the resulting messages. The receiver questions are ``Which object has green surface paint?'' and ``Which object is made of gold?''. The required answer pairs are BANQ/BANQ in world 0 and CIRV/BANQ in world 1. These are task labels, not generated receiver responses.

\begin{table}[!htbp]
\caption{\textbf{One scene from descriptions to symbol indices.}}
\label{tab:current_example}
\centering\small
\setlength{\tabcolsep}{3pt}\renewcommand{\arraystretch}{1.08}
\begin{tabular*}{\linewidth}{@{\extracolsep{\fill}}lllp{0.36\linewidth}@{}}
\toprule
World & Object & Described facts & Transmitted indices (decoded here) \\
\midrule
0 & BANQ & green, gold & 62 (gold), 160 (green) \\
 & CIRV & red, steel & 133 (steel), 166 (red) \\
1 & BANQ & red, gold & 62 (gold), 166 (red) \\
 & CIRV & green, steel & 133 (steel), 160 (green) \\
\bottomrule
\end{tabular*}
\par\vspace{3pt}
{\footnotesize\raggedright Actual sender selections for a generated confirmation scene, selected by scene ID without inspecting receiver outcomes. All selected weights are 255; other positions stay neutral. Symbol spellings in parentheses explain the packet and are not inserted as text in the internal route.\par}
\end{table}

\subsection{Development, confirmation and overlap}
\label{app:current_task_split}
The 96-scene development bank had already been evaluated during interface
design. Four scenes check implementation, and 92 determine the receiver
configuration. A later rescreen of coat materials also reused previously evaluated scenes. Neither set is reported as independent confirmation.
The final 576-scene bank is constructed and fixed before the receiver outputs
used in the main tables. Table~\ref{tab:current_overlap} specifies both what
changes and what is retained from earlier banks.

\begin{table}[!htbp]
\caption{\textbf{New material and retained task structure.}}
\label{tab:current_overlap}
\centering\small
\setlength{\tabcolsep}{3pt}\renewcommand{\arraystretch}{1.08}
\begin{tabular*}{\linewidth}{@{\extracolsep{\fill}}lrr@{}}
\toprule
Unit & In confirmation & Also in earlier banks \\
\midrule
Object names & 24 & 0 \\
Full scenes & 576 & 0 \\
Evidence templates & 12 & 0 \\
Single-object fact tuples & 48 & 48 \\
Two-object combinations & 72 & 14 \\
Paired-value cells & 36 & 7 \\
Question forms & 6 & 6 \\
\bottomrule
\end{tabular*}
\par\vspace{3pt}
{\footnotesize\raggedright Counts describe the entire generated 576-scene bank. A fact tuple retains attribute names; a value cell specifies the two candidate values for each attribute. Question forms abstract over the queried value. The reference contains 1,920 earlier scenes. No new-symbol or new-question-form generalization is claimed.\par}
\end{table}

New names and evidence templates test use of familiar attribute symbols in new
descriptions. The confirmation retains the finite symbol inventory and question
forms, and every single-object fact tuple has appeared before. The two worlds, their two questions and the longer decoding budget are
repeated measurements of one scene, never separate examples.
The four-object extension changes the number of objects and uses four
templates; its construction and selected receiver scope are reported in
Appendix~\ref{app:recovery_four}.

\paragraph{Answer scoring.}
The parser takes the first nonempty line, strips surrounding whitespace and
punctuation, and matches it to a known object alias without case sensitivity.
It does not infer an answer from an explanation or take the most likely alias
from logits. An unmatched answer is invalid and incorrect. Single-answer
accuracy uses all answers; validity measures successful alias parsing regardless
of correctness. Changed- and unchanged-question success each require that
question to be correct in both worlds. Joint success, called four-answer success in the main text, requires all four
answers.
An identical deterministic no-message input cannot satisfy the changed question
in both worlds, so its joint score is necessarily zero even if individual
answers are sometimes correct.

\FloatBarrier
\section{Final-interface confirmation and receiver diagnostics}
\label{app:final_evidence}
\subsection{Write-setting selection and automatic confirmation}
\label{app:receiver_recovery}

\paragraph{Local fitting and development.}
We keep the 221-symbol inventory, the prompted Llama-3.1-8B-Instruct sender,
the strict parser and the twelve symbol-minus-unknown fitting contexts, and fit
directions at every candidate layer (FP32, eager attention, TF32 disabled).
The refitted middle-layer directions are checked against the earlier fit.
All message sources, including complete facts, now use four public fact
positions per object, unlike the earlier two-position complete-fact study. No scene, answer or paired sender--receiver activation enters fitting.

The previously evaluated 96-scene development bank supplies four implementation-check scenes and 92 evaluation scenes.
Candidates are the original three-layer middle band and single layers at relative depths
$\{0.2,0.35,0.5,0.65,0.8\}$, each with strengths
$\{0.25,0.5,1,2,4\}$. The native-state reference is tested at the same locations: it copies the
receiver's states from a prefix containing the complete facts into the
corresponding neutral fact positions. It is a diagnostic built from
scene-specific states, not the reusable dictionary method.

A receiver-local scale multiplies each direction by the median ratio
$c_\ell=\mathrm{median}_{w,t}\|\delta_{\ell,w,t}\|/\|d_{\ell,w}\|$
between single-context and averaged contrast norms. At the depth selected
with native states, we test the same strength grid with this scale and, as
diagnostics, match the realized edit norms of fitted and native writes in
both directions. The final selection maximizes joint success among settings
with at least 90\% format coverage; ties are broken by unchanged-question
correctness, then fewer layers, raw scale, smaller strength and the listed
depth order. All four receivers select raw scaling, depth 0.2 and
$\gamma=1$; the local scale never yields a better configuration. The
selected layer indices are 13, 8, 8 and 12 (zero-based) for
Qwen3-32B, Qwen3-14B-Base, GLM-4-9B and Gemma-3-27B-IT.

For Gemma, a rule that picks the depth from native states alone, with local
scaling and strength one, was fixed before any fitted-direction outcome was
read; it reaches 100\% development success and differs from the raw-scale
setting selected afterwards (Appendix~\ref{app:gemma_depth_rule}). A single
rescreen on previously evaluated coat scenes, without retuning, gives 100\%
joint success on Qwen and Gemma and 47.9\% on GLM with 79.4\% coverage, below
both recovery thresholds defined below. All four receivers nevertheless
remain in the new confirmation.

\paragraph{Confirmation material and information separation.}
The bank in Appendix~\ref{app:current_tasks} has 576 generated scenes and
572 evaluated scenes (2,288 answers per condition), with 36 material groups
nested in twelve evidence templates. Its overlap and implementation-check
exclusions are specified there. Messages are fixed across receivers and
delivery routes. Both
internal and weighted-text routes use the same four-position layout; text
spells out the selected symbols while unused positions stay neutral. The bank
and analysis are fixed before receiver outputs, with no configuration
selection from confirmation performance.

\paragraph{Controls and inference.}
We test no message, an identity write, the original middle band, the
selected early layer, the same message as text, three symbol--direction
reassignments, and the other world's message. Reassignments keep positions
and weights and match the actual per-layer Frobenius norm over all active
positions, but not the geometry at each position. Complete facts are also
delivered as message text, as the full original evidence text, and through
native states.

Joint success, accuracy, coverage and question-specific success use every
scene, including malformed outputs. Estimates average family means equally.
Because four implementation-check scenes are excluded, the family-macro estimate
can differ slightly from the raw success fraction. Raw selected-write
counts are 555/572, 559/572, 425/572 and 559/572 for Qwen3-32B,
Qwen3-14B-Base, GLM and Gemma. The 64- and 256-token joint results agree.

Intervals use 20,000 cluster-bootstrap draws for each of three fixed seeds
and report the 95\% interval enclosing all three, computed separately over
the 36 families and the twelve parent templates. The six Holm-corrected tests
compare the early write with the middle band and with the mean reassignment
on Qwen3-14B-Base, GLM and Gemma. We declared in advance that a receiver
recovers if it reaches at least 50\% joint success and 90\% coverage and
beats reassignment with positive interval lower limits and corrected
significance; it recovers substantially if it also gains twenty points over
the middle band under the same inference. All three receivers meet both
criteria on this bank. Qwen3-32B, which already succeeded with middle-layer
writing, is kept as a reference outside these tests. Text contrasts
are descriptive and do not establish equivalence.

\begin{table}[!htbp]
\caption{\textbf{Answer validity and selective use in automatic confirmation.}}
\label{tab:recovery_controls}
\centering\small
\setlength{\tabcolsep}{3pt}\renewcommand{\arraystretch}{1.08}
\begin{tabular*}{\linewidth}{@{\extracolsep{\fill}}llrrrrr@{}}
\toprule
Receiver & Delivery & Joint & Accuracy & Valid & Changed & Unchanged \\
\midrule
Qwen3-32B & Early write & 97.0 & 99.1 & 99.6 & 97.9 & 98.8 \\
 & Same packet as text & 97.0 & 99.1 & 99.4 & 97.9 & 98.8 \\
 & Reassignment 1 & 1.9 & 31.7 & 59.5 & 8.6 & 20.3 \\
 & Reassignment 2 & 0.9 & 31.3 & 62.5 & 5.0 & 20.1 \\
 & Reassignment 3 & 0.7 & 32.5 & 64.5 & 5.1 & 22.1 \\
Qwen3-14B-Base & Early write & 97.7 & 99.3 & 100.0 & 98.1 & 99.3 \\
 & Same packet as text & 97.7 & 99.3 & 99.9 & 98.1 & 99.3 \\
 & Reassignment 1 & 2.5 & 49.0 & 99.3 & 7.5 & 40.1 \\
 & Reassignment 2 & 2.1 & 48.8 & 99.4 & 5.4 & 44.1 \\
 & Reassignment 3 & 2.4 & 49.0 & 99.0 & 6.3 & 40.9 \\
GLM-4-9B & Early write & 74.3 & 90.1 & 92.3 & 84.3 & 84.6 \\
 & Same packet as text & 87.8 & 95.2 & 95.5 & 91.1 & 93.0 \\
 & Reassignment 1 & 1.1 & 24.6 & 48.8 & 2.4 & 17.3 \\
 & Reassignment 2 & 0.5 & 19.9 & 40.5 & 1.1 & 13.9 \\
 & Reassignment 3 & 0.3 & 28.5 & 60.0 & 0.9 & 22.7 \\
Gemma-3-27B-IT & Early write & 97.7 & 99.3 & 100.0 & 98.1 & 99.3 \\
 & Same packet as text & 97.9 & 99.4 & 100.0 & 98.3 & 99.3 \\
 & Reassignment 1 & 1.9 & 43.2 & 85.2 & 9.1 & 27.8 \\
 & Reassignment 2 & 4.7 & 45.3 & 87.9 & 13.4 & 33.5 \\
 & Reassignment 3 & 2.5 & 46.8 & 96.3 & 10.2 & 34.4 \\
\bottomrule
\end{tabular*}
\par\vspace{3pt}
{\footnotesize\raggedright Percentages are family-macro averages on 572 scenes. Accuracy scores individual answers; two-choice chance is 50\%. Valid is the fraction of answers following the required format, regardless of correctness. Changed and unchanged each require the respective question to be correct in both worlds. Invalid generations remain errors. The 256-token sensitivity does not change joint success.\par}
\end{table}

\begin{table}[!htbp]
\caption{\textbf{Paired gains under both clustering levels.}}
\label{tab:recovery_contrasts}
\centering\small
\setlength{\tabcolsep}{3pt}\renewcommand{\arraystretch}{1.08}
\begin{tabular*}{\linewidth}{@{\extracolsep{\fill}}llrrr@{}}
\toprule
Receiver & Contrast & Gain (pp) & 36-family interval & 12-template interval \\
\midrule
Qwen3-32B & Early minus middle & 2.6 & [1.2, 4.2] & [1.6, 3.7] \\
 & Early minus reassigned & 95.9 & [92.1, 98.4] & [90.8, 98.9] \\
 & Early minus text & 0.0 & [-0.5, 0.5] & [-0.5, 0.5] \\
Qwen3-14B-Base & Early minus middle & 66.7 & [63.1, 70.2] & [63.0, 70.6] \\
 & Early minus reassigned & 95.4 & [92.8, 97.6] & [90.9, 98.1] \\
 & Early minus text & 0.0 & [-0.7, 0.9] & [-0.5, 0.5] \\
GLM-4-9B & Early minus middle & 73.8 & [66.5, 80.8] & [62.0, 84.8] \\
 & Early minus reassigned & 73.7 & [66.3, 80.6] & [61.9, 84.6] \\
 & Early minus text & -13.5 & [-18.3, -8.9] & [-20.4, -7.5] \\
Gemma-3-27B-IT & Early minus middle & 67.3 & [61.7, 72.6] & [60.8, 72.7] \\
 & Early minus reassigned & 94.7 & [91.4, 97.2] & [89.6, 97.8] \\
 & Early minus text & -0.2 & [-0.5, 0.0] & [-0.5, 0.0] \\
\bottomrule
\end{tabular*}
\par\vspace{3pt}
{\footnotesize\raggedright 95\% paired bootstrap intervals. Early-minus-middle and early-minus-mean-reassignment on Qwen3-14B-Base, GLM-4-9B and Gemma-3-27B-IT form the six-test Holm family; all six reject. Text contrasts are descriptive; a small or zero difference does not establish equivalence.\par}
\end{table}

\begin{table}[!htbp]
\caption{\textbf{Complete facts and contextual states at the selected depth.}}
\label{tab:recovery_capacity}
\centering\small
\setlength{\tabcolsep}{3pt}\renewcommand{\arraystretch}{1.08}
\begin{tabular*}{\linewidth}{@{\extracolsep{\fill}}lrrrrr@{}}
\toprule
Receiver & Middle write & Early write & Native states & Packet text & Evidence text \\
\midrule
Qwen3-32B & 97.7 & 99.8 & 100.0 & 99.8 & 99.5 \\
Qwen3-14B-Base & 27.1 & 99.5 & 99.8 & 100.0 & 100.0 \\
GLM-4-9B & 0.0 & 76.5 & 75.6 & 93.0 & 99.5 \\
Gemma-3-27B-IT & 26.7 & 100.0 & 100.0 & 100.0 & 100.0 \\
\bottomrule
\end{tabular*}
\par\vspace{3pt}
{\footnotesize\raggedright Four-answer success (\%) on the same 572-scene bank. All conditions supply complete facts without a model sender. Native states are the receiver\textquotesingle s own contextual fact-token states, transplanted at the selected early layer. All conditions use four public positions per object.\par}
\end{table}

\begin{table}[!htbp]
\caption{\textbf{Remaining automatic-message control conditions.}}
\label{tab:current_additional_controls}
\centering\small
\setlength{\tabcolsep}{3pt}\renewcommand{\arraystretch}{1.08}
\begin{tabular*}{\linewidth}{@{\extracolsep{\fill}}llrrrrr@{}}
\toprule
Receiver & Condition & Joint & Accuracy & Valid & Changed & Unchanged \\
\midrule
Qwen3-32B & No message & 0.0 & 47.3 & 95.5 & 0.0 & 46.7 \\
 & Identity write & 0.0 & 47.3 & 95.5 & 0.0 & 46.7 \\
 & Middle write & 94.4 & 98.4 & 99.0 & 96.9 & 97.0 \\
 & Other-world packet & 0.0 & 49.9 & 99.6 & 0.0 & 98.8 \\
Qwen3-14B-Base & No message & 0.0 & 50.8 & 100.0 & 0.0 & 51.6 \\
 & Identity write & 0.0 & 50.8 & 100.0 & 0.0 & 51.6 \\
 & Middle write & 31.0 & 74.4 & 100.0 & 47.2 & 69.3 \\
 & Other-world packet & 0.0 & 50.3 & 100.0 & 0.0 & 99.3 \\
GLM-4-9B & No message & 0.0 & 0.4 & 0.9 & 0.0 & 0.4 \\
 & Identity write & 0.0 & 0.4 & 0.9 & 0.0 & 0.4 \\
 & Middle write & 0.5 & 40.5 & 80.6 & 1.6 & 38.1 \\
 & Other-world packet & 0.0 & 45.5 & 92.3 & 0.0 & 84.6 \\
Gemma-3-27B-IT & No message & 0.0 & 49.7 & 100.0 & 0.0 & 49.4 \\
 & Identity write & 0.0 & 49.7 & 100.0 & 0.0 & 49.4 \\
 & Middle write & 30.5 & 71.1 & 100.0 & 43.6 & 64.7 \\
 & Other-world packet & 0.0 & 50.3 & 100.0 & 0.0 & 99.3 \\
\bottomrule
\end{tabular*}
\par\vspace{3pt}
{\footnotesize\raggedright Percentages use the same family averages and all 572 scenes as Table~\ref{tab:recovery_controls}. Other-world packets are scored here against the current world; supplied-world scoring reproduces the selected-write counts. No-message joint success is structurally zero because the relevant answer changes between worlds with identical receiver input.\par}
\end{table}

Other-world messages reproduce the opposite world's generated answers under
the same questions. Scored against that world's answers, their counts equal
the selected-write counts, so these rows check consistency and do not add independent samples.
All active identity writes reproduce no-message logits and tokens exactly.
Independent audits rederive gold labels and scores, validate message and file hashes, and verify edit norms; separate recounts reproduce estimates
and decisions. GLM's material-specific breakdown in Appendix~\ref{app:development_confirmation} is a
post-evaluation descriptive analysis using ordinary within-domain means.

\subsection{Update magnitude and the native-state reference}
\label{app:current_norm_checks}
Table~\ref{tab:current_edit_norms} reports the actual size of the selected
early-layer update relative to the receiver state at its write positions.
The ratios range from 0.88 to 1.15. These writes are comparable in magnitude
to the states they modify; high success should not be read as evidence that
a small perturbation suffices. Absolute norms also differ markedly across
receiver architectures, so a common numerical strength does not imply a
common vector length.

\begin{table}[!htbp]
\caption{\textbf{Actual update size at the selected early layer.}}
\label{tab:current_edit_norms}
\centering\small
\setlength{\tabcolsep}{3pt}\renewcommand{\arraystretch}{1.08}
\begin{tabular*}{\linewidth}{@{\extracolsep{\fill}}lrrrr@{}}
\toprule
Receiver & Block & Edit norm & Residual norm & Edit/residual \\
\midrule
Qwen3-32B & 13 & 317.6 & 320.7 & 0.99 \\
Qwen3-14B-Base & 8 & 167.8 & 161.2 & 1.04 \\
GLM-4-9B & 8 & 50.6 & 44.0 & 1.15 \\
Gemma-3-27B-IT & 12 & 2186.0 & 2490.3 & 0.88 \\
\bottomrule
\end{tabular*}
\par\vspace{3pt}
{\footnotesize\raggedright Automatic-message confirmation, zero-based blocks. Norms are Frobenius norms over the positions used by the write hook, averaged over recorded prefix executions. The last column averages the per-execution ratios, rather than dividing the two displayed means. Two question branches reuse each world prefix; these records are not independent observations.\par}
\end{table}

\begin{table}[!htbp]
\caption{\textbf{Matching actual edit norms at the early layer.}}
\label{tab:current_norm_matching}
\centering\small
\setlength{\tabcolsep}{3pt}\renewcommand{\arraystretch}{1.08}
\begin{tabular*}{\linewidth}{@{\extracolsep{\fill}}lrrrr@{}}
\toprule
 & \multicolumn{2}{c}{Fitted directions} & \multicolumn{2}{c}{Native states} \\\cmidrule(lr){2-3}\cmidrule(l){4-5} Receiver & Raw & Native-sized & Raw & Fitted-sized \\
\midrule
Qwen3-32B & 100.0 & 100.0 & 100.0 & 100.0 \\
Qwen3-14B-Base & 100.0 & 100.0 & 100.0 & 100.0 \\
GLM-4-9B & 88.5 & 87.5 & 90.6 & 89.6 \\
Gemma-3-27B-IT & 100.0 & 100.0 & 100.0 & 100.0 \\
\bottomrule
\end{tabular*}
\par\vspace{3pt}
{\footnotesize\raggedright Complete-fact joint success (\%) on 92 development scenes, averaged over twelve material groups. At the selected depth, native-sized rescales the fitted edit to the native edit norm for that prefix; fitted-sized performs the reverse comparison. This matches the total update norm, not each position independently. These diagnostic conditions were not selected for confirmation.\par}
\end{table}

The development comparison in Table~\ref{tab:current_norm_matching} rescales
the entire fitted edit to the native edit norm and then performs the reverse
comparison. Qwen and Gemma retain 100\% joint success in all four conditions;
GLM changes by about one percentage point under either rescaling. At the
selected early depth, this magnitude adjustment therefore contributes little
to the remaining native/fitted difference. It does not identify why late
writes fail: depth changes both the representations and the amount of subsequent
computation available to use them. Writing at a late layer also leaves the caches of all earlier layers computed
from the neutral prefix.

\subsection{A depth rule fixed before Gemma's fitted outcomes}
\label{app:gemma_depth_rule}
Gemma provides a limited check of whether native-state delivery can guide
where to write reusable directions. Before reading any fitted-direction
outcomes on this receiver, we choose depth using complete-fact native states
(Table~\ref{tab:current_gemma_rule}), set strength to one, and apply the
receiver-local scale from generic fitting contrasts. The selected layer is 12, at relative depth 0.2; its local scale is 1.028. This fixed rule reaches
100\% joint success on the development bank.

\begin{table}[!htbp]
\caption{\textbf{Gemma depth selection from native states.}}
\label{tab:current_gemma_rule}
\centering\small
\setlength{\tabcolsep}{3pt}\renewcommand{\arraystretch}{1.08}
\begin{tabular*}{\linewidth}{@{\extracolsep{\fill}}lrr@{}}
\toprule
Native-state write depth & Joint success (\%) & Valid answers (\%) \\
\midrule
Middle band & 43.8 & 100.0 \\
0.20 & 100.0 & 100.0 \\
0.35 & 96.9 & 100.0 \\
0.50 & 0.0 & 100.0 \\
0.65 & 0.0 & 100.0 \\
0.80 & 0.0 & 99.5 \\
\bottomrule
\end{tabular*}
\par\vspace{3pt}
{\footnotesize\raggedright Development results available before inspecting any fitted-direction cell on Gemma-3-27B-IT. The rule selects the highest joint success with at least 90\% valid answers; ties prefer fewer layers, then the frozen depth order. It selects depth 0.2. This is development-guided selection.\par}
\end{table}

The confirmation setting uses the later selected raw scale, also at depth
0.2 and strength one. Keeping the two choices distinct matters: the
native-guided result supports development-time location selection without
looking at Gemma's fitted scores, whereas the confirmation tests the final
configuration. The rule uses task outcomes from native-state writes,
so it is not calibration without task data or an independently validated
predictor of receiver performance.

\subsection{Four-object transfer with the receiver fixed}
\label{app:recovery_four}
This initial four-object extension asks single-attribute questions.
The later conjunction study in Appendix~\ref{app:wordif_conjunction}
requires two attributes jointly and retains all four receivers.
The extension contains 192 scenes of four painted blocks with unique
color/material answers, balanced over the six changed-object pairs, queried
values and answer positions. Four implementation-check scenes leave 188 evaluated scenes in 24 cells (four templates crossed with six changed pairs).
Materials are plastic, glass, rubber and copper; only colors are exchanged
between worlds. New aliases and evidence constructions have no overlap
with preceding banks; inventory symbols and question forms remain familiar.

Receiver dictionaries, depth and strength remain fixed. The renderer repeats
the same object block; the sender changes only the number of aliases and
JSON keys. Its model, inventory, decoding, parser, ordering and equal
weights remain fixed, without retries. Qwen3-32B and Gemma, the non-Qwen receiver with the highest rescreen score,
were chosen before any new confirmation outcome.
This is a selected pair of receivers, not a new success-rate sample.
All 384 world messages parse; required-fact recall is 96.6\% and complete per-object recovery 93.2\%, including implementation-check scenes.

The automatic early-write counts are 163/188 and 171/188, respectively.
Their family-macro percentages are 86.9\% and 91.1\%; complete-fact counts
are 183/188 and 188/188. The original middle band gives 80.0\% and 26.2\%.
\begin{table}[!htbp]
\caption{\textbf{The fixed receiver configuration extends to four objects.}}
\label{tab:recovery_four}
\centering\small
\setlength{\tabcolsep}{3pt}\renewcommand{\arraystretch}{1.08}
\begin{tabular*}{\linewidth}{@{\extracolsep{\fill}}lrrr@{}}
\toprule
Receiver & \shortstack[r]{Automatic write\\{[95\% CI]}} & \shortstack[r]{Same-message\\text} & \shortstack[r]{Complete-fact\\write} \\
\midrule
Qwen3-32B & \textbf{86.9}\enspace {\footnotesize [78.0, 94.6]} & 86.9 & 97.3 \\
Gemma-3-27B-IT & \textbf{91.1}\enspace {\footnotesize [83.3, 97.9]} & 91.1 & 100.0 \\
\bottomrule
\end{tabular*}
\par\vspace{3pt}
{\footnotesize\raggedright Four-answer success (\%) on 188 four-object scenes, with 24-family intervals. Receiver dictionaries, depth and strength are unchanged; the sender schema is extended to four objects. Complete facts have no model sender. This descriptive extension uses four templates; their wider intervals and receiver selection are reported in Appendix~\ref{app:recovery_four}.\par}
\end{table}

All three automatic reassignments give zero on Qwen; Gemma has one success
in one draw and zero in the other two. Both generation budgets agree.
Family intervals appear in Table~\ref{tab:recovery_four}; four-template
intervals are [67.7,97.3]\% and [73.4,100]\%, respectively. This extension
has no declared hypothesis family and remains descriptive.

Appendix~\ref{app:current_error_analysis} pairs these scenes across text and
internal delivery and checks their messages for missing facts. No-message
joint success is structurally zero under deterministic decoding; $0.25^4$
would instead assume independent uniform random guesses for the four answers.

\subsection{Message content and agreement between delivery routes}
\label{app:current_error_analysis}
Table~\ref{tab:current_sender_quality} checks the messages before any receiver
is used. It compares the decoded message symbols with the two required attributes of
each object, including the implementation-check scenes. A parsed message can
still omit a needed fact, and a message containing all facts can still fail
at the receiver. These content measures therefore complement joint answer success.
The four-object messages sometimes include an additional object noun such
as \emph{block}; such symbols are not counted as missing or incorrect attributes.

\begin{table}[!htbp]
\caption{\textbf{Facts retained by the automatic sender.}}
\label{tab:current_sender_quality}
\centering\small
\setlength{\tabcolsep}{3pt}\renewcommand{\arraystretch}{1.08}
\begin{tabular*}{\linewidth}{@{\extracolsep{\fill}}lrrrr@{}}
\toprule
Objects & Messages & Fact recall (\%) & Complete objects (\%) & Mean bytes \\
\midrule
Two & 1152 & 4570/4608 (99.2) & 2269/2304 (98.5) & 22.6 \\
Four & 384 & 2968/3072 (96.6) & 1432/1536 (93.2) & 36.0 \\
\bottomrule
\end{tabular*}
\par\vspace{3pt}
{\footnotesize\raggedright All generated messages parse, including the four engineering scenes in each bank. Recall counts required attribute symbols assigned to the correct object; complete objects contain both required facts. Extra selected symbols are allowed. These content checks are independent of the receiver and differ from four-answer task success.\par}
\end{table}

\begin{table}[!htbp]
\caption{\textbf{Which scenes succeed under the two delivery routes?}}
\label{tab:current_route_agreement}
\centering\small
\setlength{\tabcolsep}{3pt}\renewcommand{\arraystretch}{1.08}
\begin{tabular*}{\linewidth}{@{\extracolsep{\fill}}llrrrr@{}}
\toprule
Objects & Receiver & Both & Write only & Text only & Neither \\
\midrule
Two & Qwen3-32B & 553 & 2 & 2 & 15 \\
Two & Qwen3-14B-Base & 557 & 2 & 2 & 11 \\
Two & GLM-4-9B & 409 & 16 & 93 & 54 \\
Two & Gemma-3-27B-IT & 559 & 0 & 1 & 12 \\
Four & Qwen3-32B & 162 & 1 & 1 & 24 \\
Four & Gemma-3-27B-IT & 171 & 0 & 0 & 17 \\
\bottomrule
\end{tabular*}
\par\vspace{3pt}
{\footnotesize\raggedright Raw numbers of scenes with all four answers correct under the automatic message; 572 two-object and 188 four-object scenes. Counts come from independent answer-level recounts. This descriptive table keeps each scene paired across routes; it does not add another significance test.\par}
\end{table}

The scene-level pairing in Table~\ref{tab:current_route_agreement} makes the
text comparison more precise. On both Qwen receivers, two two-object scenes
succeed only through writing and two only through text. Their equal aggregate
scores therefore conceal different errors. On GLM, 93 scenes succeed as text
but fail through writing, versus sixteen in the other direction. All 93 of
the former contain complete automatic facts for both objects in both worlds,
so these failures arise after the sender has selected the facts. This does
not isolate which part of the receiver's processing causes them.

\begin{table}[!htbp]
\caption{\textbf{Automatic-message performance by object domain.}}
\label{tab:current_domain_results}
\centering\small
\setlength{\tabcolsep}{3pt}\renewcommand{\arraystretch}{1.08}
\begin{tabular*}{\linewidth}{@{\extracolsep{\fill}}llrrrr@{}}
\toprule
Receiver & Objects & Write count & Write (\%) & Text (\%) & Write valid (\%) \\
\midrule
Qwen3-32B & Blocks & 191/191 & 100.0 & 100.0 & 100.0 \\
 & Candy & 189/190 & 99.5 & 99.5 & 99.9 \\
 & Coats & 175/191 & 91.6 & 91.6 & 98.8 \\
Qwen3-14B-Base & Blocks & 191/191 & 100.0 & 100.0 & 100.0 \\
 & Candy & 190/190 & 100.0 & 100.0 & 100.0 \\
 & Coats & 178/191 & 93.2 & 93.2 & 99.9 \\
GLM-4-9B & Blocks & 174/191 & 91.1 & 97.4 & 97.6 \\
 & Candy & 162/190 & 85.3 & 99.5 & 96.2 \\
 & Coats & 89/191 & 46.6 & 66.5 & 83.1 \\
Gemma-3-27B-IT & Blocks & 191/191 & 100.0 & 100.0 & 100.0 \\
 & Candy & 190/190 & 100.0 & 100.0 & 100.0 \\
 & Coats & 178/191 & 93.2 & 93.7 & 100.0 \\
\bottomrule
\end{tabular*}
\par\vspace{3pt}
{\footnotesize\raggedright Descriptive within-domain averages on the confirmation bank, not family-macro estimates. The three domains contain 191, 190 and 191 evaluated scenes after engineering exclusions. Validity scores individual answers; success requires all four. Same packets and early-layer setting are used in every domain.\par}
\end{table}

GLM's losses concentrate in coats: 89/191 scenes succeed through writing,
versus 127/191 as text. Blocks and candy reach 174/191 and 162/190 through
writing. Table~\ref{tab:current_domain_results} retains the corresponding
breakdowns for every receiver. These are descriptive within-domain means;
they do not replace the fixed family-clustered main analysis. Coat descriptions
combine fibre and button material, so the observed domain gap alone cannot
distinguish lexical, formatting or binding difficulty.

For four objects, Gemma's seventeen failed scenes all omit a queried fact and
also fail as text. Qwen has nineteen failures with missing queried facts and
six with complete automatic facts in both worlds. Its text and write routes
each succeed uniquely on one scene. These counts motivate treating message
selection and receiver use as separate questions, even when their aggregate
accuracies happen to match.

\begin{figure}[!htbp]
\centering
\includegraphics[width=.6\linewidth]{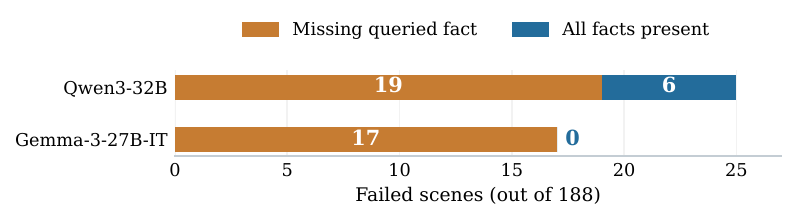}
\caption{\textbf{Missing facts accompany most initial four-object failures.}
Raw scene counts among automatic early-write failures, out of 188 scenes per
receiver in the single-attribute extension. The two categories exhaust
observed failures. This content check describes errors rather than their
unique causes.}
\label{fig:recovery_errors}
\end{figure}

\subsection{Timing internal and text delivery}
\label{app:recovery_cost}
The timing implementation constructs complete-fact messages on a fixed
eight-scene development subset; it does not load the automatic sender's
actual messages. Each of the sixteen worlds answers two questions with batch
size one and a 64-token budget. Of five repetitions per route, the first is
kept as cold (model already loaded) and the other four as warm; model loading
is excluded. Writing includes neutral prefill, direction construction,
intervention, cache reuse and decoding; text includes its prefix and decoding.
Sender work is common to the two routes and reported separately, while
local dictionary fitting is an offline cost reused across messages.

\begin{table}[!htbp]
\caption{\textbf{Receiver-side delivery cost for complete-fact messages.}}
\label{tab:recovery_cost}
\centering\small
\setlength{\tabcolsep}{3pt}\renewcommand{\arraystretch}{1.08}
\begin{tabular*}{\linewidth}{@{\extracolsep{\fill}}lrrrr@{}}
\toprule
 & \multicolumn{2}{c}{Early write (ms)} & \multicolumn{2}{c}{Text (ms)} \\\cmidrule(lr){2-3}\cmidrule(l){4-5} Receiver & Median & p90 & Median & p90 \\
\midrule
Qwen3-32B & 1301 & 1456 & 1018 & 1177 \\
Qwen3-14B-Base & 640 & 717 & 515 & 591 \\
GLM-4-9B & 651 & 736 & 545 & 656 \\
Gemma-3-27B-IT & 1300 & 1406 & 1051 & 1163 \\
\bottomrule
\end{tabular*}
\par\vspace{3pt}
{\footnotesize\raggedright One B200, FP32 eager, batch one, 64-token budget; each item answers two questions. Eight development scenes yield sixteen world-items, each repeated five times; four warm repetitions are summarized. These use supplied complete facts, not actual automatic packets. Sender generation and offline fitting are excluded.\par}
\end{table}

The complete-fact messages are 20 bytes and the padded weighted text averages
226 UTF-8 bytes on this subset. These are actual serialized representations,
not an optimized equal-budget comparison: the same indices can be rendered
as text at the receiver. This measurement supports no bandwidth or latency
advantage for internal delivery and does not measure automatic-system
end-to-end time.

\FloatBarrier
\section{Four-object questions and simpler representations}
\label{app:word_interface_followup}

These two studies reuse the selected early-layer interface on separate new
banks. Their scenes are not pooled with the preceding confirmation.

\subsection{Four-object conjunction queries}
\label{app:wordif_conjunction}

\paragraph{Task and fixed configuration.}
Each scene contains all four combinations of a color pair and a material
pair. Colors are green, orange, red and yellow; materials are plastic,
glass, rubber and copper. Crossing the six color pairs with the six material pairs yields 36 groups. Within
each group, sixteen scenes balance the changed attribute, the pair that
exchanges it, and the targets of the changed and unchanged queries.
All 24 object orders occur, with each answer position equally frequent.
Twelve evidence templates cross the groups; each contributes 48 scenes.
Queries alternate color-first and material-first wording. Automatic checks verify that each question has a unique answer, that each
attribute alone matches two blocks, and that one answer changes and the other
is preserved between worlds.

The 576 evaluation scenes have separate implementation-check scenes and
follow a 96-scene development bank. New names and description templates
exclude those in earlier banks; inventory symbols and attribute combinations
are reused. The four-object sender, receiver directions, layout, depth 0.2
and $\gamma=1$ are fixed before evaluation. The sender sees one world's
descriptions, never a query or the other world. Each message serves both queries and all four receivers. Complete-fact controls supply eight correct
attributes without a model sender.

\paragraph{Controls and statistical unit.}
Every condition contains 576 scenes and 2,304 answers. Invalid outputs
remain errors. Three symbol--direction reassignments keep the message bytes and positions and
match the actual edit norm; identity writes reproduce no-message outputs. The
other world's message reproduces that world's answers under the same
questions: scored against that world, its joint counts equal those of the
ordinary write. This is a paired consistency check, not additional independent data.
The 64- and 256-token internal joint results are identical.

Intervals use 20,000 bootstrap draws per seed for three fixed seeds,
resampling 36 groups and twelve templates separately. We report their
enclosing interval and the larger two-sided $p$ value. The contrast between internal writing and mean reassignment is
Holm-corrected across the four receivers, separately for each message
source. We declared in advance that reliable use additionally requires
$J\geq 0.5$, at least 90\% valid answers, and a positive lower bound of the
enclosing interval. Qwen and Gemma pass; GLM misses the absolute-success
threshold despite a positive corrected contrast. Text comparisons are
descriptive.

\begin{table}[t]
\caption{\textbf{Communicated facts support four-object conjunction queries.}}
\label{tab:wordif_conjunction}
\centering\small
\setlength{\tabcolsep}{3.5pt}\renewcommand{\arraystretch}{1.12}
\begin{tabular*}{\linewidth}{@{\extracolsep{\fill}}lrrrr@{}}
\toprule
Receiver & \textbf{Internal} & Text & \shortstack[r]{Full\\evidence} & \shortstack[r]{Reassigned\\Success / Valid} \\
\midrule
Qwen3-32B & \textbf{95.3} & 94.1 & 98.6 & 0.0 / 97.3 \\
Qwen3-14B-Base & \textbf{95.7} & 96.9 & 100.0 & 0.0 / 99.9 \\
GLM-4-9B & \textbf{41.8} & 70.3 & 87.7 & 0.0 / 62.2 \\
Gemma-3-27B-IT & \textbf{94.4} & 97.7 & 99.7 & 0.1 / 76.8 \\
\bottomrule
\end{tabular*}
\par\vspace{3pt}
{\footnotesize\raggedright Four-answer success (\%) on 576 new scenes requiring color and material jointly. Receiver configurations remain fixed. Text renders the same automatic message; full evidence provides the original descriptions. Intervals: Appendix~\ref{app:wordif_conjunction}. Reassigned averages three mappings. Full controls: Appendix~\ref{app:wordif_conjunction}.\par}
\end{table}

Table~\ref{tab:wordif_conjunction} summarizes the outcomes. Without receiver
retuning, three receivers from two families reach
\WIConjGemma{}--\WIConjBase{}\% four-answer success; GLM reaches only
\WIConjGLM{}\%, the main remaining limit. Pointing symbols to wrong directions
leaves at most 0.2\%. The sender's messages contain \WIConjRecall{}\% of the
required facts, so selection is not the bottleneck; GLM reaches only 44.6\%
even with complete facts written internally, versus 89.2\% as text.

\begin{table}[!htbp]
\caption{\textbf{Answer accuracy and selective use on conjunction queries.}}
\label{tab:wordif_conjunction_diagnostics}
\centering\small
\setlength{\tabcolsep}{3pt}\renewcommand{\arraystretch}{1.12}
\begin{tabular*}{\linewidth}{@{\extracolsep{\fill}}llrrrrrr@{}}
\toprule
Receiver & Message & \shortstack[r]{Joint\\count} & Accuracy & Valid & $R$ & $U$ & Text $J$ \\
\midrule
Qwen3-32B & Automatic & 549/576 & 98.8 & 99.5 & 96.4 & 99.0 & 94.1 \\
 & Complete facts & 558/576 & 99.2 & 99.6 & 97.6 & 99.3 & 97.7 \\
Qwen3-14B-Base & Automatic & 551/576 & 98.9 & 100.0 & 97.2 & 98.4 & 96.9 \\
 & Complete facts & 563/576 & 99.4 & 100.0 & 98.4 & 99.3 & 98.4 \\
GLM-4-9B & Automatic & 241/576 & 80.1 & 99.1 & 66.5 & 68.9 & 70.3 \\
 & Complete facts & 257/576 & 80.6 & 99.9 & 67.0 & 70.1 & 89.2 \\
Gemma-3-27B-IT & Automatic & 544/576 & 98.5 & 100.0 & 97.2 & 97.2 & 97.7 \\
 & Complete facts & 548/576 & 98.7 & 100.0 & 97.6 & 97.6 & 99.0 \\
\bottomrule
\end{tabular*}
\par\vspace{3pt}
{\footnotesize\raggedright All rates are percentages. Each source uses the same 576 scenes and 2,304 answers per receiver. $R$ and $U$ require the changed and unchanged queries, respectively, correct in both worlds. Complete facts contain the eight gold attributes without a model sender; all other settings remain fixed.\par}
\end{table}

\begin{table}[!htbp]
\caption{\textbf{Paired conjunction comparisons and reassignment validity.}}
\label{tab:wordif_conjunction_contrasts}
\centering\small
\setlength{\tabcolsep}{3pt}\renewcommand{\arraystretch}{1.12}
\begin{tabular*}{\linewidth}{@{\extracolsep{\fill}}lrrr@{}}
\toprule
Receiver & \shortstack[r]{Internal $-$ text\\{[95\% CI]}} & \shortstack[r]{Internal $-$ reassigned\\{[95\% CI]}} & \shortstack[r]{Valid answers (\%)\\Draws 1 / 2 / 3} \\
\midrule
Qwen3-32B & 1.2 [-0.5, 3.0] & 95.3 [92.9, 97.6] & 94.4/99.0/98.4 \\
Qwen3-14B-Base & -1.2 [-2.8, 0.2] & 95.7 [93.4, 97.6] & 99.8/100.0/100.0 \\
GLM-4-9B & -28.5 [-35.4, -20.8] & 41.8 [36.1, 47.6] & 57.1/63.4/66.1 \\
Gemma-3-27B-IT & -3.3 [-5.2, -1.6] & 94.4 [92.1, 96.4] & 88.6/56.9/84.9 \\
\bottomrule
\end{tabular*}
\par\vspace{3pt}
{\footnotesize\raggedright Automatic messages; differences are percentage points. Intervals enclose the separately resampled 36-group and twelve-template intervals. Reassigned is the within-scene mean of three mappings. All four reassignment contrasts pass the prespecified Holm correction; text differences are descriptive. Low reassignment validity contributes to losses on Gemma and GLM; Qwen3-14B retains at least 99.8\% validity.\par}
\end{table}

Across 1,152 evaluation messages, 1,150 parse fully and two partially;
1,142 contain every required fact. Required-fact recall is 99.86\%, and four
messages assign a scene value to the wrong object. GLM's complete-fact
write succeeds in 257/576 scenes, versus 514/576 as text, so correcting
selection alone does not remove its delivery gap. Reassignment can also
damage formatting, as Table~\ref{tab:wordif_conjunction_contrasts} shows.

\FloatBarrier
\subsection{Single-context directions and input embeddings}
\label{app:wordif_representations}

\paragraph{Construction and calibration.}
We compare the twelve-context mean with the first, middle and last
individual contexts in Appendix~\ref{app:final_fitting}, chosen by list
position before outcomes. The first places the symbol in the first field
of a two-position object layout; the middle and last use later fields in
four-position layouts. All use the same symbol-minus-unknown contrast and
the selected depth 0.2. Thus the comparison varies fitting context, including
position and layout width. It does not isolate either factor alone.

Each construction receives the same development strengths
$\{0.25,0.5,1,2,4\}$ and the existing coverage-constrained selection rule.
Selected strengths are one except GLM's first context (four) and input
embeddings (two). A separate 576-scene two-object bank retains the known
attributes and query patterns with new names and descriptions; four
implementation-check scenes are additional. No strength is selected on
confirmation outcomes. Tables~\ref{tab:wordif_representations}
and~\ref{tab:wordif_representation_contrasts} retain every single context
for both automatic and complete-fact messages.

Paired intervals separately resample material families and templates, with
four receivers in each comparison family for each message source. Independent
recounts reproduce estimates, intervals and all correction decisions.

\paragraph{Embedding identity and interpretation.}
The input control replaces a neutral embedding by that embedding plus
$\gamma$ times the symbol-minus-neutral embedding difference at its native
input location, preserving receiver-specific scaling. Every active symbol is
a single token in its field context on both banks, for all receivers and
both message sources. At $\gamma=1$ this is exactly ordinary message-text input; the full-span identity check gives zero logit difference. Multi-token
mean pooling is implemented but never used by these data.

Twelve-context fitting improves over the mean of three single contexts on
Qwen3-32B and GLM after correction, driven by the first context's poor
performance. The other two contexts differ by at most three points from
the mean direction, with no significant paired difference. Input embeddings
show no corrected disadvantage; they outperform internal writing on GLM.
The bank therefore supports the sufficiency of simpler representations at this early layer, rather than a necessity for twelve-context averaging.

\begin{table}[H]
\caption{\textbf{Simple local representations on a separate two-object bank.}}
\label{tab:wordif_representations}
\centering\small
\setlength{\tabcolsep}{3pt}\renewcommand{\arraystretch}{1.12}
\begin{tabular*}{\linewidth}{@{\extracolsep{\fill}}llrrrrrr@{}}
\toprule
Receiver & Message & Twelve & First & Middle & Last & Text & Embedding \\
\midrule
Qwen3-32B & Automatic & 95.3 & 87.2 & 95.1 & 95.3 & 94.8 & 94.8 \\
 & Complete facts & 98.8 & 94.6 & 98.8 & 98.6 & 99.3 & 99.3 \\
Qwen3-14B-Base & Automatic & 99.0 & 97.0 & 99.1 & 99.0 & 99.3 & 99.3 \\
 & Complete facts & 99.1 & 97.2 & 99.8 & 99.0 & 100.0 & 100.0 \\
GLM-4-9B & Automatic & 80.6 & 9.9 & 82.8 & 83.5 & 87.2 & 88.9 \\
 & Complete facts & 76.4 & 8.9 & 75.5 & 76.7 & 92.7 & 93.2 \\
Gemma-3-27B-IT & Automatic & 99.3 & 99.7 & 99.3 & 99.3 & 99.1 & 99.1 \\
 & Complete facts & 100.0 & 99.8 & 100.0 & 100.0 & 100.0 & 100.0 \\
\bottomrule
\end{tabular*}
\par\vspace{3pt}
{\footnotesize\raggedright Four-answer success (\%) on 576 new scenes. First, Middle and Last use fitting contexts 0, 6 and 11 separately. Strengths are chosen on the same development grid: $\gamma=1$ throughout except GLM First ($\gamma=4$) and Embedding ($\gamma=2$). Unit-strength embeddings exactly equal Text on this single-token bank. Embedding denotes the selected-strength input-layer control, not a projection into the internal writing layer.\par}
\end{table}

\begin{table}[H]
\caption{\textbf{What twelve-context fitting adds to the tested alternatives.}}
\label{tab:wordif_representation_contrasts}
\centering\small
\setlength{\tabcolsep}{3pt}\renewcommand{\arraystretch}{1.12}
\begin{tabular*}{\linewidth}{@{\extracolsep{\fill}}llrr@{}}
\toprule
Receiver & Message & \shortstack[r]{Twelve $-$ mean single\\{[95\% CI]}} & \shortstack[r]{Twelve $-$ embedding\\{[95\% CI]}} \\
\midrule
Qwen3-32B & Automatic & \textbf{2.8}\enspace {\footnotesize [1.0, 5.1]} & 0.5\enspace {\footnotesize [-0.7, 1.9]} \\
 & Complete facts & \textbf{1.4}\enspace {\footnotesize [0.6, 2.4]} & -0.5\enspace {\footnotesize [-1.2, 0.0]} \\
Qwen3-14B-Base & Automatic & 0.6\enspace {\footnotesize [0.0, 1.2]} & -0.3\enspace {\footnotesize [-1.2, 0.3]} \\
 & Complete facts & 0.5\enspace {\footnotesize [-0.8, 1.6]} & -0.9\enspace {\footnotesize [-2.1, 0.0]} \\
GLM-4-9B & Automatic & \textbf{21.8}\enspace {\footnotesize [17.1, 26.2]} & \textbf{-8.3}\enspace {\footnotesize [-12.0, -4.9]} \\
 & Complete facts & \textbf{22.7}\enspace {\footnotesize [17.3, 27.7]} & \textbf{-16.8}\enspace {\footnotesize [-27.1, -7.3]} \\
Gemma-3-27B-IT & Automatic & -0.1\enspace {\footnotesize [-0.3, 0.0]} & 0.2\enspace {\footnotesize [0.0, 0.5]} \\
 & Complete facts & 0.1\enspace {\footnotesize [0.0, 0.2]} & 0.0\enspace {\footnotesize [0.0, 0.0]} \\
\bottomrule
\end{tabular*}
\par\vspace{3pt}
{\footnotesize\raggedright Paired percentage-point differences at each construction\textquotesingle s development-selected strength. Intervals enclose family- and template-bootstrap intervals. Bold marks rejection after Holm correction within each four-receiver comparison family, separately by message source. GLM embedding uses $\gamma=2$; its unit-strength differences are $-6.6$ and $-16.3$ points for automatic and complete-fact messages. No other receiver shows a corrected embedding difference.\par}
\end{table}

\clearpage
\section{Earlier extraction: implementation and outcomes}
\label{app:earlier_interface}
The earlier system read sparse symbol weights (coefficients) from the sender's
hidden states and adjusted the corresponding coefficients in the receiver. It differs
from both the prompted sender and the fact-position write used in the main
results. Its positive binding effects and failures of selective fact use
motivated separating message selection from receiver-side delivery.

\subsection{Sparse messages and coefficient writing}
\label{app:protocol}\label{app:channel_overflow}
Each model independently fits a 221-symbol template-contrast dictionary at
relative depth 0.5. For each symbol, twelve frames each supply the mean of their last eight
residual-stream states; subtracting the corresponding states with
\emph{thing} in place of the symbol and normalizing gives a dictionary row.
This construction is distinct from the final-token, symbol-minus-\emph{unknown}
directions of Appendix~\ref{app:final_fitting} and from Jacobian-derived J-lens.

The sender averages description states after the first two tokens and fits
nonnegative coefficients by elastic net, with $\lambda_1=0.05$ and
$\lambda_2=10^{-3}$ \citep{jcot2607}. It transmits the four largest
coefficients per object as symbol indices and normalized 8-bit weights.
At each object's reserved positions and relative depths 0.4, 0.5 and 0.6,
the receiver fits its current coefficients $a^0$ and applies
\[
h[p]\leftarrow h[p]+D_S^{\mathsf T}(\hat a_S-a^0_S),
\]
where $S$ is the received support and $\hat a_S$ includes receiver-local
scale calibration. Nonorthogonal dictionary rows make this a replacement
of selected synthesis contributions, not independent latent coordinates.
The sender never sees the question or answer choices.

\subsection{Binding effects and generated answers}
\label{app:earlier_binding}\label{app:entdisj}
The initial task asks which of two named objects has a given attribute.
Its score is the log-probability gap between the first tokens of the two
names. The message-by-assignment interaction is the effect of exchanging the two
objects' messages from the sender minus the same exchange with an unrelated
episode's messages. A zero write through the same code must reproduce the
no-message computation exactly. Random-direction controls, with and without
run-time norm matching, keep the message bytes and assignments.

Seven original pairs were selected from a fourteen-receiver development
screen for the primary sender. All seven show a positive controlled score
effect; two later GLM pairs extend the attribute result. Replication uses
596 new examples per pair drawn from a previously evaluated entity pool,
with new combinations but no retuning on that pool. These selected pairs
do not estimate an unselected receiver success rate. Restricting the unrelated-episode messages to those within tolerance of the
sender's coefficients and planned write norm reduces
\ROBTOLFALL{}/\ROBTOLCELLS{} interactions to a median \ROBTOLRATIO{}
of their unrestricted values, so message statistics may contribute to the
original effect.

Symbol--direction reassignment is also sensitive to the random draw: the
original mapping beats the fixed reassignment on \ASEMSIG{}/\ASEMCELLS{}
pairs, but only \ADRSIG{}/\ADRCELLS{} pass the stricter rule of beating every
one of \ADRDRAWS{} draws. A later
128-draw analysis on three selected pairs supports an average advantage,
not robustness to every reassignment. Later GLM controls retain positive
contrasts on all six tested draws. Relation transfer does not pass the follow-up criteria on GLM or OLMo.

At the original strengths, no pair meets the answer-format requirement
for free generation. Lower strengths selected on separate development
material yield a net accuracy gain in \FGCNETGAIN{}/\FGCNETCELLS{}
evaluable pairs; invalid outputs remain errors. The largest receiver
does not yield an eligible generation configuration. Controlled first-token
effects therefore provide weaker evidence than the selective generated-answer
criterion used in the main paper.

\subsection{Single-fact confirmation}
\label{app:counterfactual}
The stronger test uses two candies with color/taste attributes or two
blocks with color/material attributes. One attribute exchanges objects
between worlds while the other stays fixed. The sender encodes each world
once, without either question, and that message must answer both questions.
The public receiver prompt is identical across worlds.

The 96-scene development bank contains twelve attribute-pair groups;
confirmation contains 720 scenes in thirty other groups, with new names
and description templates but shared concept vocabulary. The four pairs
are Llama-3.1-8B to Qwen3-14B-Base, Qwen3-32B and Qwen3.5-27B, and
Qwen3-8B-Base to GLM-4-9B. Development selects strengths from $\{0,1,2,4,8,16\}$ by answer accuracy
over all scenes, subject to 90\%
valid answers and smaller-strength tie breaking. Zero is retained when
writing offers no eligible improvement. Generation uses 64 tokens, with
256 as a sensitivity check; only an exact object name after case and
boundary-punctuation normalization is accepted. Invalid answers count as
errors. The receiver uses FP32 eager inference with batch size one, separately
from the earlier BF16 binding regime.

\begin{table}[!htbp]
\caption{\textbf{Selective fact use with the earlier sparse extractor.}}
\label{tab:earlier_counterfactual}
\centering\footnotesize
\setlength{\tabcolsep}{3pt}
\renewcommand{\arraystretch}{1.05}
\begin{tabular*}{\linewidth}{@{\extracolsep{\fill}}lrrrrr@{}}
\toprule
Receiver & $\gamma$ & Internal & Message text & Full evidence & Internal validity \\
\midrule
Qwen3-14B-Base & 1 & 0.1 & 24.3 & 100.0 & 99.2 \\
Qwen3-32B & 1 & 0.3 & 11.8 & 100.0 & 90.1 \\
GLM-4-9B & 0 & 0.0 & 7.9 & 99.0 & 96.8 \\
Qwen3.5-27B & 0$^{\dagger}$ & 0.0 & 17.8 & 100.0 & 60.5 \\
\bottomrule
\end{tabular*}
\par\vspace{3pt}
{\footnotesize\raggedright All values except $\gamma$ are percentages. Internal, message text and full evidence report all four answers correct; validity reports the fraction of generated internal answers that parse. Each pair has 720 scenes. Text prints the selected symbols and their weights. All no-message joint scores are zero. Strength zero is no write; $\dagger$ marks a failed format requirement. Invalid answers remain errors.\par}
\end{table}

The prespecified behavioral criterion requires at least five percentage
points of gain, a positive 95\% lower limit and Holm-adjusted significance.
The primary family has twelve tests: four pairs by answer-accuracy gain,
four-answer-success gain and relevant full-name score contrast. Intervals
resample thirty attribute-pair groups with both worlds and questions kept
together; the reported interval encloses those from three seeds of 20,000 draws.
No automatic sparse-message pair meets the selective-fact criterion.
Full-evidence text verifies task solvability, while weighted-message text
also exposes information lost before delivery. These outcomes concern the earlier sparse extraction and coefficient
writing; they are not results for the early-layer
configuration in Tables~\ref{tab:recovery_main}--\ref{tab:recovery_four}.

\FloatBarrier
\section{Message selection and direction construction}
\label{app:mechanism_diagnosis}
The following comparisons use previously evaluated development scenes and
the earlier middle-layer configuration. They distinguish selection errors
from delivery failures and compare direction constructions. They are
separate from the early-layer confirmation in Appendix~\ref{app:final_evidence}.

\subsection{Same-checkpoint message sources}
\label{app:automatic_development_matrix}
Two 96-scene banks, each with twelve attribute-pair groups, compare four
message sources using Llama-3.1-8B-Instruct throughout. Original sparse
extraction encodes the sender's hidden state; centered extraction subtracts
a generic reference state before encoding. Prompted selection uses the
current JSON-based selector. Complete facts are supplied directly, without
a model sender. Within each row, the message is fixed across the three delivery routes in Table~\ref{tab:automatic_development_matrix}.

\begin{table}[!htbp]
\caption{\textbf{Message sources and delivery routes on development material.}}
\label{tab:automatic_development_matrix}
\centering\small
\setlength{\tabcolsep}{4pt}\renewcommand{\arraystretch}{1.08}
\begin{tabular*}{\linewidth}{@{\extracolsep{\fill}}lrrrrrr@{}}
\toprule
 & \multicolumn{3}{c}{Candy / blocks} & \multicolumn{3}{c}{Coats} \\
\cmidrule(lr){2-4}\cmidrule(l){5-7}
Message source & Text & Old & Targeted & Text & Old & Targeted \\
\midrule
Original sparse & 4.2 & 1.0 & 1.0 & 5.2 & 1.0 & 7.3 \\
Centered sparse & 70.8 & 0.0 & 38.5 & 8.3 & 1.0 & 13.5 \\
Prompted selection & 100.0 & 0.0 & 95.8 & 100.0 & 0.0 & 77.1 \\
Complete facts & 100.0 & 2.1 & 99.0 & 100.0 & 1.0 & 87.5 \\
\bottomrule
\end{tabular*}
\par\vspace{3pt}
{\footnotesize\raggedright Four-answer success (\%), 96 previously evaluated scenes per bank; Llama-3.1-8B-Instruct $\rightarrow$ Qwen3-32B. Old strengths are selected on candy/blocks only (2, 2, 2, 1 in row order) and reused on coats. Old and fact-position configurations differ in dictionary, actual write positions and update rule. Automatic messages reserve four positions per object; complete facts reserve two.\par}
\end{table}

The old configuration uses one middle-layer dictionary, object-level
positions and coefficient updates; the targeted configuration uses a
dictionary at each written layer, one position per fact and
neutral-state-plus-direction writes at depths 0.4/0.5/0.6 with $\gamma=1$. Old strengths are chosen
on candy/block scenes from $\{1,2,4,8,16\}$, maximizing joint success
among settings with 90\% valid answers, then accuracy and smaller strength;
the same choices apply to coats. Their comparison changes multiple
receiver components and does not isolate the update rule.

Prompted selection outperforms same-checkpoint sparse extraction through
both writing and text. Prompted-minus-original internal success is
94.8 points [87.5, 100.0] on candy/blocks and 69.8 [54.2, 84.4] on coats.
Its difference from complete-fact writing is $-3.1$ [$-8.3$, 2.1] and
$-10.4$ [$-19.8$, $-2.1$], respectively. These paired 95\% intervals
resample attribute-pair groups. Message sources differ in support and
weights; complete facts also use two positions per object rather than
four. The comparison therefore does not isolate content loss or establish
equivalence with complete facts. Delivering the same message as text compares content, not equal byte
counts.

\FloatBarrier
\subsection{Each-layer directions and the update rule}
\label{app:receiver_ablation}
This development comparison holds complete-fact messages, prompt, fact
positions, neutral states, fitting contexts and the 0.4/0.5/0.6 layer band
fixed. Each-layer dictionaries use the contrast fitted at the written
layer; the midpoint condition reuses the center-layer contrast everywhere.
Rescaled midpoint rows match each-layer direction norms. The coefficient
update uses the same fitting contexts and one symbol at each fact position,
with calibration recomputed per dictionary and layer; it is an update-rule
ablation, not a rerun of the old end-to-end system.

\begin{table}[!htbp]
\caption{\textbf{Dictionary depth and update rule at a common strength.}}
\label{tab:receiver_ablation_summary}
\centering\small
\setlength{\tabcolsep}{4pt}\renewcommand{\arraystretch}{1.08}
\begin{tabular*}{\linewidth}{@{\extracolsep{\fill}}lrrrrr@{}}
\toprule
 & \multicolumn{3}{c}{Neutral + direction} & \multicolumn{2}{c}{Coefficient update} \\
\cmidrule(lr){2-4}\cmidrule(l){5-6}
Receiver & Each layer & Midpoint & Rescaled mid & Each layer & Midpoint \\
\midrule
Qwen3-32B & 99.0 & 54.2 & 61.5 & 0.0 & 0.0 \\
Qwen3-14B-Base & 13.5 & 13.5 & 15.6 & 0.0 & 0.0 \\
GLM-4-9B & 6.2 & 8.3 & 8.3 & 0.0 & 0.0 \\
\bottomrule
\end{tabular*}
\par\vspace{3pt}
{\footnotesize\raggedright Four-answer success (\%) at $\gamma=1$ on 96 development scenes with supplied complete facts. Rescaled midpoint directions match each-layer direction-row norms. Message, fitting recipe, fact positions and written depths stay fixed. Equal strength does not match the edit magnitude of the two update rules. GLM coefficient-update cells fail 90\% format coverage throughout the grid.\par}
\end{table}

On Qwen3-32B at $\gamma=1$, the each-layer gain over midpoint directions
is 44.8 points [31.2, 58.3]. Rescaling the midpoint directions' norms accounts for 7.3 [3.1, 12.5],
leaving 37.5 [26.0, 49.0] when the rescaled directions are replaced by
each-layer directions. Qwen3-14B-Base shows no corresponding gain, and
GLM remains at or below 8.3\% throughout the shared strength grid
$\{0.25,0.5,1,2,4\}$. Invalid answers stay in all denominators.

Equal strength does not equalize the two update rules' realized edits.
For Qwen3-32B, mean edit norms at the three layers are 363/132/173 for
each-layer neutral-plus-direction writing at strength one, versus
72/54/131 for coefficient updates even at strength four. Pre-write
residual norms are about 198/212/206. Reused midpoint directions make
larger edits, 419/300/302, despite lower success. The each-layer advantage
within the neutral-plus-direction update therefore cannot be explained
simply by a larger edit; superiority over coefficient updates at matched
edit norms remains untested. The positions themselves are held fixed,
so this comparison also does not isolate a benefit from separate fact positions.

\FloatBarrier
\subsection{Symbol contrasts and J-lens at one layer}
\label{app:dictionary_compatibility}
A matched comparison on 96 development scenes writes supplied complete
facts at the center layer only. Symbol-contrast and Jacobian-derived J-lens
directions use identical messages, public fact positions and $\gamma=1$;
each J-lens row is scaled to the corresponding symbol-contrast row norm.

\begin{table}[!htbp]
\caption{\textbf{Dictionary construction matters at a matched single layer.}}
\label{tab:dictionary_single_layer}
\centering\small
\begin{tabular*}{\linewidth}{@{\extracolsep{\fill}}lrrr@{}}
\toprule
Receiver & Symbol contrast (\%) & J-lens (\%) & J-lens minus contrast (pp) \\
\midrule
Qwen3-32B & 97.9 & 2.1 & -95.8\enspace [-99.0, -92.7] \\
Qwen3-14B-Base & 10.4 & 1.0 & -9.4\enspace [-15.6, -4.2] \\
GLM-4-9B & 0.0 & 1.0 & 1.0\enspace [0.0, 3.1] \\
Llama-3.1-8B-Instruct & 0.0 & 0.0 & 0.0\enspace [0.0, 0.0] \\
\bottomrule
\end{tabular*}
\par\vspace{3pt}
{\footnotesize\raggedright All four answers correct on 96 exposed development scenes. Both conditions use the band-center layer, the same positions, $\gamma=1$, and per-symbol direction norms matched to the symbol-contrast dictionary. Intervals are paired 95\% family-bootstrap intervals. Row-norm matching does not equate every realized residual edit.\par}
\end{table}

On Qwen3-32B, one symbol-contrast write already supports high joint success;
three writes are not necessary in this setting. J-lens performs much worse
there and on Qwen3-14B-Base, while both constructions are near zero on
GLM and Llama. How the directions are constructed therefore matters for this update. The
comparison does not test early-layer J-lens writing or reading meanings out
with the J-lens,
and row-norm matching does not equate every realized residual edit.
The result cannot by itself distinguish direction orientation, fitting
objective and compatibility with the evaluation context.

\FloatBarrier
\section{Scope of the earlier utility results}
\label{app:utility}\label{app:repro}
Countries and TipSheets adapt the private-information tasks of
\citet{ramesh2025communicating}. Countries asks for a person's country
from location evidence; TipSheets asks for an action from a company's
profile and decision rules. Each uses 400 development and 400 test items,
with disjoint person--landmark or company--profile combinations and balanced
answer letters. The sender sees the private evidence and identifiers,
never the receiver's question or answer choices.

This study uses two senders, four receivers and both tasks, giving sixteen
configurations of the earlier sparse extractor and coefficient writer.
The models use independently fitted 20,431-symbol template-contrast
dictionaries, four selected symbols per object and 8-bit weights. Evaluation
compares four single-token answer letters. Text with the same selected symbols omits the weights, and full-evidence text
supplies the original private descriptions. Both differ from the weighted-text
route of the current method.

Useful delivery requires a positive 95\% lower limit for improvement over
no message and a mean gain at least 30\% of the full-evidence text gain.
The primary comparison is Llama-3.1-8B to Qwen3-32B on Countries;
the other fifteen form a Holm family. Intervals use 10,000 hierarchical
bootstrap draws over correct-answer entities, keeping the three recorded
seeds grouped within each example. All sixteen configurations fail this
criterion. Countries shows that selected symbols can be useful as text
despite weak internal delivery; TipSheets also has a selection limitation.
Budget-aware text outperforms the original internal route at every tested
message budget. These findings do not evaluate the prompted sender with
the early-layer fact-position writer.

The earlier implementation's latency includes repeated sparse-coefficient
solves at written positions and layers. It cannot estimate the current
writer's cost, which is measured separately in Appendix~\ref{app:recovery_cost}.
Likewise, studies with different numerical settings, message formats or
evaluation banks are reported separately rather than pooled. Only valid
completed comparisons enter the summaries above; independent recounts
check the saved predictions, counts and statistics.
Appendix~\ref{app:countries} evaluates the final configuration on a
two-world version of Countries; TipSheets remains untested with it.

\FloatBarrier


\begin{thebibliography}{22}
\providecommand{\natexlab}[1]{#1}
\providecommand{\url}[1]{\texttt{#1}}
\expandafter\ifx\csname urlstyle\endcsname\relax
  \providecommand{\doi}[1]{doi: #1}\else
  \providecommand{\doi}{doi: \begingroup \urlstyle{rm}\Url}\fi

\bibitem[Buehler(2026)]{reading2607}
Markus~J. Buehler.
\newblock Reading and steering representations of materials-science mechanisms
  in an open-weight language model.
\newblock \emph{arXiv preprint arXiv:2607.20058}, 2026.

\bibitem[Chen et~al.(2025)Chen, Merullo, Stolfo, and Pavlick]{stitching2512}
Alan Chen, Jack Merullo, Alessandro Stolfo, and Ellie Pavlick.
\newblock Transferring linear features across language models with model
  stitching.
\newblock In \emph{Advances in Neural Information Processing Systems},
  volume~38, pp.\  48531--48563. Curran Associates, Inc., 2025.
\newblock \doi{10.52202/085713-1621}.
\newblock URL
  \url{https://proceedings.neurips.cc/paper_files/paper/2025/file/4569a868e7aa891248832ec08445d071-Paper-Conference.pdf}.

\bibitem[Chen et~al.(2024)Chen, Vondrick, and Mao]{chen2024selfie}
Haozhe Chen, Carl Vondrick, and Chengzhi Mao.
\newblock {SelfIE}: Self-interpretation of large language model embeddings.
\newblock In \emph{Proceedings of the 41st International Conference on Machine
  Learning}, volume 235 of \emph{Proceedings of Machine Learning Research},
  pp.\  7373--7388. PMLR, 2024.
\newblock URL \url{https://proceedings.mlr.press/v235/chen24ao.html}.

\bibitem[Du et~al.(2024)Du, Li, Torralba, Tenenbaum, and
  Mordatch]{du2024debate}
Yilun Du, Shuang Li, Antonio Torralba, Joshua~B. Tenenbaum, and Igor Mordatch.
\newblock Improving factuality and reasoning in language models through
  multiagent debate.
\newblock In \emph{Proceedings of the 41st International Conference on Machine
  Learning}, volume 235 of \emph{Proceedings of Machine Learning Research},
  pp.\  11733--11763. PMLR, 2024.
\newblock URL \url{https://proceedings.mlr.press/v235/du24e.html}.

\bibitem[Fu et~al.(2026)Fu, Min, Zhang, Yan, Dai, Ouyang, and
  Wang]{cachetocache2510}
Tianyu Fu, Zihan Min, Hanling Zhang, Jichao Yan, Guohao Dai, Wanli Ouyang, and
  Yu~Wang.
\newblock {Cache-to-Cache}: Direct semantic communication between large
  language models.
\newblock In \emph{International Conference on Learning Representations}, pp.\
  43130--43158, 2026.
\newblock URL
  \url{https://proceedings.iclr.cc/paper_files/paper/2026/file/474ada926b331d78f06d95e8913111cc-Paper-Conference.pdf}.

\bibitem[Gan et~al.(2026)Gan, Li, and Luo]{crest2608}
Jin Gan, Xin Li, and Jun Luo.
\newblock Beyond token-level guidance: Inference-time alignment of specialized
  {LLM}s via cross-family representation steering.
\newblock \emph{arXiv preprint arXiv:2608.30319}, 2026.

\bibitem[Ghandeharioun et~al.(2024)Ghandeharioun, Caciularu, Pearce, Dixon, and
  Geva]{ghandeharioun2024patchscopes}
Asma Ghandeharioun, Avi Caciularu, Adam Pearce, Lucas Dixon, and Mor Geva.
\newblock Patchscopes: A unifying framework for inspecting hidden
  representations of language models.
\newblock In \emph{Proceedings of the 41st International Conference on Machine
  Learning}, volume 235 of \emph{Proceedings of Machine Learning Research},
  pp.\  15466--15490. PMLR, 2024.
\newblock URL \url{https://proceedings.mlr.press/v235/ghandeharioun24a.html}.

\bibitem[Gong \& Wang(2026)Gong and Wang]{firsttoken2608}
Xijie Gong and Tonghan Wang.
\newblock The first token is a clue: Verbalizing multi-token concepts from the
  {J-lens}.
\newblock \emph{arXiv preprint arXiv:2608.31084}, 2026.

\bibitem[Gurnee et~al.(2026)Gurnee, Sofroniew, Pearce, Piotrowski, Kauvar,
  Chen, Soligo, Bogdan, Ong, Wang, Thompson, Abrahams, Kantamneni, Ameisen,
  Batson, and Lindsey]{verbalizable2607}
Wes Gurnee, Nicholas Sofroniew, Adam Pearce, Mateusz Piotrowski, Isaac Kauvar,
  Runjin Chen, Anna Soligo, Paul Bogdan, Euan Ong, Rowan Wang, Ben Thompson,
  David Abrahams, Subhash Kantamneni, Emmanuel Ameisen, Joshua Batson, and Jack
  Lindsey.
\newblock Verbalizable representations form a global workspace in language
  models.
\newblock \emph{Transformer Circuits Thread}, 2026.
\newblock URL \url{https://transformer-circuits.pub/2026/workspace/index.html}.

\bibitem[Hong et~al.(2024)Hong, Zhuge, Chen, Zheng, Cheng, Wang, Zhang, Wang,
  Yau, Lin, Zhou, Ran, Xiao, Wu, and Schmidhuber]{hong2024metagpt}
Sirui Hong, Mingchen Zhuge, Jonathan Chen, Xiawu Zheng, Yuheng Cheng, Jinlin
  Wang, Ceyao Zhang, Zili Wang, Steven Yau, Zijuan Lin, Liyang Zhou, Chenyu
  Ran, Lingfeng Xiao, Chenglin Wu, and J{\"u}rgen Schmidhuber.
\newblock {MetaGPT}: Meta programming for a multi-agent collaborative
  framework.
\newblock In \emph{The Twelfth International Conference on Learning
  Representations}, 2024.
\newblock URL \url{https://openreview.net/forum?id=VtmBAGCN7o}.

\bibitem[Huh et~al.(2024)Huh, Cheung, Wang, and Isola]{huh2024platonic}
Minyoung Huh, Brian Cheung, Tongzhou Wang, and Phillip Isola.
\newblock The platonic representation hypothesis.
\newblock In \emph{Proceedings of the 41st International Conference on Machine
  Learning}, volume 235 of \emph{Proceedings of Machine Learning Research},
  pp.\  20617--20642. PMLR, 2024.
\newblock URL \url{https://proceedings.mlr.press/v235/huh24a.html}.

\bibitem[Liu et~al.(2026)Liu, Zhang, Jia, Kan, and Wang]{xkv2608}
Jiyao Liu, Qi~Zhang, Yaoyi Jia, Ziwen Kan, and Song Wang.
\newblock Dual-cache latent space communication between heterogeneous language
  models.
\newblock \emph{arXiv preprint arXiv:2608.20617}, 2026.
\newblock URL \url{https://arxiv.org/abs/2608.20617}.

\bibitem[Moschella et~al.(2023)Moschella, Maiorca, Fumero, Norelli, Locatello,
  and Rodol{\`a}]{relative2209}
Luca Moschella, Valentino Maiorca, Marco Fumero, Antonio Norelli, Francesco
  Locatello, and Emanuele Rodol{\`a}.
\newblock Relative representations enable zero-shot latent space communication.
\newblock In \emph{International Conference on Learning Representations}, 2023.

\bibitem[Pham et~al.(2024)Pham, Liu, Yang, Chen, Liu, Yuan, Plummer, Wang, and
  Yang]{cipher2310}
Chau Pham, Boyi Liu, Yingxiang Yang, Zhengyu Chen, Tianyi Liu, Jianbo Yuan,
  Bryan~A. Plummer, Zhaoran Wang, and Hongxia Yang.
\newblock Let models speak ciphers: Multiagent debate through embeddings.
\newblock In \emph{International Conference on Learning Representations}, 2024.

\bibitem[Ramesh \& Li(2025)Ramesh and Li]{ramesh2025communicating}
Vignav Ramesh and Kenneth Li.
\newblock Communicating activations between language model agents.
\newblock In \emph{Proceedings of the 42nd International Conference on Machine
  Learning}, volume 267 of \emph{Proceedings of Machine Learning Research},
  pp.\  51094--51116. PMLR, 2025.
\newblock URL \url{https://proceedings.mlr.press/v267/ramesh25a.html}.

\bibitem[Rimsky et~al.(2024)Rimsky, Gabrieli, Schulz, Tong, Hubinger, and
  Turner]{rimsky2024caa}
Nina Rimsky, Nick Gabrieli, Julian Schulz, Meg Tong, Evan Hubinger, and
  Alexander Turner.
\newblock Steering {Llama 2} via contrastive activation addition.
\newblock In \emph{Proceedings of the 62nd Annual Meeting of the Association
  for Computational Linguistics (Volume 1: Long Papers)}, pp.\  15504--15522,
  2024.
\newblock \doi{10.18653/v1/2024.acl-long.828}.
\newblock URL \url{https://aclanthology.org/2024.acl-long.828/}.

\bibitem[Turner et~al.(2024)Turner, Thiergart, Leech, Udell, Vazquez, Mini, and
  MacDiarmid]{turner2023actadd}
Alexander~Matt Turner, Lisa Thiergart, Gavin Leech, David Udell, Juan~J.
  Vazquez, Ulisse Mini, and Monte MacDiarmid.
\newblock Steering language models with activation engineering, 2024.
\newblock URL \url{https://arxiv.org/abs/2308.10248}.

\bibitem[Wang et~al.(2026)Wang, Schwarzschild, Robey, Payani, Fleming, Sun, and
  Ippolito]{wang2025commandv}
Barry Wang, Avi Schwarzschild, Alexander Robey, Ali Payani, Charles Fleming,
  Mingjie Sun, and Daphne Ippolito.
\newblock Command-v: Training-free representation finetuning transfer.
\newblock In \emph{The Fourteenth International Conference on Learning
  Representations}, 2026.
\newblock URL \url{https://openreview.net/forum?id=oRYzpI3cmJ}.

\bibitem[Wu et~al.(2026)Wu, Zhu, Liu, Hu, and Pan]{jcot2607}
Junde Wu, Jiayuan Zhu, Fengling Liu, Minhao Hu, and Jiazhen Pan.
\newblock {J-CoT}: Chain-of-thought in {J-Space}.
\newblock \emph{arXiv preprint arXiv:2607.21981}, 2026.

\bibitem[Wu et~al.(2024)Wu, Bansal, Zhang, Wu, Li, Zhu, Jiang, Zhang, Zhang,
  Liu, Awadallah, White, Burger, and Wang]{wu2024autogen}
Qingyun Wu, Gagan Bansal, Jieyu Zhang, Yiran Wu, Beibin Li, Erkang Zhu,
  Li~Jiang, Xiaoyun Zhang, Shaokun Zhang, Jiale Liu, Ahmed Awadallah, Ryen~W.
  White, Doug Burger, and Chi Wang.
\newblock {AutoGen}: Enabling next-gen {LLM} applications via multi-agent
  conversations.
\newblock In \emph{First Conference on Language Modeling}, 2024.
\newblock URL \url{https://openreview.net/forum?id=BAakY1hNKS}.

\bibitem[Yang et~al.(2026)Yang, Huang, Zhang, Wang, Yu, and Lee]{xbridge2608}
Wooseong Yang, Wei-Chieh Huang, Weizhi Zhang, Yu~Wang, Philip~S. Yu, and
  Junhyun Lee.
\newblock {XBridge}: Entity-grounded latent bridge for heterogeneous {LLM}
  communication.
\newblock \emph{arXiv preprint arXiv:2608.11676}, 2026.

\bibitem[Zou et~al.(2026)Zou, Qiu, Li, Yang, Tieu, Lu, Shen, Tong, Choi, He,
  Zou, Wang, and Yang]{zou2025latentmas}
Jiaru Zou, Ruizhong Qiu, Gaotang Li, Xiyuan Yang, Katherine Tieu, Pan Lu,
  Ke~Shen, Hanghang Tong, Yejin Choi, Jingrui He, James Zou, Mengdi Wang, and
  Ling Yang.
\newblock Latent collaboration in multi-agent systems.
\newblock In \emph{Forty-third International Conference on Machine Learning},
  2026.
\newblock URL \url{https://openreview.net/forum?id=syG9I9ofd8}.

\end{thebibliography}
\end{document}